\documentclass{article}

\usepackage[margin=2cm]{geometry}

\usepackage{multirow}
\usepackage{xspace}
\usepackage{hyperref}

\usepackage{multirow}
\usepackage{xspace}
\usepackage{longtable}
\usepackage{ulem}
\usepackage{float}
\usepackage{amsmath}

\newcommand{\rev}[1]{{\color{black}#1}\xspace}
\newcommand{\revDue}[1]{{\color{black}#1}\xspace}

\newcommand{\rispmean}[1]{\textcolor{red}{Mean: #1 }}
\newcommand{\rispF}[1]{\textcolor{blue}{F: #1 }}
\newcommand{\rispacc}[1]{\textcolor{green}{Acc: #1 }}

\usepackage{amsfonts}
\usepackage{amssymb}
\usepackage{amsthm}
\usepackage{amsmath}

\newcommand{\N}{\mathbb{N}}
\newcommand{\R}{\mathbb{R}}
\newcommand{\bA}{\boldsymbol{A}}
\newcommand{\bI}{\boldsymbol{I}}
\newcommand{\bD}{\boldsymbol{D}}
\newcommand{\bAt}{\boldsymbol{\tilde{A}}}
\newcommand{\bDt}{\boldsymbol{\tilde{D}}}
\newcommand{\bW}{\boldsymbol{W}}
\newcommand{\bL}{\boldsymbol{L}}
\newcommand{\bLt}{\boldsymbol{\tilde{L}}}
\newcommand{\bX}{\boldsymbol{X}}
\newcommand{\bZ}{\boldsymbol{Z}}
\DeclareMathOperator{\lrelu}{LeakyReLU}
\DeclareMathOperator{\relu}{ReLU}

\newcommand{\bS}{\boldsymbol{S}}
\DeclareMathOperator{\sftmax}{softmax}
\DeclareMathOperator{\aucroc}{AucROC}

\DeclareMathOperator{\diag}{diag}
\DeclareMathOperator{\argmax}{argmax}
\DeclareMathOperator{\score}{Score}

\newcommand{\gcn}{\textsc{Gcn}\xspace}
\newcommand{\gat}{\textsc{Gat}\xspace}
\newcommand{\sage}{\textsc{GraphSage}\xspace}
\newcommand{\cheb}{\textsc{Cheb}\xspace}
\newcommand{\gin}{\textsc{Gin}\xspace}
\newcommand{\mincut}{\textsc{MinCutPool}\xspace}
\newcommand{\setset}{\textsc{Set2Set}\xspace}
\newcommand{\graphconv}{\textsc{GraphConv}\xspace}

\newcommand{\cam}{\textsc{Cam}\xspace}
\newcommand{\gradcam}{\textsc{GradCam}\xspace}
\newcommand{\gradexp}{\textsc{GradExplNode}\xspace}
\newcommand{\guidedbp}{\textsc{GuidedBp}\xspace}
\newcommand{\ig}{\textsc{IntegratedGradients}\xspace}
\newcommand{\ignode}{\textsc{IgNode}\xspace}
\newcommand{\pgmexpl}{\textsc{PgmExpl}\xspace}
\newcommand{\gnnexpl}{\textsc{GnnExpl}\xspace}
\newcommand{\pgexpl}{\textsc{PgExpl}\xspace}
\newcommand{\igedge}{\textsc{IgEdge}\xspace}
\newcommand{\sal}{\textsc{GradExplEdge}\xspace}

\newcommand{\rgexpl}{\textsc{RgExpl}\xspace}
\newcommand{\subgraphx}{\textsc{SubX}\xspace}
\newcommand{\letterExpl}{\textsc{H}\xspace}

\newcommand{\datasetgcone}{\textsc{Grid}\xspace}
\newcommand{\datasetgctwo}{\textsc{Grid-House}\xspace}
\newcommand{\datasetgcthree}{\textsc{Stars}\xspace}
\newcommand{\datasetgcfour}{\textsc{House-Color}\xspace}
\newcommand{\datasetncone}{\textsc{Shapes}\xspace}
\newcommand{\datasetnctwo}{\textsc{Infection}\xspace}

\usepackage{todonotes}
\usepackage{makecell}
\usepackage{enumitem}

\usepackage{colortbl}

\newlength{\toprulewidth}
\patchcmd{\toprule}% <cmd>
  {\heavyrulewidth}{\toprulewidth}% <search><replace>
  {}{}% <success><failure>

\usepackage{pifont}

\newcommand{\filter}{\ding{34}}%

\usepackage{appendix}

\usepackage{authblk}
\usepackage{booktabs}
\begin{document}

%\title{Are Explainers for Graph Neural Networks Explainable? An Experimental Survey}
\title{Explaining the Explainers in Graph Neural Networks: a Comparative Study}
\author[1,2]{Antonio Longa}
\author[1]{Steve Azzolin}
\author[3]{Gabriele Santin}
\author[4]{Giulia Cencetti}
\author[5]{Pietro Li{\`o}}
\author[2]{Bruno Lepri}
\author[1]{Andrea Passerini}

\affil[1]{University of Trento}
\affil[2]{Fondazione Bruno Kessler}
\affil[3]{University of Venice}
\affil[4]{CNRS, Centre de Physique Theorique}
\affil[5]{University of Cambridge}

\maketitle

\begin{abstract}
Following a fast initial breakthrough in graph based learning, Graph Neural Networks (GNNs) have reached a widespread application in many science and engineering fields, prompting the need for methods to understand their decision process. GNN explainers have started to emerge in recent years, with a multitude of methods both novel or adapted from other domains. To sort out this plethora of alternative approaches, several studies have benchmarked the performance of different explainers in terms of various explainability metrics. However, these earlier works make no attempts at providing insights into why different GNN architectures are more or less explainable, or which explainer should be preferred in a given setting.
In this survey we fill these gaps by devising a systematic experimental study, which tests twelve explainers on eight representative \revDue{message-passing} architectures trained on six carefully designed graph and node classification datasets. With our results we provide key insights on the choice and applicability of GNN explainers, we isolate key components that make them usable and successful and provide recommendations on how to avoid common interpretation pitfalls. We conclude by highlighting open questions and directions of possible future research.
\end{abstract}

\section{Introduction}\label{sec:introandmotivation}
Graph Neural Networks (GNNs) have emerged as the de-facto standard for graph-based learning tasks. Regardless of their apparent simplicity, that allows most GNN architectures to be expressed as variants of Message Passing~\cite{Gilmer2017}, i.e., exchanging messages between nodes, GNNs have proved extremely effective in preserving the natural symmetries present in many real-world physical systems~\cite{wang2022approximately, jaini2021learning, puny2021frame, fu2022simulate, basu2022equivariant}. The versatility of GNNs allowed them to be also applied to emulate classical algorithms~\cite{ijcai2021p595}, addressing tasks like bipartite matching~\cite{dobrik2020}, graph coloring~\cite{Li2022} or the Traveling Salesperson Problem~\cite{prates2019}, and approximate symbolic reasoning tasks like propositional satisfiability~\cite{selsam2018learning,Selsam2019,wang:arxiv21} and probabilistic logic reasoning~\cite{Zhang2020Efficient}. Despite recent works trying to adapt the Transformer architecture, made popular by a wide success first in language~\cite{VaswaniAttention, roccabruna-2022-multi, raffel2020exploring} and then in vision applications~\cite{ramesh2022hierarchical,andreini2021two,liu2021survey}, to the graph domain~\cite{kim2022pure, rampavsek2022recipe, yun2019graph, dwivedi2020generalization, Zhang2022HierarchicalGT}, the natural inductive bias of GNNs remains at the basis of the current success of GNNs.
A major drawback of GNNs with respect to alternative graph processing approaches~\cite{haussler1999convolution, patra2020review, shervashidze2011weisfeiler, grover2016node2vec} is the opacity of their predictive mechanism, which they share with most deep-learning based architectures. This severely limits the applicability of these technologies to safety-critical scenarios.

The need to provide insights into the decision process of the network, and the need to provide explanations for automatic decisions affecting human's life~\cite{Dai2022ACS, khosla2022privacy}, have stimulated research in techniques for shading light into the black box nature of deep architectures~\cite{ribeiro2016should, selvaraju2017grad, global_cnn, zeiler2014visualizing, sundararajan2017axiomatic, simonyan2013deep, springenberg2014striving, pope2019explainability,guidotti2018survey}. The approaches have also been adapted to generate explanations for GNN models~\cite{simonyan2013deep, springenberg2014striving, pope2019explainability, baldassarre2019explainability}. However, networked data have peculiarities that pose specific challenges that explainers developed for tensor data struggle to address. The main challenge comes from the lack of a regular structure, as nodes have a variable number of edges, which requires ad-hoc strategies to be properly addressed. Indeed, several approaches have been recently developed that are specifically tailored to explain GNN architectures, and Section~\ref{sec:explainers} will summarize the main contributions.

It is often the case, however, that each work proposes a new set of benchmarks or metrics, making the comparison across works complicated. We thereby stress the need for a comprehensive evaluation that can fairly benchmark the explainers under a unified lens.  One of the first attempts to provide such a comparative analysis is the up mentioned work by Yuan et al.~\cite{yuan2020explainability}, where a taxonomy of the available explainers was proposed. 
In addition to this, the authors reported a detailed overview of the most common datasets used to benchmark explainers, along with the adopted evaluation metrics. 

However, despite the wide coverage of explainers, datasets, and
evaluation metrics, only a single GNN architecture, namely a simple
Graph Convolutional Network~\cite{kipf2016semi}, was evaluated, so
that nothing can be said about the impact of different architectures
in the resulting explanations. A similar limitation affects the works
of Zhao et al.~\cite{zhao2022consistency} and Agarwal et
al.~\cite{agarwal2022evaluating,agarwal2022probing} that, despite
presenting interesting insights in terms of consistency of explainers,
desired properties of explanation metrics and even introducing a
generator for synthetic graph benchmarks, focus their analysis to a
single GNN architecture. Li et al.~\cite{li2022explainability}
conducted the first empirical study comparing different GNN
architectures. However, their study is limited to node classification
and the three explainers under analysis~\cite{ying2019gnnexplainer,
  luo2020parameterized, schlichtkrull2020interpreting} are not well
representative of the diversity of explanation strategies that have
been proposed, as summarized in the aforementioned
taxonomy~\cite{yuan2020explainability}. The most comprehensive study
to date is the recent work by Rathee at al.~\cite{rathee2022bagel},
that evaluated four GNN architectures over nine explainers for both
node and graph classification. However, the main goal of this study is
proposing a benchmarking suite to quantitatively evaluate explainers,
with no attempts at providing insights into why different GNN
architectures behave differently in terms of explainability, or which
explainer should be preferred in a given setting.

In spite of the aforementioned recent studies benchmarking explainability methods for GNNs, no investigation has been done in characterizing the typical explanation patterns associated to the topological concepts learned by the network and how different architectures affect the explanation.
In our work, we address these issues by answering the following research questions:

\begin{itemize}
\item \textbf{RQ1:} How does the architecture affect the explanations?
\item \textbf{RQ2:} How do explainers affect the explanations?
\item \textbf{RQ3:} How do different types of problems affect the explanations?
\end{itemize}

Overall, our work aims to go beyond a merely quantitative evaluation of the performance of explainer-GNN pairs and to make a significant step towards {\it explaining explainability}. We run an unprecedented number of experiments involving eight GNN architectures, twelve instance-based explainers, and six datasets (divided into node and graph classification) and enrich the quantitative results we obtain by providing a deep understanding of the reasons behind the observed behaviors, together with a set of recommendations on how to select and best use the most appropriate explainer for the task under investigation while avoiding common pitfalls, as well as a number of open problems in GNN explainability that we believe deserve further investigation.
\revDue{Following previous analyses~\cite{yuan2022explainability,li2022explainability,li2022survey,agarwal2022evaluating,rathee2022bagel}, our study investigates message-passing GNNs applied to static graphs. Exploring explainability in non-message passing architectures (e.g., Transformers) or different graph types (e.g., temporal graphs) is an interesting direction for future research, that may however necessitate distinct benchmarking approaches.}
% \revDue{Following previous analysis\cite{}, our work mainly focuses on massage passing GNNs on static graphs, as explanbility for non message passing architectures (i.e. Transformers) or type of graphs (i.e. temporal) may require different benchmarks. However, investigating expalanability in this settings constitutes a promising future development for future works.}
%Following previous analysis, our work encompasses mainly spatial GNNs for static graphs, as explainability for spectral and temporal graphs are currently understudied, therefore constituting a promising future development for our work.

%commento paper corto \antonio{commentato}
The remainder of the paper is structured as follows: Section \ref{sec:nets} presents an overview of GNN architectures, with greater detail on the models that we adopted in our study. On the same vein, Section \ref{sec:explainers} introduces the explainers for graph models, while Section \ref{sec:dataset} describes the benchmark datasets. Section \ref{sec:metrics} presents the evaluation metrics employed to asses the explanation's quality. Section \ref{sec:experimental_setting} summarizes how we trained the tested architectures, and Section \ref{sec:results} presents the results expressed with respect to the research questions defined above. In Section \ref{sec:discussion} we discuss the results. \revDue{Finally, in Section \ref{sec:future} we propose future research directions} and in Section \ref{sec:conclusion} we draw some conclusions.

%In doing so, we performed an extensive analysis of eight GNN architectures, ten instance-based explainers, and six datasets divided into node and graph classification. Instead of merely reporting the performances of each explainer for every architecture, we aim at providing general insights for the observed behaviours as well as pointing to possible future directions. 

%\section{Limitations of existing studies on GNN explainers}
%\label{sec:introandmotivation}

\section{Graph Neural Networks}\label{sec:nets}

In this section we first introduce the notation to deal with the GNN formalism, then we review the GNN architectures explicitly used in our study. 
\rev{To facilitate the reading of the paper, recurring mathematical notation is summarized in Table~\ref{tab:symbols}.}

\begin{table}
\begin{longtable}{p{5cm}p{9cm}}
\hline
Symbol & Description\\
\hline
$G:=(V, E, \bX)$  &  A graph.\\
$V:=\{1, \dots, n_V\}$  & The set of $n_V\in\N$ nodes of a graph.\\
$E\subset V\times V$  & The set of $n_E\in\N$ edges of a graph.\\
$\bX\in\R^{n_V\times d}, \bX_i\in\R^d$ & The matrix of $d$-dimensional node features, and the feature vector of the node $i\in V$.\\
$\bI\in\R^{n_V\times n_V}$& The identity matrix.\\
$\bA,\bAt\in\R^{n_V\times n_V}$& The adjacency and normalized adjacency matrix of a graph.\\
$\bD,\bDt\in\R^{n_V\times n_V}$& The degree and normalized degree matrix of a graph.\\
$\bL,\bLt\in\R^{n_V\times n_V}$& The Laplacian and normalized Laplacian matrix of a graph.\\
$N(i)\subset V$ & The first order neighborhood of the node $i\in V$.\\
$\bW\in\R^{d\times d'}$ & The trainable weights of a layer.\\
\hline
\end{longtable}
\caption{\rev{
List of the mathematical symbols used throughout the paper, and their meaning. A few symbols used only in specific settings are omitted and defined in the text.}}
\label{tab:symbols}
\end{table}
We consider a graph $G:=(V, E, \bX^V\rev{, \bX^E})$, with $n_V\in\N$ nodes $V:=\{1, \dots, {n_V}\}$, $n_E\in\N$ edges $E\subset V\times V$, a matrix of $d$-dimensional node features $\bX^V\in\R^{n_V\times d}$, where the $i$-th row of $\bX$ is the vector of $d\in\N$ features of the $i$-th node\rev{, and similarly a matrix of $d'$-dimensional edge features $\bX^E\in\R^{n_E\times d'}$. Most graphs considered in this paper do not have edge features, and we will simply write $G:=(V, E, \bX)$ in order to denote a graph with node features only.}
We use the matrices $\bA,\bL,\bI,\bAt, \bDt,\bLt\in\R^{n_V\times n_V}$, where $\bA$ and $\bL$ are the adjacency and Laplacian matrices of $G$, $\bI$ is the $n_V$-dimensional identity matrix, $\bAt:=\bA+\bI$, $\bDt$ is its diagonal matrix, and $\bLt:=\frac{2}{\lambda_{\max}(\bL)} \bL - \bI$ is the scaled and normalized Laplacian, where $\lambda_{\max}(\bL)$ is the largest eigenvalue of $\bL$. Furthermore, $N(i):=\{j\in V: (i,j)\in E\}$ is the first order neighborhood of the node $i\in V$.

Each GNN layer takes as input the graph $G$, and maps the node features $\bX\in\R^{n_V\times d}$ to updated node features $\bX'\in\R^{n_V\times d'}$ for a given $d'\in\N$. Some specific GNN layers, like Hierarchical Pooling layers~\cite{zhou2020graph, ying2018hierarchical, lee2019self, cangea2018towards}, instead of refining the embedding for each input node aggregate nodes in order to coarsen the graph in a similar way as done by pooling methods for vision models~\cite{cirecsan2011high, springenberg2014striving, krizhevsky2017imagenet}, thus resulting in a node feature matrix $\bX'\in\R^{n'\times d'}$ where $n' < n_V$. Overall, this new feature matrix $\bX'$ is the embedding or representation of the nodes after the application of one layer of the network.
When needed, we denote as $\bX_i, \bX_i'$ the original and transformed feature vector of the $i$-node, i.e., the transpose of the $i$-th row of the matrices $\bX,\bX'$.
Specifying the map $\bX\to \bX'$ is thus sufficient to provide a full definition of the different layers. 
These transformations are parametric, and they depend on trainable weights that are learned during the optimization of the network. We represent these weights as matrices $\bW$. Additional terms specific to single layers are defined in the following.

After an arbitrary number $t$ of GNN layers stacked in sequence, the node embedding matrix $\bX^{(t)}$ is further processed in a way that depends on the task to perform. In node classification settings~\cite{kipf2016semi, hu2020open}, where the aim is predicting one or more node properties, a Multi-Layer Perceptron (MLP)~\cite{Hastie2009} (with shared parameters across nodes) is applied to each node's embedding independently in order to output its predicted class. For graph classification settings~\cite{hu2020open} instead, where the goal is predicting a label for the entire graph, a permutation invariant aggregation function (like mean, max, or sum) is applied over nodes' embedding to compress $\bX^{(t)}$ into a single vector %$h\in\R^{d'}$ 
which is then mapped to the final prediction via a standard MLP.

\begin{figure}
     \centering
     \includegraphics[width=0.7\textwidth]{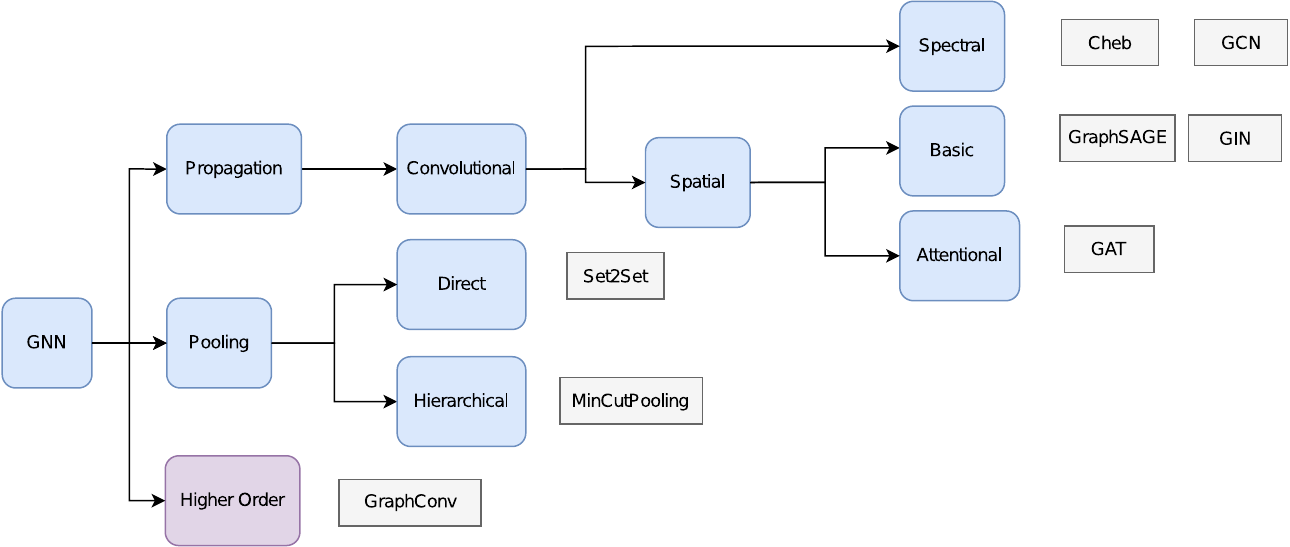}
     \caption{An overview of the adopted GNN architectures structured in a taxonomy as defined by Zhou et al.~\cite{zhou2020graph}. In particular, blue boxes represent the categories as defined by the aforementioned work, while the purple box corresponds to the newly introduced Higher Order category.}
     \label{fig:taxonomy}
 \end{figure}

With this notation settled, we can now fully define the architectures that we are going to consider. In selecting the architectures to be included in our study, we relied on the comprehensive taxonomy of GNN methods published by Zhou et al.~\cite{zhou2020graph}. Since our goal is to provide an extensive overview of explainability methods for GNNs, we selected the models to benchmark aiming at covering as much as possible the different categories of the taxonomy. The specific methods are also selected depending on their popularity, their ease of training, their performance on our benchmark datasets, their code availability \rev{and their compatibility with the explainers being investigated}. Overall, we analyzed the following categories: \emph{Convolutional} whose computation can be roughly intended as a generalization of the convolution operation on the image domain. Such convolution can either be \emph{Spectral}~\cite{defferrard2016convolutional, kipf2016semi}, theoretically grounded in graph signal processing~\cite{shuman2013emerging}, or \emph{Spatial}~\cite{velivckovic2017graph, hamilton2017inductive, xu2018powerful, gilmer2017neural}, where the operations are usually defined in terms of graph topology; The \emph{Pooling} category contains all approaches that aggregate node representations in order to perform graph-level tasks. They can be further differentiated into \emph{Direct}~\cite{vinyals2015order, zhang2018end}, where nodes can be aggregated with different aggregation strategies, often called readout functions, and \emph{Hierarchical}~\cite{ying2018hierarchical, cangea2018towards, Yuan2020StructPool:, lee2019self, bianchi2020spectral}, where nodes are progressively hierarchically aggregated based on their similarity. The latter methods often allow one to cluster nodes both based on their features and their topological neighborhood~\cite{ying2018hierarchical, bianchi2020spectral}. Despite covering the major aspects of GNN architectures, the aforementioned taxonomy lacks some of the fundamental works that we will analyze in our study. Particularly, to compensate that, we decided to respectively include the Graph Isomorphism Network (\gin)~\cite{xu2018powerful} and the GraphConv Higher Order Network (\graphconv)~\cite{morris2019weisfeiler} as \emph{Spatial Convolution} and \emph{Higher Order}, the latter being a new category added to the taxonomy. A summary of such categorization can be found in Figure \ref{fig:taxonomy}. In the following, we broadly describe each GNN model used in this work, reporting for each one the category as identified by Zhou et al.~\cite{zhou2020graph}. Although the following definitions are often given as global functions of the graph for notational convenience, we remark that all the layers of a GNN can be efficiently implemented as local operations by means of a message-passing or sparse matrix multiplications.

%In the following, we broadly describe each GNN model used in this work, reporting for each one the category as identified by Zhou \emph{et al.}~\cite{zhou2020graph}. Two GNN methods, namely \gin~\cite{xu2018powerful} and \graphconv~\cite{morris2019weisfeiler}, are assigned to the new category \newCategory, since they are not covered by the aforementioned taxonomy.

\begin{description}[leftmargin=0cm]
\item[Chebyshev Spectral Graph Convolution](\cheb, Spectral Convolution)~\cite{defferrard2016convolutional}:
%\item[\cheb](Spectral Convolution)~\cite{defferrard2016convolutional}:
%\textbf{Cheb} (Spectral Convolution)~\cite{defferrard2016convolutional}:
The Chebyshev spectral graph convolutional operator~\cite{defferrard2016convolutional} was aimed to generalize the convolution operation from the image to the graph domain. In doing so, it approximates the convolution of the node features with a trainable filter where such approximation is defined in the Fourier domain by means of a Chebyshev polynomial. Since explicitly computing the convolution in the Fourier domain is very computationally expensive, \cheb adopts the truncated recursive Chebyshev expansion~\cite{hammond2011wavelets}, where the Chebyshev polynomial $T_k(x)$ of order $k$ can be computed by the recurrence $T_k(x) = 2xT_{k-1}(x) - T_{k-2}(x)$ with $T_1 = 1$, $T_2 = x$. Given a degree $K\in\N$, we thus have:
% This convolution approximate the spectrum of a graph with a Chebyshev polynomial. The convolution is defined as follow:
\begin{equation}
    \bX' =  \sum_{k=0}^{K} \bZ^{(k)} \cdot \bW^{(k)}, 
\end{equation}
where $\bZ^{(k)}$ is computed recursively as:
\begin{align*}
    \bZ^{(1)} = \bX,\quad\bZ^{(2)} = \bLt\cdot \bX,\quad \bZ^{(k)} = 2 \bLt\cdot \bZ^{(k-1)} - \bZ^{(k-2)}.
\end{align*}
In our analysis we kept $K = 5$.

\item[Graph Convolutional Network](\gcn, Spectral Convolution)~\cite{kipf2016semi}: \gcn~\cite{kipf2016semi} represents a first-order approximation of localized spectral filters on graphs~\cite{defferrard2016convolutional}.
For a single layer, the node representation is computed as:
\begin{equation}
\label{eq:GCN}
\bX' = \sigma (\bDt^{-1/2}\cdot \bAt \cdot \bDt^{-1/2} \cdot \bX \cdot \bW),
\end{equation}
with $\bW\in \R^{d\times d'}$, and where $\sigma:\R\to\R$ is a nonlinear activation function that is applied entry-wise. The normalization applied to $\bAt$ is in place to avoid numerical instabilities after successive applications of the above propagation rule. It effectively normalizes the node-wise aggregation by a weighted sum where each neighbor is weighted by the incident edge weight normalized by the degree of the two nodes.
The connection with \cheb introduced above can be seen by keeping the number of convolutions per layer to 1, thus by setting $K = 1$. After some further simplifications detailed in \cite{kipf2016semi}, and here omitted for brevity, we can rewrite the spectral convolution as $\bW^{(k)} \cdot (\bI + \bDt^{-1/2}\cdot \bAt \cdot \bDt^{-1/2}) \cdot x$. After applying the renormalization trick introduced in~\cite{kipf2016semi} and by generalizing the formulation to a node feature matrix $\bX\in\R^{n_V\times d}$ we get the formulation of Eq \ref{eq:GCN}.

\item[Graph SAmple and aggreGatE](\sage, Spatial Convolution)~\cite{hamilton2017inductive}: This work was proposed as an extension of \gcn. Contrary to \gcn, which inherently implements a weighted mean neighborhood aggregation, GraphSAGE generalizes to different kinds of aggregations like mean, max or Long Short-Term Memory (LSTM) aggregations~\cite{hamilton2017inductive, LSTM}.
\begin{equation}
    \bX'_i = \bW_1 \bX_i + \bW_2 \cdot aggregation_{j\in N(i)} \bX_j,
\end{equation}
where $\bW_1,\bW_2$ have both dimension $n_V\times d'$ and the function aggregation can be any permutation invariant function. Note that in the case of the LSTM aggregation, the authors adapted LSTMs to operate on sets by simply applying the LSTMs to a random permutation of the nodes. In our study we used the mean of node features as aggregator.

\item[Graph Isomorphism Network](\gin, Spatial Convolution)~\cite{xu2018powerful}: The work of Xu et al.~\cite{xu2018powerful} was among the first ones to study the expressive power of GNNs in relation to the Weisfeiler-Lehman (WL) test of isomorphism~\cite{Leman2018THERO}. The key insight is that a GNN can have as large discriminative power as the WL test if the GNN’s aggregation scheme is highly expressive and can model injective functions, like the aggregation of the WL test. They studied the conditions for which a GNN has the same discriminative power as the WL test, and then they proposed the Graph Isomorphism Network (\gin) that provably satisfies such conditions~\cite{xu2018powerful}. The \gin computes node representations as:

\begin{equation}
    \bX' = h_{\bW'} (( \bA + (1+\epsilon) + \bI) \cdot \bX),
\end{equation}
where $\epsilon\in\R$ is a small number, and $h_{\bW'}$ is a neural network whose weights $\bW'$ are the trainable part of the \gin layer. This model generalizes the WL test and hence achieves maximum discriminative power with respect to the WL test~\cite{xu2018powerful}. In our work, to limit the complexity of the network, we limited $h_{\bW'}$ to have a single layer.

\item[Graph Attention Network](\gat, Spatial Attentional Convolution)~\cite{velivckovic2017graph}: Following the successes of attention mechanism in natural language processing~\cite{VaswaniAttention, raffel2020exploring, roccabruna-2022-multi} and in computer vision~\cite{xu2015show, ramesh2022hierarchical}, Veli{\v c}kovi{\'c} et al.~\cite{velivckovic2017graph} proposed a GNN layer which computes each node's representation as a weighted sum of neighborhood features, where the weights are computed via an attention mechanism. Such attention mechanism helps the layer to focus on neighbors which are considered to be important for the current node, instead of treating each neighbor equally importantly~\cite{bahdanau2014neural}. Specifically, the propagation rule for each node can be defined by:
\begin{equation}
    % \bX'_i = \alpha_{i,i} \bW\cdot \bX_i + \sum_{j\in N(i)} \alpha_{i,j} \bW\cdot \bX_j,
    \bX'_i = \sum_{j\in N(i)\cup\{i\}} \alpha_{i,j} \bW\cdot \bX_j,
\end{equation}
where the $\alpha_{i,j}$ are attention coefficients, which are the output of a single-layer feed forward neural network applied to $\bW\cdot \bX_i, \bW\cdot \bX_j$. Namely, if $[x||y]$ denotes the concatenation of the vectors $x,y$, and if $\bW'\in\R^{2d'\times 1}$ is a vector of trainable weights, then $\alpha_{i,j}$ is defined by
\begin{equation}
    \alpha_{i,j} := \dfrac{\exp(\lrelu((\bW')^T [\bW \bX_i || \bW \bX_j]))}{\sum_{k \in N(i) \cup  \{i\}}  \exp(\lrelu((\bW')^T [\bW \bX_i || \bW \bX_k]))},
\end{equation}
where $\lrelu$ has usually a slope $\alpha:=0.2$ if $x\leq 0$. Similarly as previous works on attention~\cite{VaswaniAttention, clark2020electra, raffel2020exploring, roccabruna-2022-multi, ramesh2022hierarchical}, \gat can implement multi-head attention in order to increase the attention's expressive power in modelling different aspects of node features.

%\item[Diffpool]\cite{ying2018hierarchical}: This network is used in graph classification. It differs from the other cases since each application of the layer groups the nodes into clusters, and thus the number $n_V'\in\N$ of output nodes differs from the number $n_V$ of input nodes. In particular, we have that $\bX'\in R^{n_V'\times d'}$. The method is based on aggregation and coarsening, and thus usually $n_V'<n_V$.
%The map from the nodes to the clusters is obtained by training a differentiable assignment matrix $\bS\in\R^{n_V \times n_V'}$, where the $i$-th entry of the $j$-th column represents the weight of the soft asignement of the $i$-node to the $j$-th cluster.

%The new node representation and new adjacency matrix $\bA'\in\R^{n_V'\times n_V'}$ are then defined by 
%\begin{align}
 %   \bX' &= \sftmax(\bS)^T \cdot \bX,\\
  %  \bA' &= \sftmax(\bS)^T \cdot \bA \cdot \sftmax(\bS)\nonumber.
%\end{align}
%Repeated applications of the layer with $n_V'=1$ in the final layer, provide an embedding of the entire graph.

\item[MinCutPooling](\mincut, Hierarchical Pooling)~\cite{bianchi2020spectral}: Similarly to other graph pooling mechanisms~\cite{ying2018hierarchical, cangea2018towards, Yuan2020StructPool:, lee2019self}, \mincut is able to hierarchically aggregate nodes into coarser representations to summarize local components and remove redundant information. Graph pooling follows a similar idea as pooling in standard architectures for vision applications~\cite{cirecsan2011high, springenberg2014striving, krizhevsky2017imagenet}, where it helps to reduce the memory and computation footprint of the model. However, the structured graph domain poses new challenges which required ad-hoc techniques for achieving good performing pooling methods. \mincut~\cite{bianchi2020spectral} achieves so by formulating a relaxation of the MinCut problem and by training a GNN to compute cluster assignment by jointly optimizing the downstream supervised loss and the additional unsupervised \emph{mincut} loss. In particular, a soft cluster assignment matrix $\bS\in\R^{n_V \times n_V'}$ is computed with a MLP with softmax activation over a refined set of node features (e.g. after one or more layers of message passing). Then, both the adjacency matrix and the node features' matrix are updated accordingly:
\begin{align}
    \bX' = \bS^T \cdot \bX,\quad
    \mathbf{\tilde{A}} &= \bS^T \cdot \bA \cdot \bS -\boldsymbol{\mathit{I}_{n_V'}}\diag(\bS^T \cdot \bA \cdot \bS),\quad
    \bA' = \tilde{D}^{\frac{1}{2}} \cdot \mathbf{\tilde{A}} \cdot \tilde{D}^{\frac{1}{2}}\nonumber.
\end{align}

We refer the interested reader to \cite{bianchi2020spectral} for the details about the mathematical formulation of the differentiable MinCut problem. Since pooling methods modify the graph structure, they are not typically suitable for node classification and link predictions tasks~\cite{ying2018hierarchical, cangea2018towards, Yuan2020StructPool:, lee2019self}. We will henceforth refer to as \mincut for an architecture implemented with \gcn layers interleaved with the \mincut operator.

\item[\setset](Direct Pooling)~\cite{vinyals2015order}: The \setset model~\cite{vinyals2015order} is a specific approach for global graph pooling which takes as input the node-level representations, as computed by any GNN layer, and outputs a single representation for the entire graph suitable for graph-level tasks (e.g., graph classification~\cite{hu2020open}, molecular property prediction~\cite{MolData, borgwardt2005protein}, etc.). Whilst traditional approaches simply taking the mean/max/sum of every node's representations, \setset can be seen as learning the aggregation via an LSTM. Thus, the node features are treated as sequences, which are processed through an LSTM network to obtain an embedding of the entire graph, represented as a vector of length $2d$. To enforce permutation equivariance, the ordering of the nodes can be either learned or randomized during training.
The underlying LSTM works as follows:
\begin{align}
    \textbf{q}_t &= \text{LSTM}(\textbf{q}^{*}_{t-1}),\quad
    \alpha_{i,t} = \sftmax(\bX_i \cdot \textbf{q}_t)\nonumber,\\
    \textbf{r}_t &= \sum_{i = 1}^{n_V} \alpha_{i,t}\bX_i\nonumber,\quad
    \textbf{q}_{t}^{*} = \left[\textbf{q}_t || \textbf{r}_t\right]\nonumber.
\end{align}
where $\textbf{q}_{t}^{*}$ is the output of the layer. The subscript $t\in\{1,\dots, T\}$ indicates that the process may be repeated a number $T\in\N$ of times. In our work we refer to \setset as a GNN architecture implemented as a number of \gcn layers with the \setset global pooling operator and with a value of $T = 7$.

\item[\graphconv](Higher Order)~\cite{morris2019weisfeiler}: To increase the expressivity of GNNs, researchers developed techniques to capture not only 1-hop connections, i.e., between the node neighborhood, but also to capture higher order connections~\cite{morris2019weisfeiler,li2021higher,duval2022higher,roddenberry2019hodgenet,zhang2019hyper,li2022improving,gao2021higher}. 
%Many works have been proposed~\cite{morris2019weisfeiler,li2021higher,duval2022higher,roddenberry2019hodgenet,zhang2019hyper,li2022improving,gao2021higher}, however, in this work we focused on the one proposed by \cite{morris2019weisfeiler} \graphconv.
This network captures higher order connections of a graph (up to order two) by aggregating information from neighbourhood nodes and incident edges. It is defined as:
\begin{equation}
    \bX'_i = \bW_1 \bX_i + \bW_2 \sum_{j \in N(i)} e_{i,j} \cdot \bX_j,
\end{equation}
where $\bW_1$ and $\bW_2$ are learnable weight matrices, while $e_{i,j}$ is the edge weight from node $i$ to node $j$. 

\end{description}

%In this work we explain those networks, covering the major categories of GNN networks developed so far on both graph and node classification. Recall that pooling networks cannot used in node classification tasks~\cite{ying2018hierarchical, cangea2018towards, Yuan2020StructPool:, lee2019self}. 
%The aim of this work is to capture the intrinsic explanation exploited by a specific architecture, thus networks that does not achieve a reasonable accuracy are not carried on in the analysis.

\section{GNN Explainability}\label{sec:explainers}

To analyze and understand the strengths and weakness of graph explanation algorithms, we selected instances of GNN explainers which are representative of the current state of the art.
To this end, we follow the systematization proposed by Yuan et al.~\cite{yuan2020explainability}, and choose to investigate \emph{instance-based} explainers \cite{zhou2016learning,selvaraju2017grad,simonyan2013deep,sundararajan2017axiomatic,springenberg2014striving,vu2020pgm,ying2019gnnexplainer,baldassarre2019explainability,schwarzenberg2019layerwise,huang2022graphlime,ribeiro2016should, yuan2021explainability, gcexplainer, NEURIPS2021_be26abe7, pmlr-v151-lucic22a, zhang2021relex, schnake2021higher, funke2020hard, li2023dag, lin2021generative}, i.e., those which aim at identifying components of the input that are responsible for the model's output. This is in contrast with \emph{model-based} explainers, which rather try to provide a global understanding of a trained model\rev{~\cite{GLGExplainer, yuan2020xgnn, DAGExplainer, GLocal, wang2022gnninterpreter}}. Since the available model-based explainers are very heterogeneous (i.e., it is not available a unified evaluation setting), and since previous works on benchmarking graph explainers have focused on instance-based methods~\cite{yuan2020explainability,zhao2022consistency,agarwal2022evaluating,agarwal2022probing,li2022explainability,rathee2022bagel}, we thereby omit model-based explainers.
In particular, Yuan et al. \cite{yuan2020explainability} identifies four macro categories of instance-based explainers, namely \textit{gradient}-, \textit{perturbation}-, \textit{decomposition}- and \textit{surrogate-based} models. Roughly speaking, gradient-based explainers exploit gradients of the input neural network~\cite{springenberg2014striving, sundararajan2017axiomatic, pope2019explainability}, perturbation-based models perturb the input aiming to obtain explainable subgraphs \cite{ying2019gnnexplainer, luo2020parameterized, funke2020hard, schlichtkrull2020interpreting}, decomposition-based models try to decompose the input identifying the explanations~\cite{baldassarre2019explainability, pope2019explainability, schnake2021higher}, while surrogate-based models use a simple interpretable surrogate to explain the original neural network~\cite{huang2022graphlime, zhang2021relex, vu2020pgm}. \rev{Furthermore, in order to account for a number of new approaches based on modeling the underlying graph distribution via a generative process, and following the categorization proposed by Kakkad~\cite{kakkad2023survey}, a fifth category named \textit{generation-based} is added~\cite{NEURIPS2021_be26abe7, li2023dag, lin2021generative}}.

\subsection{Explanation masks}\label{sec:masks}
Independently from this categorization, a further fundamental distinction is among explainers providing explanations in terms of edge~\cite{ying2019gnnexplainer, luo2020parameterized, schlichtkrull2020interpreting, yuan2021explainability} or node masks~\cite{springenberg2014striving, sundararajan2017axiomatic, baldassarre2019explainability, pope2019explainability, schnake2021higher}. 
\rev{Given a graph $G:=(V, E, \bX^V, \bX^E)$ (with possibly empty node or edge feature matrices - see Section~\ref{sec:nets}), a node explanation mask is a graph $G_{\exp}:=(V, E, \bX^V_{\exp})$ where the node features $\bX^V_{\exp}\in\R^{n_V\times 1}$ are node explanation weights. 
Similarly, an edge explanation mask is a graph $G_{\exp}:=(V, E, \bX^E_{\exp})$, where now $\bX^E_{\exp}\in\R^{n_E\times 1}$ are edge explanation weights.
For both nodes and edges, we have hard masks if the weights have binary values in $\{0,1\}$, and soft masks if they have continuous values in $[0,1]$. Any soft mask $G_{\exp}$ can be converted in an hard mask $G_{\exp}(t)$ by thresholding its weights with a given threshold value $t\in(0,1)$.
Given an hard mask $G_{\exp}$,  its complement $G\setminus G_{\exp}$ is another hard mask where the value of binary weights is flipped.
}

% \rev{We recall that, for a graph $G:=(V,E,\bX)$, a node mask $M_V:V\to[0,1]$ and an edge mask $M_E:E\to[0,1]$ are maps that assign a weight to each node or edge in $G$. Up to normalization, we assume that both type of masks have values in $[0,1]$: this facilitate the comparison of different masks, and it does not modify the explanation since it only relies on the relative importance of nodes and edges.
% }
% We refer to edge mask every time the explainer gives as output any sort of likelihood for each edge in the input graph. Conversely, a node mask represents likelihoods for every node in the graph. 
% commento paper corto 
%Despite this distinction, it is anyway possible to transform an edge mask into a node mask with a loss of information (and vice-versa) by, for example, averaging for each node the likelihoods of the edges incident to that specific node. 
\rev{In addition to these two types of masks, a few explainers return also an explanation for the node features}~\cite{ying2019gnnexplainer, simonyan2013deep}. However since single node features are not representative of the underlying topological structure which we are interested in, and in line with most previous works~\cite{yuan2020explainability,zhao2022consistency,agarwal2022evaluating,agarwal2022probing,li2022explainability,rathee2022bagel}, we do not consider single node features' explanations.
% commento paper corto 
%Furthermore, to maintain a unique and homogeneous evaluation setting, we convert every edge mask into a node mask, accordingly to the procedure described before.

%relying on edges, and thus provide an explanation which is an edge mask, or on nodes, thus providing a node mask. In both cases, all the explainers that we consider produce a soft explanation, i.e., the masks are defined by assigning real valued weights (or scores) to the nodes or to the edges, rather than binary values.

\subsection{Selection of the explainers}
Below we report a brief overview of our benchmark explainers. 
%Note that in this study we did not aim to cover every explainer for GNNs, but we rather focused on a subset. 
Despite the existence of other works proposing explainers, which occasionally fall outside the aforementioned categorization \cite{yuan2021explainability, gcexplainer, pmlr-v151-lucic22a, huang2022graphlime, zhang2021relex, schnake2021higher}, we limited our analysis on a subset. More specifically, the criteria for selecting a given explainer can be roughly summarized by \emph{i)} representativity of a specific category as outlined before; \emph{ii)} code availability; and \emph{iii)} feasibility of usage, i.e., whether the explainer is not too computationally heavy to be used. %Unfortunately, for the \emph{decomposition} category, we did not find an explainer with a working codebase that could be promptly included in our study. 

Given a GNN $g$ to be explained, let $g(e)^c = y^c = \left(w^c\right)^T e$ be the prediction of the model where $e$ corresponds to the final graph-level or node-level embedding, 
% corresponds to \emph{i)} a global pooling of node features at the last layer of the GNN for graph classification settings ($e := pooling_{i\in V}(\bX^{(t)}_i)$); \emph{ii)} the single refined node feature for node classification settings ($e := \bX^{(t)}_i$). 
and where the vector $w^c\in\R^{d'}$ contains instead the learned Fully Connected weights for class $c$ to perform the final classification.
\rev{Furthermore, we denote with $\letterExpl^c(i)$ the importance attributed to a given explainer to the node $i\in V$ for the prediction of class $c$, i.e., the collection of the values $\letterExpl^c(i)$ for $i=1, \dots, n_V$ provides the node explanation mask $\bX_{\exp}^V$ for class $c$.}
A brief summary of the methods is available in Table~\ref{tab:explainers}.

\begin{description}[leftmargin=0cm]
\item[\gradexp] (Gradient)~\cite{simonyan2013deep}: Inspired by the previous work computing explanation in the context of Bayesian classification~\cite{baehrens2010explain}, \gradexp computes the attribution for each input by backpropagation to the input space. The general idea is that the magnitude of the derivative gives insights into the most influential features that, if perturbed, give the highest difference in the output space:
\begin{equation}
    \letterExpl_{\gradexp}^c\rev{(i)} =  \frac{\partial y^c}{\partial \bX_{\rev{i}}}
\end{equation}

To derive a single value for each input node, it is possible to take the maximum magnitude along the feature channel. Similarly, if the GNN to explain supports edge weights, then this method can be applied on edges, thus producing an edge mask. In the rest of the paper we will refer to this method with \gradexp and \sal, whether it is applied to nodes or edges, respectively.
%It produces explanations using the squared norm of its gradient with respect to the inputs. Note that this explainer can be used for both node and edge importance. Of course, in order to be used for edge masks, the input Neural Network have to accept weighted edges. To make the further analysis easier, we will distinguish node explainer with \textbf{Grad explainer (node)} and edge explainer with \textbf{Grad explainer (edge)}.

\item[\guidedbp:] (Gradient)~\cite{springenberg2014striving} Guided Back Propagation (\guidedbp) follows a similar approach as \gradexp, with the only difference that in the backpropagation it clips negative gradients which corresponds to features prone to decreasing the target activation, and thus considered as noise.

\item[\ig:] (Gradient)~\cite{sundararajan2017axiomatic} \ig
builds on top of simple Gradient-based methods, like \gradexp, by integrating the gradient along a path. Specifically, given $x' \in \R^d$ a baseline input which represents a neutral input, often represented by a zero-vector, the resulting explanation is computed as:
\begin{equation}
\label{eq:ig}
    \letterExpl_{\ig}^c\rev{(i)} = (\bX_{\rev{i}} - x')  \int_{0}^{1} \frac{\partial f(x' + \alpha(\bX_{\rev{i}}-x'))}{\partial \bX_{\rev{i}}} \,d\alpha 
\end{equation}

In short, the explanation corresponds to the integral computation of the gradients along the straight line path from an input baseline to the original value of the input. The adoption of this integral computation was shown to provide better theoretical guarantees then other approaches, as demonstrated in~\cite{sundararajan2017axiomatic}. Nonetheless, for an efficient implementation, the integral in Eq \ref{eq:ig} is substituted by a finite summation. This explainer can be applied even on edges if the GNN support edge weights. In the rest of the paper, we will refer to \ignode and \igedge, whether it is applied to nodes or edges, respectively.

%This explainer can be used for both node and edge masks. Again, we will distinguish between \textbf{Integrated gradient (node)} and \textbf{Integrated gradient (edge)}.

\item[\cam] (\rev{Decomposition})~\cite{pope2019explainability}: Class Activation Map (\cam) maps backward the node features in the final layer to the input space for identifying important nodes. It represents a straightforward adaptation of \cam originally developed for Convolutional Neural Networks~\cite{zhou2016learning} to the graph domain. The \cam heat-map for node $\rev{i}$ and class $c$ is defined as:
\begin{equation}
    \letterExpl_{\cam}^c\rev{(i)} = \relu\left(\left(w^c\right)^T \bX^{(t)}_{\rev{i}}\right)
\end{equation}

\item[\gradcam] (Gradient)~\cite{pope2019explainability}: \gradcam represents an extension of \cam. Instead of using the weights $w^c$ to weight the contribution of each feature, \gradcam uses the gradient of the output with respect to node features.
\begin{equation}
    \letterExpl_{\gradcam}^c\rev{(i)} = \relu\left(\left(\alpha^c\right)^T \bX^{(t)}_{\rev{i}}\right)
\end{equation}
%\begin{equation}
%\label{eq:grad_cam_alpha}
%    \alpha_k^c = \frac{1}{\rev{n_V}} \sum_{\rev{i}=1}^{\rev{n_V}} \frac{\partial y^c}{\partial \bX^{(t)}_{\rev{i},k}}
%\end{equation}
\begin{equation}
\label{eq:grad_cam_alpha}
    \alpha_k^c = \frac{\partial y^c}{\partial \bX^{(t)}_{\rev{i},k}}
\end{equation}
where $\alpha^c = [\alpha^c_0, \dots ,\alpha^c_{d'}]^T$. Note \gradcam allows also to compute the heat-map for each layer independently, by simply replacing the node feature matrix in the above equations with the features of any specific layer~\cite{selvaraju2017grad}.
% commento paper corto 
% \gnnexpl was the first algorithm specifically tailored for GNNs.
\item[GNNExplainer] (Perturbation)~\cite{ying2019gnnexplainer}: \gnnexpl is able to provide an explanation both in terms of a subgraph of the input instance to explain, and a feature mask indicating the subset of input node features which is most responsible for the GNN's prediction.  Overall, it is formulated as an optimization problem maximizing the Mutual Information between the GNN's predictions and distribution of possible subgraphs. However, in practical terms, since the optimization problem formulated in this way is intractable given the exponential number of subgraphs for a specific input graph, a relaxed version is actually computed, which can be interpreted as a variational approximation of the distribution of subgraphs. Nonetheless, despite the non-convex nature of the problem, the authors  empirically observed that the aforementioned approximation together with a regularizer for promoting discreteness converges to good local minima.

\item[\pgexpl:] (Perturbation)~\cite{luo2020parameterized}: The Parametrized Explainer for GNNs (\pgexpl)~\cite{luo2020parameterized} adopts a very similar formulation of the explanation problem as \gnnexpl where the two major differences are: \emph{i)} \pgexpl provides solely explanations in terms of subgraph structures, neglecting explanations in terms of node features; \emph{ii)} instead of directly optimizing  continuous edge and features masks as done by \gnnexpl, it uses Gradient Descent to train a MLP which, given the two concatenated node embeddings $[X_i^{(t)}||X_j^{(t)}]$, predicts the likelihood of the edge ($i$, $j$) being a relevant edge.

\rev{
\item[\subgraphx] (Perturbation)~\cite{yuan2021explainability}: SubgraphX uses the Monte Carlo tree search algorithm to explore and score different connected subgraphs of the given
input graph. It can be formally described as:

\begin{equation}
    G_{\exp} = \argmax_{|G_i| \le N_{\min}} \score(g, G, G_i)
\end{equation}

where $N_{\min}$ is the maximum size of the expected explanation, each candidate explanation $G_i$ is restricted to be connected to avoid disconnected explanatory graphs, and the scoring function is obtained by adapting Shapley values \cite{shapley} to the graph domain.}

\item[\pgmexpl] (Surrogate)~\cite{vu2020pgm}: \pgmexpl builds a surrogate model of the GNN to be explained by building a probabilistic graphical model. Random node features perturbations are applied to the given instance and, for each perturbation, the algorithm records the influence of the perturbation to the final prediction. After a number of perturbations, a dataset is generated and used to learn an interpretable Bayesian network which serves as final explanation~\cite{vu2020pgm}.

\rev{
\item[\rgexpl] (Generation)~\cite{NEURIPS2021_be26abe7}: RG-Explainer is a reinforcement learning enhanced explainer based on three main components: starting point selection, iterative graph generation, and stopping criteria learning. Those components are all trained via policy gradient using as the (negative) reward a similar loss as \pgexpl and \gnnexpl~\cite{luo2020parameterized, ying2019gnnexplainer} with the addition of regularization terms to encourage the explanation to be small in the size of the nodes, to have a low longest length of the shortest path, and have similarity between the original node representations and the new representation~\cite{NEURIPS2021_be26abe7}. In order to learn the stopping criteria, a special action STOP is added to the set of possible actions and a maximum number of generation steps is set to avoid generating very large subgraphs.}

%\item[LPR NON VA]
%\item[GraphLime:] solo per node features
\end{description}

\begin{table}
\centering
\begin{tabular}{ccccc}
\hline
Name & Category & Task & Mask type \\ %& Node feature mask
\hline
  \gradexp & Gradient & Graph/Node & Node\\ % & \cmark, Edge
  \sal & Gradient & Graph/Node & Edge\\  %& \cmark
  \guidedbp & Gradient & Graph/Node & Node\\ %\cmark
  \igedge & Gradient & Graph/Node & Edge \\%& \cmark Edge
  \ignode & Gradient & Graph/Node & Node \\ %& \cmark Edge
  \gradcam & Gradient & Graph/Node & Node \\ %& \cmark
  \gnnexpl & Perturbation & Graph/Node & Edge \\ %& \cmark
  \pgexpl & Perturbation & Graph/Node & Edge  \\ %& \xmark
  \rev{\subgraphx} & Perturbation & Graph/Node & \rev{Edge}  \\
  \pgmexpl & Surrogate & Graph/Node & Node  \\ %& \xmark
  \cam & \rev{Decomposition} & Graph/Node & Node \\%& \cmark
  \rev{\rgexpl} & Generation & Graph/Node & \rev{Edge}  \\
  % \gradexp & Gradient & Graph/Node & Node, Node features, Edge\\
  % \guidedbp & Gradient & Graph/Node & Node, Node features\\
  % \ig & Gradient & Graph/Node & Node, Node features, Edge\\
  % \cam & Gradient & Graph/Node & Node, Node features\\
  % \gradcam & Gradient & Graph/Node & Node, Node features\\
  % \gnnexpl & Perturbation & Graph/Node & Edge, Node features\\
  % \pgexpl & Perturbation & Graph/Node & Edge\\
  % \pgmexpl & Surrogate & Graph/Node & Node\\
\hline
\end{tabular}
\caption{Summary of explainers analyzed in this work. The columns \emph{Task} represents to which downstream task the explainer can be applied to, while \emph{Mask type} represents whether the explainer returns explanations in terms of entire node importance, single node features importance, or edge importance.}
\label{tab:explainers}
\end{table}
% \todo{A:Chiedi a @Gabri}

\section{Benchmark datasets}\label{sec:dataset}
In this section we present the graph benchmark datasets employed in our work, the majority of which represent newly proposed datasets. In designing the new benchmarks, we took inspiration from Faber et al.~\cite{faber2021comparing} who analyzed frequent biases in evaluating GNN explainers and pointed out that explainers should be evaluated on controlled benchmarks where the ground-truth evidence for target labels is known, and that different benchmarks should aim at testing different aspects of the GNN. In the following we describe in detail each dataset, both for node and graph classification.

\subsection{Datasets for graph classification}\label{sec:dataset_gc}
% \gabriele{Qui bisogna controllare di aver spiegato bene cosa sono le ground truth}

\begin{description}[leftmargin=0cm]

\item[\datasetgcone:] Inspired by the benchmarks presented in Ying et al. \cite{ying2019gnnexplainer}, the \datasetgcone dataset is composed by 1000 Barab{\'a}si-Albert (BA) graphs~\cite{barabasi1999emergence}. To half of these 1000 graphs we attach a $3 \times 3$ grid, and the resulting graphs are assigned to the positive class, while the ones without grid are the negative class.
The number of nodes in the BA graph is a uniformly distributed random number between 15 and 30 (for the negative class) and between 6 and 21 (for the positive class). This guarantees that when adding the grid, the average number of nodes in the positive class matches the one in the negative class. It is worth mentioning that in the experiments done by Ying et al. \cite{ying2019gnnexplainer}, the total number of nodes is fixed.
This benchmark evaluates the ability of the explainers to identify explanations consisting of a simple connected pattern.

\item[\datasetgctwo:] \datasetgctwo is characterized by two concepts. A $3 \times 3$ grid, as in the previous benchmark, and a house made of 5 nodes. The base structure, to which the concepts are attached, is a BA graph with a random number of nodes, and the final task corresponds to binary classification. The negative class consists of a BA graph connected to a grid or a house, while the positive class is composed by a BA graph connected to {\em both} a grid and a house. This benchmark aims at evaluating compositionality, as identifying simple patterns in isolation is insufficient to characterize the ground truth.

\item[\datasetgcthree:] The \datasetgcthree benchmark is characterized by a random graph connected to a variable number of star-shaped structures (from one to four). For the random graph generation, this time, we opted for the Erd\H{o}s-R{\'e}nyi (ER) random graph model \cite{erdHos1959random} to avoid a possible interference of stars generated in a BA graph. We defined a three-class classification task, depending on the number of stars present in each sample: class 0 corresponds to 1 star, class 1 to 2 stars, and class 2 to 3 or 4 stars. Each star has a fixed size of 16 nodes, and the total number of nodes is uniformly distributed between 30 and 50. This benchmarks is aimed at evaluating how explainers deal with counting substructures.
% commento paper corto \antonio{commentato}
% Note that standard message-passing GNNs can learn to count star-like motifs, while more densely connected substructures are out of reach~\cite{chen2020}. 

\item[\datasetgcfour:] None of the previous benchmarks involves node features. Thus, in this benchmark we test how node features affect explanations. In particular, we have a random BA graph with (one-hot encoded) random colored nodes (blue, green, red). To the base BA graphs, we attached from one to three house-like structures made of five nodes. One of these houses has a uniform color, which is blue for the negative class and green for the positive one. The other houses have random colors.
\end{description}

% \begin{figure}
%     \centering
%     \includegraphics[width=\textwidth]{Figures/draw/GC_datasets.pdf}
%     \caption{Caption}
%     \label{fig:my_label}
% \end{figure}

\subsection{Datasets for node classification}\label{sec:dataset_nc}
\begin{description}[leftmargin=0cm]
\item[\datasetncone:] The BA-Shapes dataset (henceforth referred to as  \datasetncone), introduced in Ying {et al.}~\cite{ying2019gnnexplainer}, is a widely used dataset for benchmarking GNN explainers~\cite{ying2019gnnexplainer, luo2020parameterized, NEURIPS2021_be26abe7, yuan2020explainability, yuan2021explainability, zhao2022consistency}. It is composed by 300 nodes and a set of 80 five-node house-structured network motifs which are randomly attached to the base graph, generated following the BA model~\cite{barabasi1999emergence}. 
Nodes are assigned to four categories, namely they either do not belong to a house (class 0), or they are classified depending on their structural function in the house: they may be either on the middle below the roof (class 1), on the base (class 2), or on the top of the roof (class 3).
The expected ground truth explanation is a house motif for all classes.

\item[\datasetnctwo:] This benchmark graph has been introduced in Section 5.1 of Faber et al. \cite{faber2021comparing}, and we use it with minor modifications\footnote{The code to generate this benchmark can be found at \url{https://github.com/m30m/gnn-explainability}.}.
Starting from a directed ER graph \cite{erdHos1959random} with $1000$ nodes and an edge-generation probability $p=0.004$, a set of $50$ nodes is selected uniformly at random and identified as infected. The state of each node is mapped to a node feature by one-hot encoding, i.e., a node has feature $[0, 1]$ if it is healthy and $[1, 0]$ if it is infected. 
The label of a node is defined based on the length of a minimal directed path to reach this node from an infected one. Namely, if this distance is denoted as $d$, then the node has label $0$ if $d=0$ (i.e., the node itself is infected), label $1$ if $d=1$ or $d=2$, and label $2$ otherwise, i.e., $d\geq 3$. We remark that the original dataset included $5$ classes ($d=0, 1, 2, 3$, and $d\geq 4$), but we restrict to three to simplify the presentation of the results. Accordingly, we employ two-layer networks instead of the four-layer ones used in \cite{faber2021comparing}.
% The dataset is designed to test the capacity of the classifiers to identify specific patterns in the data (a path of minimal length in this case).
From the definition of the dataset we identify the expected ground truth explanations of a node $v$ as follows. For label zero, the explanation is the node itself. For label 1, any directed path of length one or two from an infected node to $v$ is a valid explanation. For label 2, the explanation is given by the union of all directed paths of length up to 2 from any node in the graph to $v$. In the last case, indeed, the network has to check the entire set of nodes from which $v$ is reachable in at most two steps to exclude that any of them is infected.
\end{description}

\section{Assessment of the explanation quality}\label{sec:metrics}
Evaluating GNNs' explanations is a challenging task that requires to verify if and how the explainer is effective in capturing the behaviour of the model. 
There are two main strategies to evaluate explanation quality. The first is a {\em supervised} 
strategy~\cite{sanchez2020evaluating,ying2019gnnexplainer,funke2022z}, that measures the similarity of the extracted explanation with an existing ground-truth, which is assumed to be known. 
The second strategy measure in an {\em unsupervised} manner how much the prediction of a GNN on the full graph resembles the prediction computed on the 
extracted explanation only. Note that this does not require to have a ground-truth explanation available. 
%The second strategy is an {\em unsupervised} one, that measures how much the prediction of a GNN on the full graph resembles the prediction computed on the extracted explanation only, and this does not require to have a ground-truth explanation available. 
We consider a metric for each of these two strategies, in order to capture different aspects of the quality of an explanation: the {\em plausibility}\cite{rathee2022bagel} of the explanation with respect to a ground-truth concept that an accurate GNN is expected to have learned, and the {\em fidelity}\cite{deyoung2019eraser} of the explanation with respect to the prediction of the GNN to be explained. Specifically, with plausibility we quantify the consistency between the explainer mask and a human-level intuition of what a plausible explanation looks like. On the other hand, fidelity measures the consistency between the model prediction on the full graph and the prediction on the explanation subgraph, and thus it works with a sort of model-based instead of human-based ground truth.

In this work we will evaluate explainers according to both strategies, and study the trade-off between the two. 

% \rev{
% In more general terms, in this work we do not introduce new metrics, but rather rely on existing and well-established ones. The only novelty here is the use of the $f1$-score used to define $F_{f1}$ (see below), which is however a very simple and transparent aggregation, and not really a new metric.
% This choice is motivated by the need to have a few metrics that can be used systematically across all the scenarios. For this reason, we favor simple and well established metrics.
% Moreover, a methodical comparison of all the existing metrics and a choice of best ones is beyond the scope of this paper. Indeed, evaluating the quality of a metric is a difficult task in itself, and to address it one would need an independent way to measure the quality of the explainers. This is something that is lacking at the moment, and one of the reasons motivating this survey.
% }

\subsection{Single-instance metrics}
In the following we detail the metrics we employ, namely \emph{plausibility}\cite{rathee2022bagel} ($P$) and \emph{fidelity} ($F$), the latter further divided in its \emph{comprehensiveness}\cite{deyoung2019eraser} ($F_{com}$) and 
\emph{sufficiency}\cite{deyoung2019eraser} ($F_{suf}$) components.

Each of the scores or metrics is computed for a specific instantiation of a dataset with $n_c\in\N$ classes, a 
class $c\in\{0, 1, \dots, n_c-1\}$, a model, and an explainer. We thus assume that these four are fixed in the following, and we stress that the same 
computation has to be repeated for each of these configurations. Moreover, we remark that the metrics are computed on the training set alone, as we need access 
to the labels of the graphs or nodes.
 
We assume to have a graph $G$ (for graph classification tasks) or a node $v\in G$ (for node classification ones) of class $y=c$, and denote as $g$
the trained GNN. 
We have GNNs which output a class probability prediction vector in form of a soft max, so that the predicted class probabilities sum to $1$. Since we are 
considering one class at a time, in the following we assume to be working with only the output's entry corresponding to class $c$.

Only the graphs (or nodes) which are correctly classified by the trained GNN are considered further and run through the explainer, which returns a corresponding node or edge soft explanation mask $G_{\exp}$ (\rev{see Section~\ref{sec:masks}}). 

Before computing the metrics, these soft-mask explanations are processed and filtered by means of three operations:
\begin{itemize}
\item \emph{Conversion:} Edge masks are converted to node masks by assigning to each node the weight given by the average of the weights of its incident edges. 
This operation makes it easier to compare the scores of edge-based and node-based explainers, and we choose to use node masks since node-based explainers are 
more common in our taxonomy (see Section~\ref{sec:nets}).
\item \emph{Filtering:} For each mask we check the difference between the largest and the smallest weight. If the difference is below a tolerance 
$\tau=10^{-3}$, we discard the graph or node for the given combination of dataset, class, model, and explainer (the graph or node may still pass the filter for 
other settings). The goal of this filter is to discard poorly informative explanations.

\item \emph{Normalization:} The remaining explanation masks are normalized instance by instance, so that each explanation has weights in $[0, 1]$. This has 
the effect of making the computation of the metric uniform across the entire dataset, and comparing its values to those obtained with other settings.
\end{itemize}

% We use the same notation $G_{\exp}$ to indicate the soft explanation mask after this processing steps.

After these operations have been applied, we compute the metrics as follows. We formalize each metric as it is computed on a single 
instance (a graph or a node), and remark that the overall values of plausibility or fidelity for the entire (dataset, class, model, explainer)-configuration is obtained by averaging over these 
single instances.

\paragraph{Plausibility} 
Let $\overline G_{\exp}$ be the expected ground truth for class $c\in\{0, 1, \dots, n_c-1\}$, represented by a copy of the original graph $G$ with an hard mask highlighting the ground truth nodes.
Following \cite{rathee2022bagel}, the plausibility $P$ of the explanation is defined as 
\begin{equation*}
P = \aucroc(G_{\exp}, \overline G_{\exp}),
\end{equation*}
i.e., the area under the ROC curve between the computed soft mask and the ground truth hard mask.

It is clear that this metric can only be computed on benchmarks in which the ground truth explanation can be defined, and it is completely 
dependent on this definition. 
For each dataset, the ground truths that we are using to compute $P$ are defined in Section~\ref{sec:dataset}. 
Whenever multiple ground truths are possible (e.g., the shortest paths in \datasetnctwo), we compute the plausibility of each candidate and consider only the highest one.

\paragraph{Sufficiency}
The fidelity sufficiency $F_{suf}$ \cite{deyoung2019eraser} is the difference in the predicted probability when computed on the graph and on the 
explanation. 
Since the explanation is a soft mask, we fix a number of levels $N_t\in\N$ and apply an incremental thresholding with $N_t+1$ threshold levels $t_k=k / N_t$, 
$k=0, \dots, N_t$, where we define $G_{\exp}(t_k)$ to be the hard mask explanation derived from $G_{\exp}$ with threshold $t_k$. 

Using $N_t=100$, we define the metric by
\begin{equation*}
F_{suf} = \frac{1}{N_t - 1} \sum_{k=1}^{N_t-1} \left(g(G) - g(G_{\exp}(t_k))\right),
\end{equation*}
i.e., the average change in prediction over all the possible hard masks. 
% Observe that for computational convenience omit non used masks

This metric may possibly be negative, and a smaller value indicates a better result. This indeed may happen only if the explanation provides an higher 
probability for the correct class than the entire graph, and thus the explanation mask manages to filter unnecessary parts of the graph.
For this reason this metric is harder to compare to other scores, so when used alone we transform it to a renormalized metric $F_{suf}'$, 
which has values in $[0,1]$ and where $F_{suf}'=1$ means a good quality of the explanation. 
The normalization takes into account the number of classes $n_c$, and it is defined for $p = \frac{n_c-1}{n_c}$ as
\begin{equation*}
F_{suf}' 
= 1-\frac{F_{suf}+p}{1+p} 
% = 1-\frac{f+\frac{n_c-1}{n_c}}{1+\frac{n_c-1}{n_c}}
% = 1-\frac{f n_c + (n_c-1)}{n_c+(n_c-1)}
% = 1-\frac{n_c (f + 1) -1}{2 n_c -1}
% = \frac{2 n_c - 1 - n_c (f + 1) + 1}{2 n_c -1}
% = \frac{n_c (2 - f - 1)}{2 n_c -1}
% = \frac{n_c (1 - f)}{2 n_c -1}
= \frac{n_c}{2 n_c -1}(1 - F_{suf}).
% = \frac{1 - f}{2 -1 / n_c}
\end{equation*}

\paragraph{Comprehensiveness}
The fidelity comprehensiveness $F_{com}$ \cite{deyoung2019eraser} is instead the difference in the predicted probability when computed on the graph and on 
the complement of the explanation.
Proceeding as in the computation of the sufficiency, we define 
\begin{equation*}
F_{com} =  \frac{1}{N_t - 1} \sum_{k=1}^{N_t-1} \left(g(G) - g(G\setminus G_{\exp}(t_k))\right),
\end{equation*}
where now $G\setminus G_{\exp}(t_k)$ is the complement of the hard mask $G_{\exp}(t_k)$.
This metric may as well assume negative values, but good explanations have in this case $F_{com}$ close to $1$ (the 
complement of the explanation provides low probability).

We are not using this metric for node classification datasets, since its evaluation would require to compute the model prediction on $G\setminus G_{\exp}(t_k)$, 
which is a graph that may possibly not contain the node whose classification we are willing to explain.

\paragraph{Fidelity} 

To aggregate $F_{com}$ and $F_{suf}$ into a unique fidelity metric, for graph classification tasks we compute what we call \emph{$f1$-fidelity} ($F_{f1})$, 
which is defined by
\begin{equation*}
F_{f1} = 2  \frac{(1-F_{suf}) \cdot F_{com}}{(1-F_{suf})+F_{com}}.
\end{equation*}
This is indeed the $f1$ score \cite{jones1976information} between $F_{com}$ and $(1-F_{suf})$.
In graph classification tasks, we use this metric in place of $F_{com}$ and $F_{suf}$.

\subsection{Aggregation}\label{sec:aggregation}
After we evaluate any of these metrics on each (dataset, class, model, explainer)-configuration, we need an aggregation mechanism to assign a unique score to the models and the explainers over all classes and datasets. This permits to avoid visualizing the detailed metrics over the entire set of configurations, and make the results easier to be interpreted. \rev{However, we provide plausibility and fidelity values for each (dataset, class, model, explainer)-configuration in Appendix~\ref{sup:tabs}.}

To define these aggregate metrics we proceed as follows, where the same procedures are repeated for both plausibility and fidelity: \textit{(1)} For each (dataset, class, model, explainer)-configuration we keep only the class with the highest value of the metric, i.e., the best explained class. \textit{(2)} For a given dataset, we rank the model-explainer pairs according to the values selected in point (1). The aggregated score of each pair is the ranking number $1, 2, \dots$. \textit{(3)} The dataset-level scoring of an architecture, of an explainer, or of a category of explainers (e.g., grad-based or edge-based) is the average of the scores of point (2) over all the corresponding pairs. \textit{(4)} To obtain global answers (over all the datasets), the scores of point (2) are averaged over the datasets, and the operations of point (3) are repeated using these values.

%\begin{enumerate}
%\item\label{item:point1} For each (dataset, class, model, explainer)-configuration we keep only the class with the highest value of the metric, i.e., the best explained class.
%\item\label{item:point2} For a given dataset, we rank the model-explainer pairs according to the values selected in point \eqref{item:point1}. The aggregated score of each pair is the ranking number $1, 2, \dots$.
%\item\label{item:point3} The dataset-level scoring of an architecture, of an explainer, or of a category of explainers (e.g. grad-based, or edge-based) is the average of the scores of point \eqref{item:point2} over all the corresponding pairs. 
%\item To obtain global answers (over all the datasets), the scores of point \eqref{item:point2} are averaged over the datasets, and the operations of point 
%\eqref{item:point3} are repeated using these values.
%\end{enumerate}
We mention in particular that the answer to the research questions (Section~\ref{sec:rq_definition}) are computed from these aggregations.

To assess instead the stability the explanations over an entire dataset, we propose a qualitative visualization of the masks which is discussed for each experimental setting.

\section{Experimental setting}\label{sec:experimental_setting}
Any explainer provides an explanation of the prediction of a given instance of a model, as it is obtained after an optimization process on a specific dataset. It is thus of paramount importance to identify the choices made in the training of the networks that will be analyzed in the following.

\paragraph{Graph classification}
% \antonio{ricordare che tutte le uniche gnn che hanno global aggreation = sum è numebr of stars } \gabriele{guarda}
We report in Table \ref{tab:gc_model_param} the details of the networks used in each graph classification task, the parameters used for their optimization, and the resulting train and test accuracies.

For each dataset and each architecture, the table shows the dimensions of the hidden layers of GNN type (column \textit{GNN}) and of fully connected type (column \textit{Fully conn.}), and any additional parameter used for the definition of the architecture (see Section~\ref{sec:nets} for a definition of these hyperparameters). For example, in the first row the numbers $30-30-30$ and $10-2$ mean that three \gcn layers are applied, each mapping to a target dimension of $30$, followed by two fully connected layers with target dimensions $10$ and $2$. 
We remark that the final aggregation function is a mean for all datasets except for \datasetgcthree, where we used a sum aggregation.
The table additionally reports the learning rate (column \textit{LR}) and number of epochs used in the training, where an ADAM optimizer has been used in each case. 
We remark that these configurations have been chosen with the guiding principle of obtaining the simplest configuration achieving a target $0.95$ train accuracy. 
\rev{In particular, no validation set has been used. This choice is motivated by the fact that the explanations are computed on the training set. This is a common (sometimes implicit) choice in the explainability literature, with the rationale that explanations should identify what the model learned during training, possibly highlighting patterns that do not generalize to test examples. However, to avoid extracting spurious explanations for models that did not learn anything sensible, only the models with a training accuracy of at least $0.95$ have been further analyzed, while all the others have been discarded.}
The last two columns of Table \ref{tab:gc_model_param} show the resulting train and test accuracies obtained by the models trained according to these specifications, \rev{where an "X" indicates that it was not possible to achieve the desired target accuracy (in this case, the row reports the configuration of the largest architecture which has been tested).}
%We remark that for some configurations it was not possible to achieve the desired target accuracy, so the table reports an "X" in place of the accuracy values, and it reports the values corresponding to the largest architecture which has been tested. Being not sufficiently accurate, these models have been removed from the successive analysis.
\rev{ It is easy to see that with a $0.95$ threshold on training accuracy, training and test accuracies end up being very similar.}

\begin{table}[ht]
    \small
    \centering
    \begin{tabular}{ c| c | c | c | c |c |c | c || c | c}
         Dataset & Architecture & \makecell{GNN} & \makecell{Fully conn.} & HyperParams & LR & Epochs & \makecell{Train\\ Acc} & \makecell{Test\\ Acc}\\ 
         \hline \hline 
         \multirow{7}{*}{\datasetgcone}& \gcn & 30-30-30 & 10-2 & - & 0.001 & 1500 & 0.994 & 0.998\\ %&2522
          & \sage & 30-30-30 & 10-2& - & 0.01 & 3000 & X & X\\ %&4622
          & \gat &  30-30-30 & 10-2  & heads = 1 & 0.01 & 3000 & X & X\\ %&2702
          & \gin & 30-30 & 30-2  & - & 0.001 & 1000 & 1.0 & 1.0\\ %&2252
          & \cheb & 30-30  & 30-2 & - & 0.001 & 1000 & 1.0 & 1.0\\ %&7052
          %& DiffPool & 64-64-64  & 64-2 & DenseGCN & 0.001 & 5000 & X & X\\ %&92836
          & \mincut & 32-32-32  & 32-2 & - & 0.001 & 700 & 0.92 & 0.93\\ 
          & \setset &30-30-30 & 10-2 & - & 0.001 & 1500 & 0.97 & 0.97\\ %&13862
          & \graphconv & 30-30 & 30-2 & - & 0.001 & 500 & 1.0 & 1.0\\ %&3452
          \hline
         \multirow{7}{*}{\datasetgctwo}& \gcn & 60-60-60-60 & 60-10-2 & - & 0.001 & 7000 & 0.97 & 0.97\\ 
          & \sage & 60-60-60-60 & 60-10-2& - & 0.01 & 3000 & X & X\\  
          & \gat &  60-60-60-60 & 60-10-2  & heads = 3 & 0.01 & 3000 & X & X\\
          & \gin & 30-30 & 30-2  & - & 0.001 & 1000 & 0.99 & 1.0\\
          & \cheb & 30-30-30  & 30-2 & - & 0.001 & 1000 & 1.0 & 0.98\\
          %& DiffPool & 64-64-64  & 64-2 & DenseGCN & 0.001 & 5000 & X & X\\
          & \mincut & 32-32-32  & 32-2 & - & 0.001 & 700 & 0.95 & 0.95\\ 
          & \setset & 60-60-60-60 & 60-10-2 & - & 0.001 & 1500 & 0.97 & 0.97\\
          & \graphconv & 30-30 & 30-2 & - & 0.001 & 500 & 1.0 & 1.0\\ 
          \hline
          \multirow{7}{*}{\datasetgcthree}& \gcn & 70-70-70 & 30-3 & - & 0.005 & 1000 & 0.99 & 1.0\\ 
          & \sage & 30-30-30 & 30-3 & - & 0.01 & 3000 & X & X\\
          & \gat &  30-30-30 & 10-3 & heads = 1 & 0.01 & 3000 & X & X\\
          & \gin & 40-40 & 30-3  & - & 0.001 & 3000 & 0.99 & 1.0\\
          & \cheb & 30-30  & 30-3 & - & 0.001 & 1000 & 0.99 & 0.99\\
          %& DiffPool & 64-64-64 & 30-3 & DenseGCN & 0.001 & 1500 & X & X\\
          & \mincut & 32-32-32 & 32-3 & - & 0.001 & 400 & 0.99 & 0.99\\
          & \setset & 70-70-70 & 30-3 & - & 0.001 & 1500 & 0.99 & 0.99\\
          & \graphconv & 30-30 & 30-3 & - & 0.001 & 500 & 0.99 & 0.99\\ 
          \hline
          \multirow{7}{*}{\datasetgcfour}& \gcn & 30-30 & 15-2 & - & 0.001 & 4000 & 0.99 & 0.99\\ 
          & \sage & 30-30-30 & 30-2 & - & 0.001 & 1000 & 1.0 & 0.99\\  
          & \gat & 10-20-40 & 10-2 & heads = 2 & 0.001 & 500 & 0.99 & 0.99\\
          & \gin & 30-30 & 30-2  & - & 0.001 & 1000 & 1.0 & 0.99\\
          & \cheb & 30-30  & 30-2 & - & 0.001 & 500 & 1.0 & 1.0\\
          %& DiffPool & 30-30  & 15-2 & DenseGCN & 0.001 & 4000 & X & X\\
          & \mincut & 32-32  & 32-2 & - & 0.001 & 400 & 0.96 & 0.97\\
          & \setset & 30-30  & 15-2 & - & 0.001 & 1500 & 1.0 & 0.99\\
          & \graphconv & 30-30  & 30-2 & - & 0.001 & 500 & 1.0 & 1.0\\
    \end{tabular}
    \caption{Configuration of the graph classification models, and corresponding accuracies. The table reports for each dataset and each architecture the dimension, number, and hyperparameters defining the hidden layers, together with the optimization parameters, and the obtained train and test accuracies.  
    The configuration-dataset pairs which did not reach the target $95\%$ train accuracy are marked with an "X", and are not further analyzed in this work.
    }
    \label{tab:gc_model_param}
\end{table}

\paragraph{Node classification}
In the same way, we report in Table~\ref{tab:nc_model_param} the configurations and accuracies related to the node classification tasks.

% questi includono anche il fine tuning che abbiamo fatto per far funzionare gli explainers

\begin{table}[ht]
    \small
    \centering
\begin{tabular}{ c| c | c | c | c |c || c | c}
         Dataset & Architecture & \makecell{GNN} & \makecell{Fully conn.}  & LR & Epochs & \makecell{Train\\ Acc} & \makecell{Test\\ Acc}\\ 
         \hline \hline 
         \multirow{5}{*}{\datasetncone}
          & \gcn & 30-30-30  & 10-4   & 0.0005 & 2000 & 0.98 & 0.98\\ 
          & GraphSAGE & 30-30 & 4 & 0.005 & 2000 & 1.0 & 1.0\\
          & \gat & 30-30-30 & 10-4 & 0.0005 & 2000 & X & X\\ 
          & \gin & 70-70-70 & 4 & 0.0005 & 5000 & 0.96 & 0.95\\ 
          & \cheb & 30-30 & 4 & 0.0005 & 300 & 1.0 & 1.0\\
          \hline
          
          \multirow{5}{*}{\datasetnctwo}
          & \gcn & 30-30  & 30   & 0.005 & 500& 0.951 & 0.950\\ 
          & GraphSAGE & 30-30 & 30  & 0.005 & 500 & 1.000 & 0.995\\ 
          & \gat & 30-30 & 30 & 0.005 & 500& 0.977 & 0.975\\ 
          & \gin & 30-30 & 30 & 0.005 & 500 & 0.952 & 0.950\\ 
          & \cheb & 30-30 & 30 & 0.0005 &  600 &0.993& 0.950 \\ 
    \end{tabular}
    \caption{Configuration of the node classification models, and corresponding accuracies. The table reports for each dataset and each architecture the dimensions and number of hidden layers, together with the optimization parameters, and the obtained train and test accuracies.  
    The configuration-dataset pairs which did not reach the target $95\%$ train accuracy are marked with an "X", and are not further analyzed in this work.
    }
    \label{tab:nc_model_param}
\end{table}

\section{Results}\label{sec:results}
\subsection{Research questions}\label{sec:rq_definition}
The comparative analysis of the behavior of the explainers is developed along the following research questions, which will apply to both node and graph classification tasks.
\begin{itemize}
\item \textbf{RQ1: How does the architecture affect the explanations?}
This research question can be naturally divided into the following three subquestions:
\begin{itemize}
    \item \emph{\textbf{RQ1.1:} Which is the architecture that has the best explanation? } With this question we would like to understand which is the architecture that achieves the best score, either in terms of f1-fidelity or plausibility.
    \item \emph{\textbf{RQ1.2:} Which is the easiest architecture to explain?} This question aims to find what is the architecture that is well explained by the greatest number of explainers.
    \item \emph{\textbf{RQ1.3:} Which is the hardest architecture to explain?} In this case we want to search for an architecture that achieves the lowest score.
\end{itemize}

\item \textbf{RQ2: How do explainers affect the explanations?}
Even this question can be divided into subquestions, which try to cover different open problems related to state-of-the-art GNN explainers. We identify them as follows:
\begin{itemize}
    \item \emph{\textbf{RQ2.1}: Which is the explainer that explains in the best way?} Here we are interested in finding the explainer able to obtain the highest plausibility or fidelity.
    \item \emph{\textbf{RQ2.2}: Which is the explainer that explains the maximum number of architectures?} This aspect is particularly important because we need explainers which are robust with respect to different GNN architectures. 
    \item \emph{\textbf{RQ2.3}: Which is the category of explainers that provides the best explanations?} The subquestion searches for the best category of explainers. As defined by \cite{yuan2020explainability}, we consider three macro-categories, namely gradient-based (Grad), perturbation-based (Pert), and decomposition-based (Dec). 
    \item \emph{\textbf{RQ2.4}: Which is the best mask type between node and edge?} By answering this question we investigate if there is an advantage for explainers based on node or edge importance.
\end{itemize}
We would like to remark that \textbf{RQ2.3} and \textbf{RQ2.4} are particularly relevant for future research in GNN explainability, since they may provide actionable guidelines for the development of new explainers.

\item \textbf{RQ3: How do different types of problems affect the explanations?} To address this question we analyze a series of different problems such as: single versus multiple concepts, counting substructures and others. Each problem has been investigated throw different datasets, that will be discussed in relation to this question.
\end{itemize}

\subsection{Graph classification}
\subsubsection{RQ1: How does the architecture affect the explanations?}\label{subsec:rq1}
Table \ref{tab:rq1_all} visualizes in a compact form the answer to research questions \textbf{RQ1}, where the aggregation mechanism described in Section~\ref{sec:aggregation} has been used to identify a ranking of the architectures for each dataset across all explainers, both in terms of plausibility and fidelity and the highest ranking architecture is reported for each research question and dataset. 
Each column in the table refers to a specific dataset (\datasetgcone to \datasetgcfour, see Section~\ref{sec:dataset}), while the first column shows the overall answer obtained by considering all the datasets at once. Also in this case we refer to Section~\ref{sec:aggregation} for the details of the computation of this ranking.

% \textbf{:} 
% The best explanation among all datasets is achieved with \graphconv in both plausibility and fidelity. However, GNN's explanations have different behaviours for each dataset. 
Although the architecture with the best explanation (\textbf{RQ1.1}) varies depending on the dataset, the overall best performer is \graphconv both for plausibility (paired with \sal) and fidelity (paired with \igedge). This indicates a sort of consistency in the performances of \graphconv: It may not be the single best one for any dataset, but it is always among the best performing ones, in a way that makes it to be the best in terms of aggregated scores.

% \textbf{:} 
The overall easiest architecture to explain on average (\textbf{RQ1.2}) is \gcn for both plausibility and fidelity, \rev{and even at the single dataset level it prevails in three out of five datasets for plausibility, in four out of five for fidelity. Moreover, the two metrics agree to identify \gcn as the easiest to explain in three out of five datasets}. %, even if each dataset has a different behaviour. 
A possible motivation behind this result may come from the fact that \gcn are among the simplest models based on message passing, even if other architectures (e.g., \sage and \gin) are equally simple. A difference so large may thus be related to other aspects that we are unable to identify. 
% However, it is worth mentioning that the only difference between \gcn and \setset is on the global aggregation function, since the underlying GNN is still a \gcn in both cases. 
Moreover, we remark that \gcn and \setset work with the same underlying GNN layers (\gcn), and they differ only in the final aggregation operation (a sum in \gcn, and an LSTM in \setset). The better performances of \gcn are thus hinting to the fact that a different global aggregation alone is responsible for changing a network's explainability, and that linear aggregations (\gcn) are, perhaps unsurprisingly, easier to explain than nonlinear ones (\setset). In general terms, the role of the global aggregation function and its stability across different tasks is yet to be fully understood and has not received great attention in the explanation literature. Thus, we believe that a systematic study in this direction may be an interesting future topic of research.

% \rev{\sout{The two metrics of plausibility and fidelity agree only in \datasetgcfour. This is the only dataset with meaningful node features, and we expect that this makes the learning process more stable and closer to the identification of concepts closer to GT, so that the same explainer is able to provide high score explanations for both metrics.}}
% , and this may be due to the fact that meaningful node features are enough for the pooling mechanism to aggregate accurate and faithful concepts.

Finally, from the table it is easy to see that \gin is the most difficult network to explain (\textbf{RQ1.3}), for both plausibility and fidelity in each dataset except for \datasetgcthree. This stark difficulty in explaining \gin is not directly understandable, especially because it is implemented with a single-layer MLP, which does not introduce stronger non-linearity than a simpler \gcn. \revDue{A possible explanation for this behaviour is the fact that most explainers have been developed and tested on \gcn, potentially introducing a bias that affects the performance of that architecture. This aspect warrants further investigation in future studies.}
% may be due to the presence of a MLP in the definition of the layer, which introduces a strong nonlinearity that may render the model's predictions more difficult to explain. 
% \gabriele{amenable to future studies. Dire che forse e' il pooling (confornta nodi).}

Moreover, the difficulty in explaining \mincut on \datasetgcthree may be due to the fact that in this case the task is to count the number of occurrences of a concept (the stars) in a dataset. Although the concept itself is the easiest possible to identify for message passing based GNNs, it seems that concept-counting is hard to explain for a pooling mechanism. %\antonio{too strong?}

%\begin{table}[ht]
%\small
%\centering
%\begin{tabular}{c|ccccc}
%\multicolumn{6}{c}{Plausibility}\\
%&\textbf{All} & \datasetgcone& \datasetgctwo& \datasetgcthree & \datasetgcfour \\
%\hline
%{RQ1.1} & \textbf{\graphconv} & \graphconv & \cheb & \setset & \gcn    \\
%{RQ1.2} & \textbf{\gcn}       & \cheb      & \gcn  & \setset & \mincut \\
%{RQ1.3} & \textbf{\gin}       & \gin       & \gin  & \mincut & \gin    \\
%\multicolumn{6}{c}{}\\
%\multicolumn{6}{c}{Fidelity}\\ 
%&\textbf{All} & \datasetgcone& \datasetgctwo& \datasetgcthree & \datasetgcfour \\
%\hline
%{RQ1.1} & \textbf{\graphconv} & \cheb & \setset & \graphconv & \setset \\
%{RQ1.2} & \textbf{\gcn}       & \gcn  & \mincut & \graphconv & \mincut \\
%{RQ1.3} & \textbf{\gin}       & \gin  & \gin    & \mincut    & \gin    \\
%\end{tabular}
%\caption{Experimental answer to \textbf{RQ1} for graph classification. The  table shows the top-ranking architecture with respect to each subquestion \textbf{RQ1.1}, \textbf{RQ1.2}, \textbf{RQ1.3}, for each dataset \datasetgcone to \datasetgcfour, and overall. The rankings are computed with respect to the Plausibility and the Fidelity metrics, and the colors identify different architectures.}
%    \label{tab:rq1_all}
%\end{table}

\begin{table}[ht]
\small
\centering
\begin{tabular}{c|ccccc}
\multicolumn{6}{c}{Plausibility}\\
&\textbf{All} & \datasetgcone& \datasetgctwo& \datasetgcthree & \datasetgcfour \\
\hline
{RQ1.1} & \textbf{\graphconv} & \graphconv & \cheb & \setset & \gcn    \\
{RQ1.2} & \textbf{\gcn}       & \rev{\gcn}     & \gcn  & \setset & \mincut \\
{RQ1.3} & \textbf{\gin}       & \gin       & \gin  & \mincut & \gin    \\
\multicolumn{6}{c}{}\\
\multicolumn{6}{c}{Fidelity}\\ 
&\textbf{All} & \datasetgcone& \datasetgctwo& \datasetgcthree & \datasetgcfour \\
\hline
{RQ1.1} & \textbf{\graphconv} & \rev{\graphconv} & \setset & \graphconv & \setset \\
{RQ1.2} & \textbf{\gcn}       & \gcn       & \rev{\gcn}   & \graphconv & \rev{\gcn}   \\
{RQ1.3} & \textbf{\gin}       & \gin       & \gin    & \mincut    & \gin    \\
\end{tabular}
\caption{\rev{Experimental answer to \textbf{RQ1} for graph classification. The table shows the top-ranking architecture with respect to each subquestion \textbf{RQ1.1}, \textbf{RQ1.2}, \textbf{RQ1.3}, for each dataset \datasetgcone to \datasetgcfour, and overall. The rankings are computed with respect to the plausibility and the fidelity metrics, and the colors identify different architectures.}}
    \label{tab:rq1_all}
\end{table}

% \begin{figure}[ht]
%     \centering
%     \includegraphics[width=0.7\textwidth]{Figures/tab/RQ1_all_data_2.pdf}
%     \caption{Experimental answer to \textbf{RQ1} for graph classification. The  table shows the top-ranking architecture with respect to each subquestion \textbf{RQ1.1}, \textbf{RQ1.2}, \textbf{RQ1.3}, both for each dataset \datasetgcone to \datasetgcfour, and overall. The rankings are computed with respect to the Plausibility and the Fidelity metrics, and the colors identify different architectures.}
%     \label{fig:tab:rq1_all}
% \end{figure}

\begin{figure}[ht]
    \centering
    \includegraphics[width=0.99\textwidth]{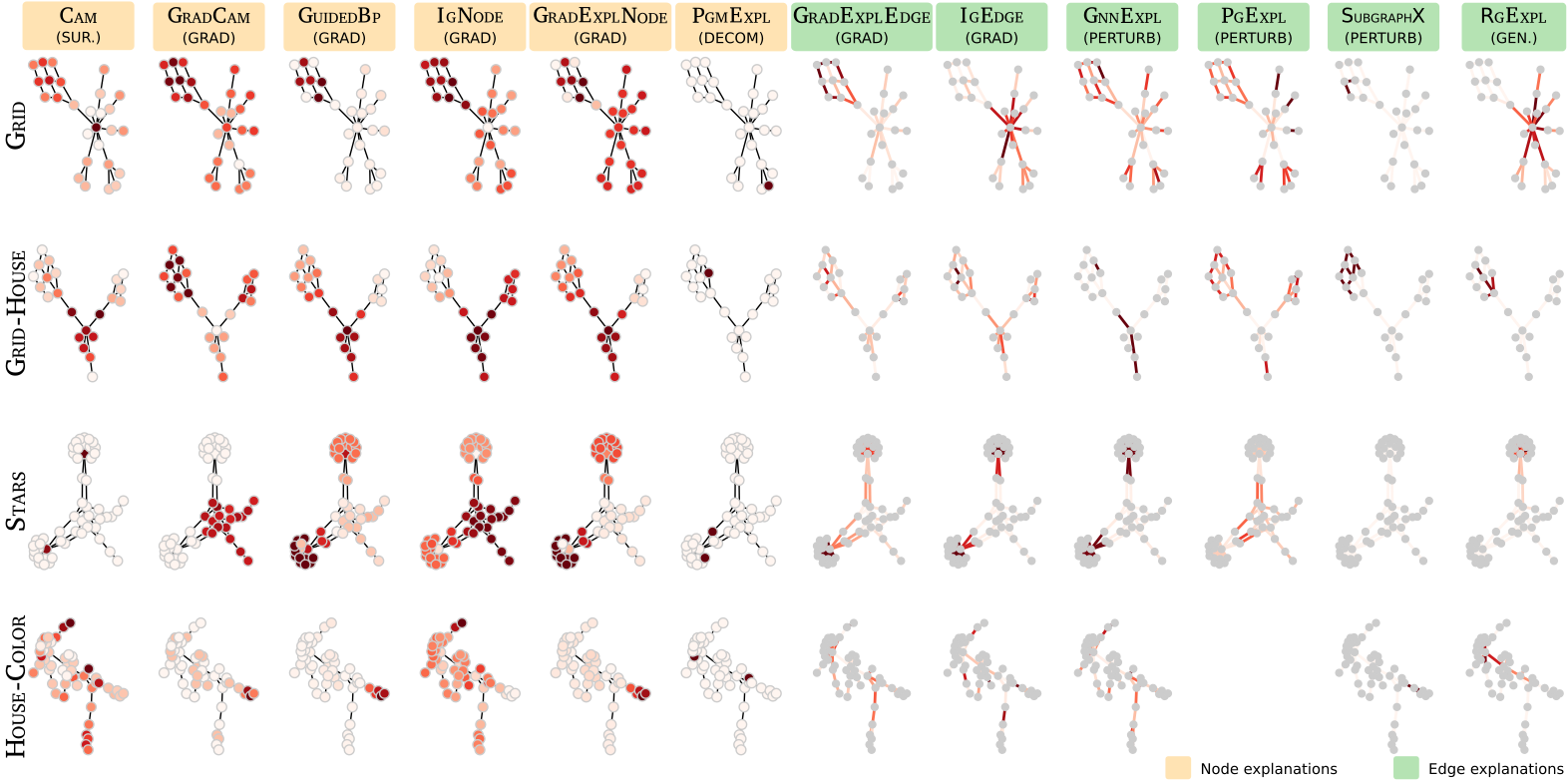}
    \caption{\rev{Explanation masks (node- or edge-based) computed by the different explainers on the predictions of \gcn. Each row visualizes the mask computed for a given random graph from each dataset.}}
    \label{fig:rq12}
\end{figure}

% Although , we wish to stress that there is a significant amount of variability that has still to be taken into account, and which will be addressed through the following research questions.
To offer additional insight into this fine-grained behavior of \gcn, which is the easiest architecture to explain according to RQ1.2 in terms of both metrics, we pick a random element from each dataset and analyze the mask provided by each explainer. Figure~\ref{fig:rq12} reports these masks, where the rows show the representative graph from each dataset, and each column corresponds to a different explainer.
The first five explainers return node importance, while the last four are edge-based. In both cases, the node or edge importance is rendered by a different color intensity. 
We stress once again that only one graph per dataset is shown in the figure, and thus the following discussion is of a rather qualitative nature, to be complemented with the metrics discussed in the first part of this section.

\gcn manages to achieve good results in terms of both plausibility and fidelity, meaning that its explanations are both close to the expected ground truth and to the actual one used by the models to realize their prediction. This duality can be observed across the examples of Figure~\ref{fig:rq12}. Indeed, there are cases where the masks identify clearly the grids in \datasetgcone (\cam, \gradcam, \ignode, \sal, \pgexpl), the grid, the house and the path connecting them in \datasetgctwo (most node-based explainers, and \pgexpl), the stars in \datasetgcthree (\guidedbp and \gradexp), and the colored house in \datasetgcfour (\gradcam, \guidedbp, \gradexp). On the other hand, for many other dataset-explainer combinations the mask is less localized and interpretable by a human eye, but the overall high scores of this explainer suggest these explanations could still be good, at least in terms of fidelity and thus from a model perspective, even if they deviate from the expected ground truth. A more in-depth analysis of the actual behavior on each dataset is presented in Section~\ref{subsec:rq3}.

%  In the first dataset, the grid motif is captured by almost every explainer, except for \pgmexpl and \igedge. However \gradcam, \ignode, \gradexp, and \gnnexpl capture many nodes not belonging to the grid motif, while \pgexpl is able to separate the grid motif from the other parts of the network. Finally, \guidedbp and \sal capture only the grid motif, or part of it. 
% The second dataset is much more complex, because we have to capture both the house and the grid together. 
% Node-importance based explainers are not good in capturing both motifs, while \sal and \igedge succeed in capturing them both (with some noise), with the caveat of trying to find a path that connects them. Interestingly, \pgexpl is not only able to capture both motif, but it also manages to keep them disconnected. 
% In the third dataset, both \cam and \gradcam are able to capture the center of the star, but not the entire star. 
% \guidedbp and \gradexp capture the stars, but also a lot of non important nodes. 
% In terms of edge importance explanations, \igedge and \gnnexpl identify the star structure without adding spurious edges. 
% The last row shows the results on the fourth dataset. In this case, node-based explainers work better than those based on edges. In particular, \gradcam and \guidedbp outperform the other ones. 

\subsubsection{RQ2: How do explainers affect the explanations?}\label{subsec:rq2}
The answers to this question are summarized in Table \ref{tab:rq2_all}, where we used again the aggregation strategies defined in Section~\ref{sec:aggregation} to establish a ranking of the explainers and select the best ones, both for each dataset and overall.

\rev{Interestingly, the overall best explainer is different for plausibility and fidelity in absolute terms (\textbf{RQ2.1}), but it is the same (\subgraphx) when looking at the best performing one on average across all architectures (\textbf{RQ2.2}). They are however all edge-based.}
% In fact, the highest plausibility is achieved by \sal, while the highest fidelity is achieved by \igedge. Although different, they both produce an edge importance mask with a gradient-based mechanism. 

At the dataset level, it is worth remarking that \datasetgcfour is the only one whose absolute best explainer (RQ2.1, \guidedbp) is node-based, and this may clearly be due to the fact that this dataset is the only one with meaningful node features.

In terms of average performances (\textbf{RQ2.3}), perturbation-based explainers are those that best explain all the models for plausibility\rev{, while generation-based explainers prevail with respect to fidelity, both at the aggregate and single-dataset levels}. \rev{In the case of plausibility, the single best explainer is instead gradient-based (RQ2.1)}, but this discrepancy is similar to what happens in node classification (Section \ref{subsec:rq2_nc}). However, while for node tasks the local gradient-based explainers worked better, here perturbation mechanisms are more effective, and this is understandable since graph classification may benefit from these more global types of explanation.
% The answer to this question is particularly remarkable, since perturbation-based explainers are selected as the best ones for both metrics and for all datasets, with the single exception of fidelity for \datasetgcthree, and also overall.

When looking at the average over the entire groups (\textbf{RQ2.4}), edge-mask based explainers are clearly overperforming node-based ones, in accordance with RQ2.1 and RQ2.2. 
We argue that this may be due to the fact that edge-based explainers have been developed specifically for graph-explanation tasks, while node-based ones are all adaptations of existing explainers, introduced for other settings. We remark once again that this is the case only for graph classification, while for node-based tasks (Section \ref{subsec:rq2_nc}) node-based explainers appear to be superior.

\begin{table}[ht]
\small
\centering
\begin{tabular}{c|ccccc}
\multicolumn{6}{c}{Plausibility}\\
&\textbf{All} & \datasetgcone& \datasetgctwo& \datasetgcthree & \datasetgcfour \\
\hline
{RQ2.1} & \textbf{\sal}       & \rev{\rgexpl}   & \pgexpl &  \igedge & \guidedbp    \\
{RQ2.2} & \rev{\textbf{\subgraphx}}  & \rev{\subgraphx} & \pgexpl  & \sal    & \rev{\subgraphx} \\
{RQ2.3} & \textbf{Pert}       & \rev{Gen}      & Pert  & Grad & Pert    \\
{RQ2.4} & \textbf{Edge}       & Edge       & Edge  & Edge & Edge    \\
\multicolumn{6}{c}{}\\
\multicolumn{6}{c}{Fidelity}\\ 
&\textbf{All} & \datasetgcone& \datasetgctwo& \datasetgcthree & \datasetgcfour \\
\hline
{RQ2.1} & \textbf{\igedge}     & \rev{\subgraphx} & \igedge    & \sal     & \pgexpl    \\
{RQ2.2} & \rev{\textbf{\subgraphx}} & \rev{\rgexpl}   & \rev{\subgraphx}& \gnnexpl & \rev{\subgraphx}\\
{RQ2.3} & \rev{\textbf{Gen}}       &\rev{Gen}    & \rev{Gen}     & \rev{Gen}    & Pert    \\
{RQ2.4} & \textbf{Edge}        & Edge  & Edge    & Edge    & Edge    \\
\end{tabular}
\caption{\rev{Experimental answer to \textbf{RQ2} for graph classification. The table reports the top-ranking explainer with respect to each subquestion \textbf{RQ2.1}-\textbf{RQ2.4}, for each dataset \datasetgcone to \datasetgcfour, and overall. The rankings are computed with respect to the plausibility and the fidelity metrics, and the colors identify different explainers.}
    \label{tab:rq2_all}}
\end{table}

% \begin{figure}[ht]
%     \centering
%     \includegraphics[width=0.7\textwidth]{Figures/tab/RQ2_all_data.pdf}
%       \caption{Experimental answer to \textbf{RQ2} for graph classification. The  table reports the top-ranking explainers with respect to each subquestion \textbf{R02.1}-\textbf{RQ2.4}, both for each dataset \datasetgcone to \datasetgcfour, and overall. The rankings are computed with respect to the Plausibility and the Fidelity metrics, and the colors identify different explainers.}
%     \label{fig:tab:rq2_all}
% \end{figure}

\begin{figure}[ht]
    \centering
    \includegraphics[width=0.99\textwidth]{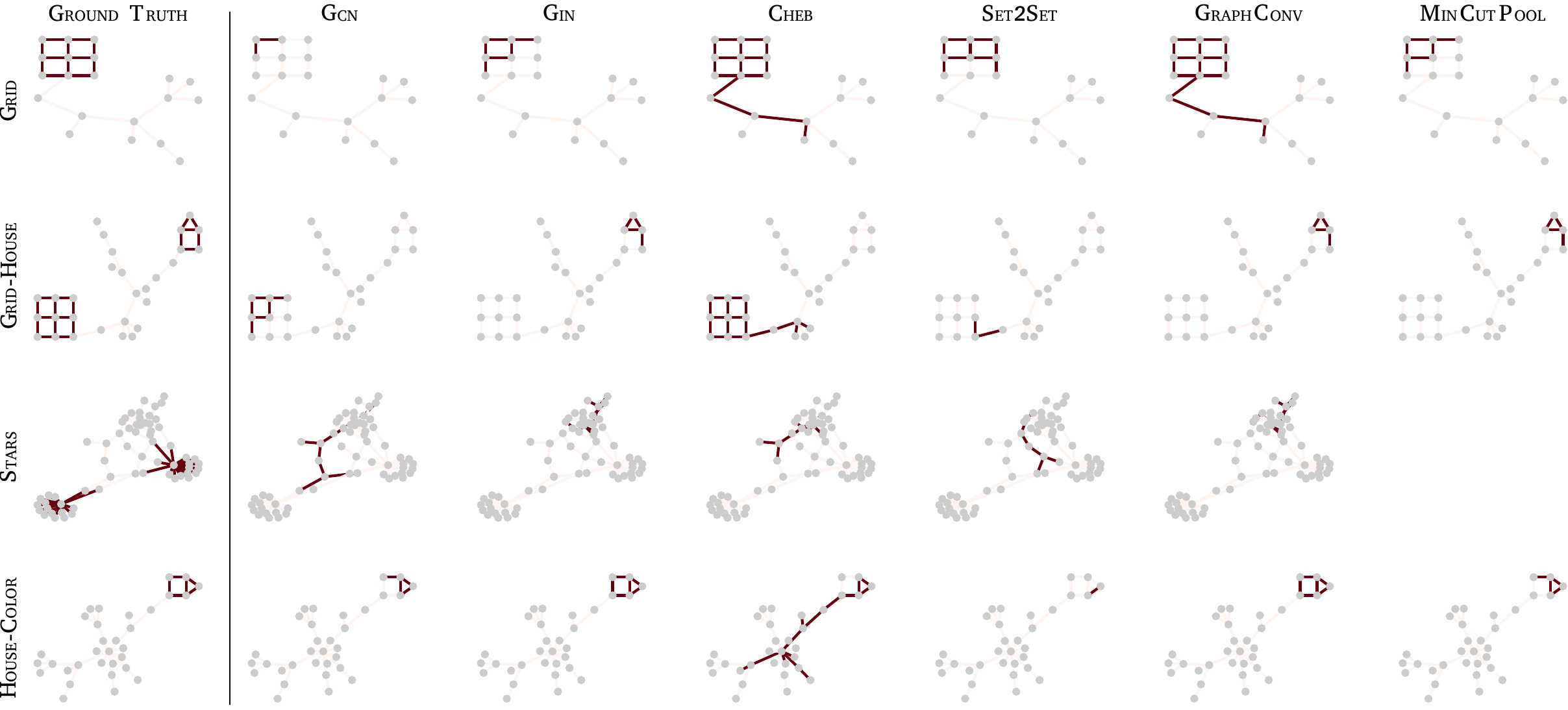}
    \caption{\rev{Explanation masks computed by \subgraphx on the predictions of the different models. Each row visualizes the mask computed for a given random graph from each dataset.}}
    \label{fig:rq2}
\end{figure}

Similarly to the previous section, Figure \ref{fig:rq2} zooms into \rev{\subgraphx}, which is the best-ranking explainer according to RQ2.1 and with respect to plausibility. 
% \rev{\sout{We show the results for each dataset and each GNN for which \sal works, i.e., those that accept edge weights in the training phase.}}
The high plausibility of the explainer means that it is effective in identifying the human-expected explanations in the graphs, and this is clearly visible in the examples of Figure \ref{fig:rq2}: with a few exceptions, the dark red edges identify the grid in \datasetgcone, a path connecting the grid and the house in \datasetgctwo, the stars in \datasetgcthree, and the colored house in \datasetgcfour.
% The first row of figure \ref{fig:rq2} shows the explanations obtained on \datasetgcone. \gcn and \setset have a similar behaviour in terms of explanations, and this is not surprising since the only difference is on the global pooling mechanism. \sal on \cheb does capture the grid, although it also includes the edge connecting it to the BA structure. Finally, \sal on \graphconv captures the grid structure with some noise.
% The second row shows the result obtained on the \datasetgctwo, in which we have multiple concepts (house and grid). Again, we have a similar behavior between \gcn and \setset. Even if \graphconv is good in capturing both the concepts, it has some noise edges.
% In the third row of Figure~\ref{fig:rq2}, all the GNNs are able to capture the star. This result is not surprising, since each GNN is based on message passing, thus stars are the easiest concepts that can be captured. However, \graphconv not only capture the stars, but it also in not affected by noise as in the case of \cheb.
% Finally, in the last row of figure \ref{fig:rq2}, we report the explanations obtained on the \datasetgcfour. Again almost all the GNNs are able to capture the ground truth. Interestingly, simpler GNNs (\gcn and \setset) are those which are less influenced by noisy edges. This may indicate that, with meaningful node features, simpler GNNs architecture are more effective.

\subsubsection{RQ3: How do different types of problems affect the explanations?}\label{subsec:rq3}

To address the third question we analyze the datasets separately. We remark that each dataset has been chosen to represent different types of challenges, which will be discussed in each of the following paragraphs.

\paragraph{\textbf{\datasetgcone}}
In the first dataset, the concept is a grid attached to a random Barabási–Albert (BA) network. Since the BA component is identical in the positive and negative classes, the only discriminative subgraph is the grid (or part of it). In fact, the minimal discriminant subgraph for this dataset is a square, because the BA component does not contain it.

\begin{figure}[ht]
    \centering
    \includegraphics[width=0.99\textwidth]{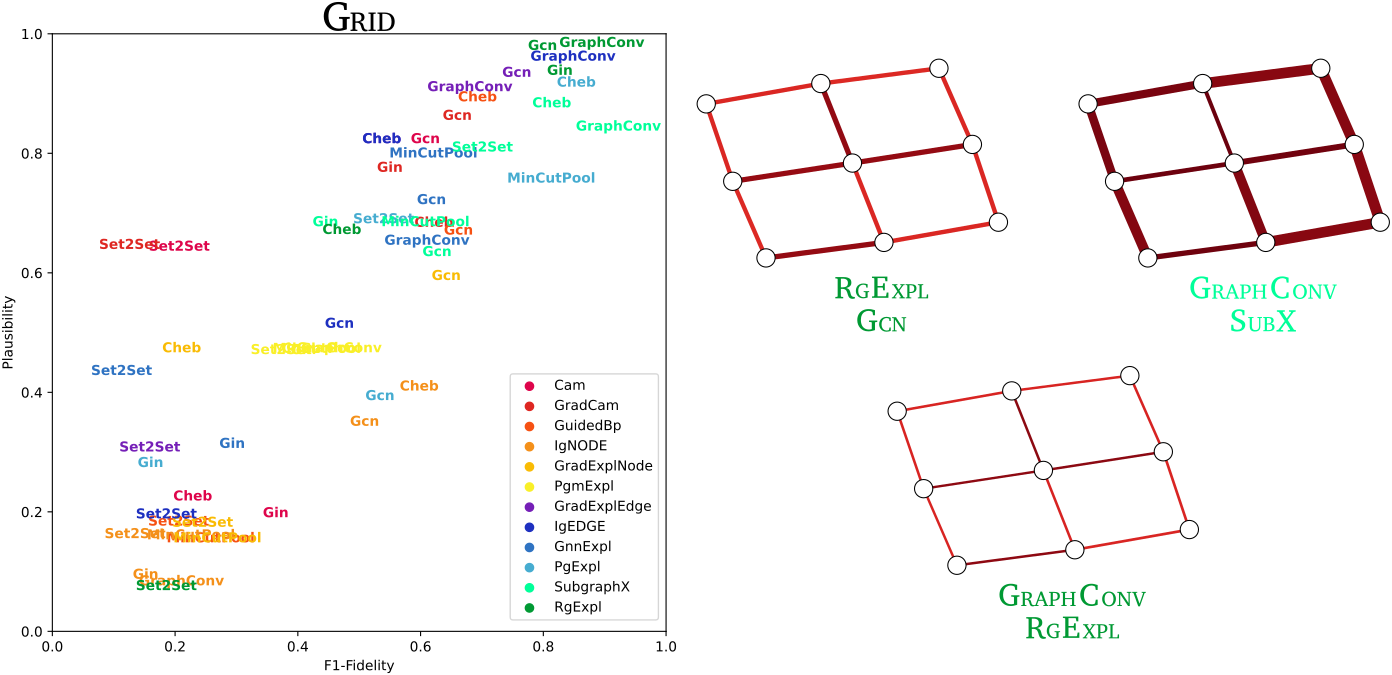}
    \caption{\rev{Left: fidelity and plausibility achieved by all the model-explainer pairs when applied to \datasetgcone. In each pair, the name refers to the model while the color identifies the explainer. Right: stability of the explanations for the three top-performing model-explanation pairs. The colors identify important edges (dark red), and the edge thickness the variability of the importance in the dataset.}}
    \label{fig:pareto_d1_and_stability}
\end{figure}

The left panel of Figure \ref{fig:pareto_d1_and_stability} visualizes the performance of each model-explanation pair when applied to this dataset. 
Each pair is located according to the two-dimensional coordinate given by the resulting fidelity (horizontal axis) and plausibility (vertical axis), and it is identified by the model name and by a color representing the explainer. 
We use warm colors for node-based explainers, and cold colors for edge-based ones. 

This visualization permits to identify those model-explanation pairs that strike the best balance between the two scores, namely, models that maximize both plausibility and fidelity are in the top right corner of the figure.
It is first relevant to observe that a clear positive correlation emerges for the top-performing pairs, in the sense that there are no cases where high fidelity is achieved without a correspondingly high plausibility, and vice-versa.
Moreover, the highest plausibility is achieved by \graphconv with \rev{\rgexpl}, while \rev{\graphconv} with \rev{{\subgraphx}} obtains the highest fidelity. 
These are also the two Pareto-optimal pairs, i.e., any other pair reaches either a smaller fidelity or a smaller accuracy. In this sense, they are the best ones according to this evaluation.

Moreover, it is remarkable to observe that the best performing pairs (thus including also the pairs \rev{\gcn-\rgexpl, \graphconv-\igedge, \gin-\rgexpl, \cheb-\pgexpl}) have all edge-based explainers. This fact is in perfect accordance to the answer to RQ2.4 (Section~\ref{subsec:rq2}), which identifies this type of explainer as superior to node-based ones. 
Observe however that \datasetgcone has no node features, and this may bias this aspect.
% \andrea{se e' per task senza node features, va rimarcato! e.g. for tasks where node features are unavailable.}

For these \rev{two top-performing pairs (\graphconv-\rgexpl and \graphconv-\subgraphx)} we further investigate the quality of the explanations by quantifying their stability. Namely, we are interested in understanding how the different instances of graphs in the dataset are explained, and if there is any recurring pattern in these explanations.

This stability is shown in the right panel of Figure~\ref{fig:pareto_d1_and_stability}, where each edge in the grid motif is colored according to its importance averaged over all the networks in \datasetgcone, with a color scale ranging from white (for importance 0) to dark red (for importance 1).  
The width of each edge is instead proportional to the standard deviation of the explanation across the dataset, such that thicker edges describe a larger deviation, hence a smaller stability, and vice-versa.

\rev{While \graphconv-\subgraphx provides significant explanations (dark red edges), we can observe a very high variability across the dataset (thick edges). On the other hand, \graphconv-\rgexpl provides consistent explanations over the entire dataset (thin edges), but with a less pronounced emphasis on the significant edges (light red edges).}

\begin{figure}[ht]
    \centering
        \includegraphics[width=0.99\textwidth]{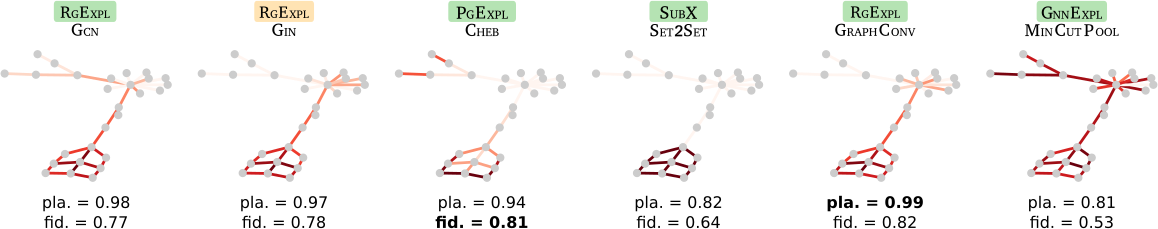}
    \caption{\rev{Examples of explanations provided for each model and its highest plausibility explainer, when applied to a random sample from \datasetgcone. The plausibility and fidelity values are those of the entire dataset, as reported in Figure~\ref{fig:pareto_d1_and_stability}}}
    \label{fig:best_expl_d1}
\end{figure}

% This last combination proves unstable also for BA grid house. 
To conclude the analysis on \datasetgcone, Figure \ref{fig:best_expl_d1} shows a prototypical explanation for each GNN, paired with its best explainer as identified by the highest combination of the two metrics.
Overall, we can assert that each GNN can be explained fairly well if the concept is a simple subgraph of the network. As anticipated in subsection \ref{subsec:rq1}, \gin is the hardest to explain, while the best explanations are obtained with gradient, perturbation\rev{, and generation}-based explanations producing edge masks. 

\paragraph{\textbf{\datasetgctwo}} 
In this dataset class 0 contains either a house or a grid, while class 1 
contains both of them. Thus just identifying the presence of a simple pattern (a grid or a house) is insufficient for discrimination, and the GNN needs to learn how to combine them.
In addition to investigating compositionality, this dataset can also help investigating an aspect that we name \emph{laziness}. Namely, a network can address a binary classification problem by learning patterns characterizing only one of the two classes and predicting the other one when these patterns are absent.

\begin{figure}[ht]
    \centering
    \includegraphics[width=\textwidth]{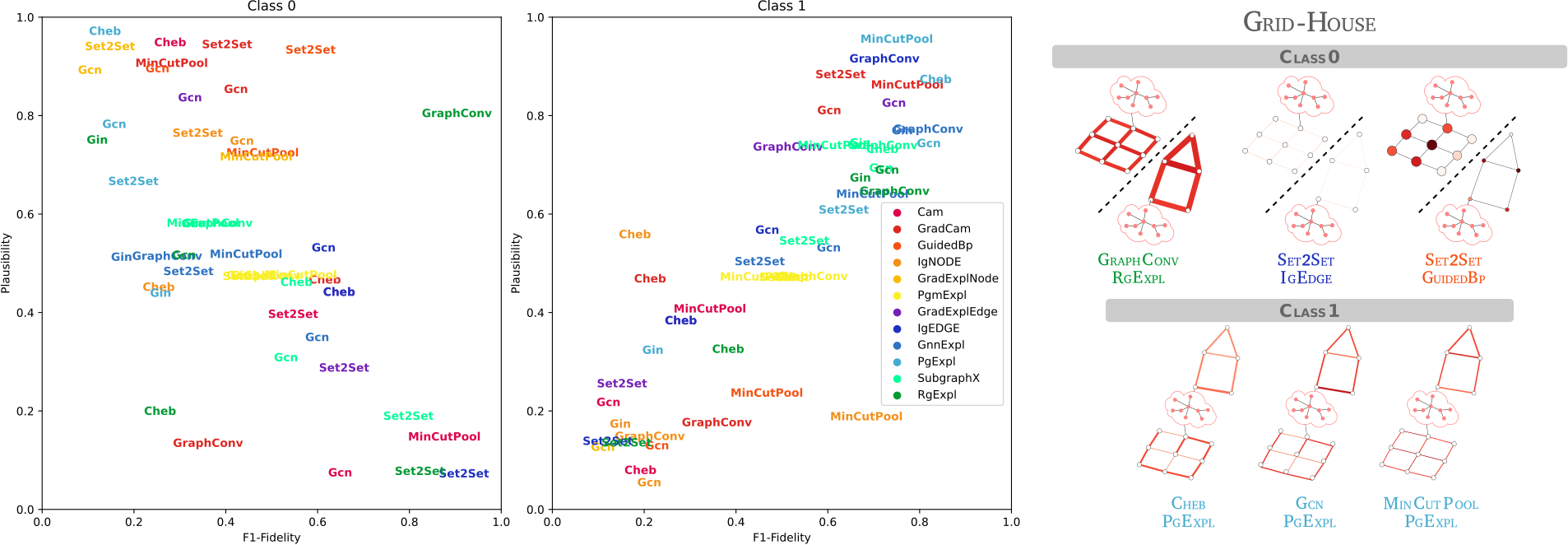}
    \caption{\rev{Left: Fidelity and plausibility achieved by all the model-explainer pairs when applied to \datasetgctwo, for class 0 (left) and class 1 (right). In each pair the name refers to the model, while the color identifies the explainer. Right: Stability of the explanations for the three top-performing model-explanation pairs, for each of the two classes. The colors identify important edges (dark red), and the edge thickness the variability of the importance in the dataset.}}
    \label{fig:pareto_d2_and_stabiliyt}
\end{figure}

As with \datasetgcone, we start with the comparative analysis of plausibility and fidelity for each GNN-explainer pair in the left of Figure~\ref{fig:pareto_d2_and_stabiliyt}, where in this case the results are reported for both class 0 (left panel) and class 1 (right panel). Here a remarkable difference can be observed between the two classes
since it is clear that for class 1 a linear correlation is present between the two metrics for each model-explainer pair, while for class 0 a high 
plausibility is associated with a low fidelity, and vice-versa, \rev{the only partial exception being the pair \graphconv-\rgexpl, which achieves a good balance between the two metrics.}
The best explanations are clearly those for class 1, confirming the laziness phenomenon explained above. This finding suggests that care should be taken in evaluating instance-based explanations, as the reason for predicting a certain class could lie in the absence of evidence in favor of the alternative one.

In more general terms, this result shows a substantial discrepancy between plausibility and fidelity, and it indicates that it is crucial to jointly consider both metrics to properly evaluate GNN explainability.

Moreover, for class 1 it is easy to see that  \mincut, \cheb and \gcn stay on the Pareto front when paired with \pgexpl, which is thus the best explainer in this case. 
In the top right of the figure, the majority of the colors are cold, corresponding to edge-based explainers.
In the case of class 0, instead, it can be verified that the high-plausibility model-explanation pairs (top left corner in the figure) all 
capture either the house or the grid, but no other structure in the graph.
On the other hand, the high-fidelity ones (bottom right, i.e., \setset plus \igedge, \mincut plus \cam, and \gcn plus \cam) have explainers which capture 
both part of the motif (grid or house), and part of the BA graph. This confirms the unreliability of the explanations extracted from the ``default'' class.

In terms of stability, right of Figure \ref{fig:pareto_d2_and_stabiliyt} shows the average explanations for the three top-performing explainers for each of the two classes. 
Regarding class 0, we can see that both \rev{\graphconv-\rgexpl} and \setset-\guidedbp identify both the grid and the house, and this reflects the high plausibility shown in the left panel in Figure \ref{fig:pareto_d2_and_stabiliyt}. 
On the other hand, \igedge does not capture neither the grid nor the house, but it captures part of the BA component, and this explains the low plausibility 
and high fidelity. %These observations confirm the insights obtained from Figure~\ref{fig:pareto_d2_and_stabiliyt}. 

For class 1 the situation is more uniform, since the three optimal pairs all achieve a rather similar stability. Indeed, the two motifs are colored in dark 
red, and the edges are rather thin. A partial exception is the case of \cheb-\guidedbp, where the the house motif has a lighter color (and thus a smaller 
average importance across the dataset), and the grid has some variability over the edges' thickness, indicating a larger standard deviation in their importance.

\begin{figure}[ht]
    \centering
        \includegraphics[width=0.99\textwidth]{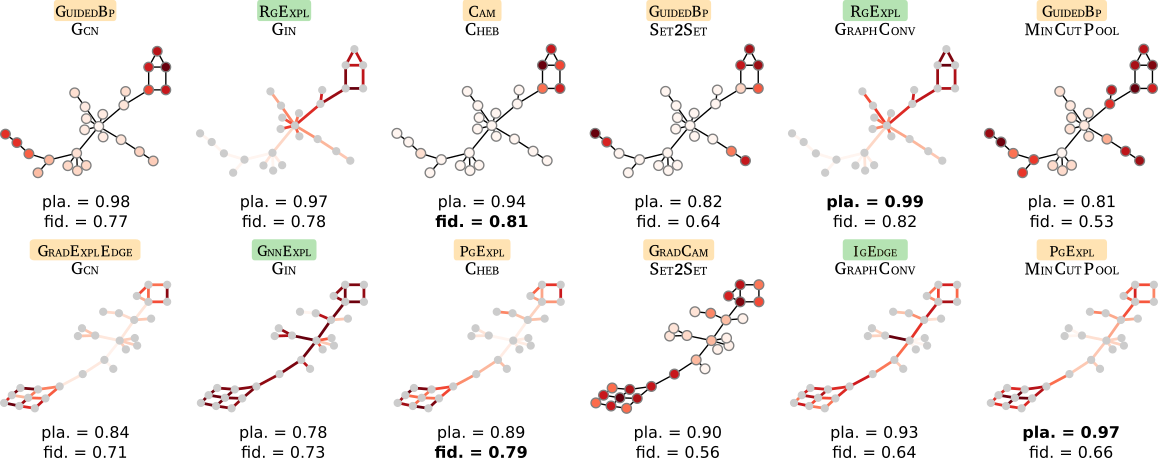}    
        \caption{\rev{Examples of explanations provided for each model and its highest plausibility explainer, when applied to a random sample from \datasetgctwo. Each row shows the results for one of the two classes. The plausibility and fidelity values are those of the entire dataset, as reported in Figure~\ref{fig:pareto_d2_and_stabiliyt}.}}
    \label{fig:best_expl_d2}
\end{figure}

Finally, in Figure \ref{fig:best_expl_d2} we show an example of the explanation provided by the best explainer associated to each model, for both class 0 and class 1. These explanations are visualized with the same color scale used for the previous dataset, and computed for a graph randomly selected from each class in \datasetgctwo. The high plausibility pairs are clearly visible since they identify the house for class 0 (\cheb-\pgexpl and \mincut-\pgexpl) and both structures and their connection for class 1 (\mincut-\pgexpl). 
Interestingly, even the high fidelity ones are easy to spot (\gradcam-\setset for class 0, \cheb-\pgexpl for class 1), and this indicates a good agreement between the expected and learned concepts.

% Even if in class 0, two models (\cheb and pooling) achieve an high plausibility, the fidelity is quite low, for this reason, we can asses that the class 0 is more difficult to explain with respect to the class 1. Probably, GNNs easily learn complex concepts (grid plus house) than multiple but simpler concepts. Class 1 is well explained in almost all the GNNs. Once again, for both classes the best explainer is gradient based.\\
% The right part of figure \ref{fig:stanbility_d2} shows the same analysis on class 1. It is evident that in the three GNNs with PGEx are well explained. In particular we can even infer that the pooling mechanism has the lowest standard deviation on the prediction of edge importance.

\paragraph{\textbf{\datasetgcthree}}
This dataset has three classes: each graph is obtained starting from an Erd{\H o}s–R{\'e}nyi (ER) graph, to which we attach one (class 0), two (class 1), three or four (class 2) stars. This dataset thus evaluates GNN explainers on concepts involving counting. 

In order to enable this motif-counting capability, we used for all architectures a sum-based global aggregation in place of a mean-based one, which would prevent the networks from being able to count occurrences of motifs. Indeed, mean-based versions of all models were trained as well, yet without reaching the thresholded train accuracy of $95\%$ (see Section \ref{sec:experimental_setting}).

\begin{figure}[ht]
    \centering
    \includegraphics[width=0.99\textwidth]{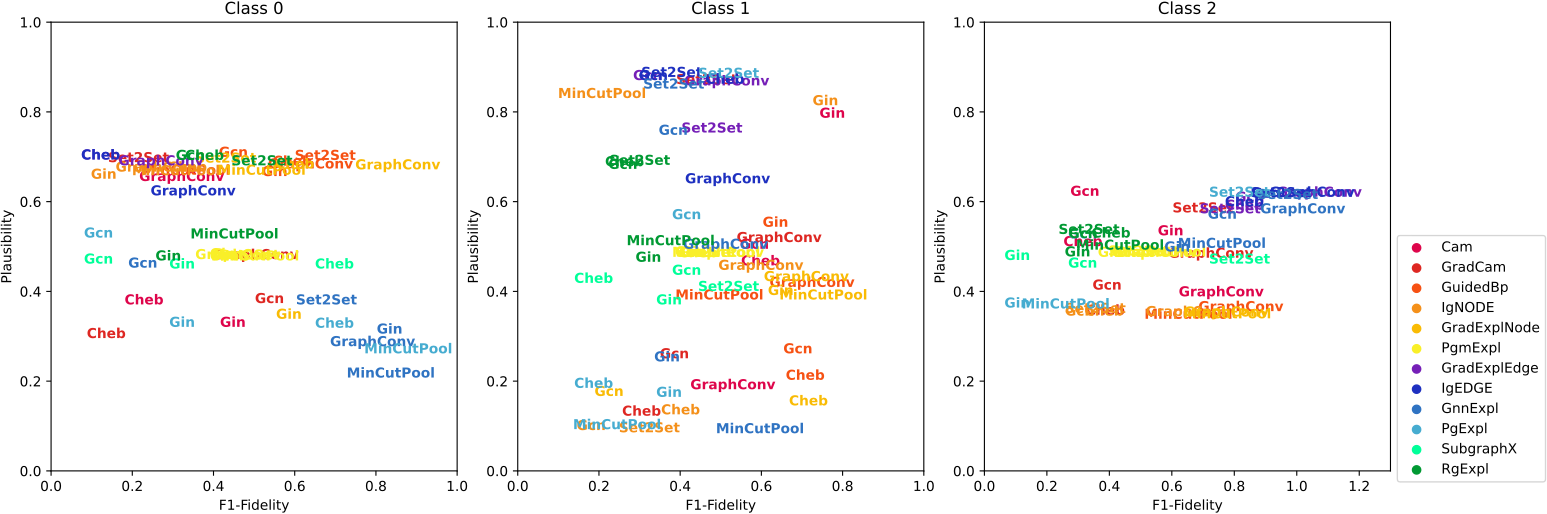}
    \caption{\rev{Fidelity and plausibility achieved by all the model-explainer pairs when applied to \datasetgcthree, for class 0 (left) and class 1 (right). In each pair the name refers to the model, while the color identifies the explainer.}}
    \label{fig:pareto_d3}
\end{figure}

The behaviors of each model-explainer pair is visualized in Figure \ref{fig:pareto_d3}, with a panel for each class. 

Class 0 is explained well by explainers producing node masks, and especially by \gradexp on \graphconv. However, all the model-explanation pairs have a fairly limited plausibility when compared with the other datasets, and none of them has a plausibility larger than $0.8$.

For class 1 the best combined performances are achieved by \gin explained by \ignode and \cam, even if also in this case the fidelity is almost always below 0.8, with instead a quite high plausibility. This indicates that these explainers may be good in identifying the expected ground truth (the stars), thus obtaining high plausibility, but that the presence of these motifs alone may be insufficient for predicting the class, when not complemented by a non-negligible part of the ER graph.

The right panel of Figure \ref{fig:pareto_d3} shows the results for class 2. Also in this case the plausibility is very limited across the entire set of models and explanations. Looking at the fidelity alone, it is evident that the best performing explainers are those that produce an edge mask. In this case, a direct inspection of the explanations shows that these explainers not only capture the stars, but also paths connecting them.
This makes the explanations farther away from the expected ground truth (thus obtaining low plausibility), but apparently provide better masks for the model to correctly identify the class.

From these results, it seems clear that there may be some clash between the identification of the motifs (high plausibility) and the fact that these motifs are sufficient to predict the class (low fidelity).
%the identification of patterns connecting them and covering parts of the BA component. 
To try to explain this behavior, we evaluated all the trained networks on the exact expected ground truth explanations, i.e., four graphs obtained by one, two, three or four disconnected stars.
For all these graphs, all the models predict class 2 (i.e., that with three or four stars attached to an ER graph), thus completely missing the correct classification for class 0 and class 1. These stars are thus clearly insufficient to characterize what the networks use for prediction. This result reveals the difficulty in defining the plausibility of an explanation, even in the presence of explicitly defined ground truths.

\begin{figure}[ht]
    \centering
        \includegraphics[width=0.99\textwidth]{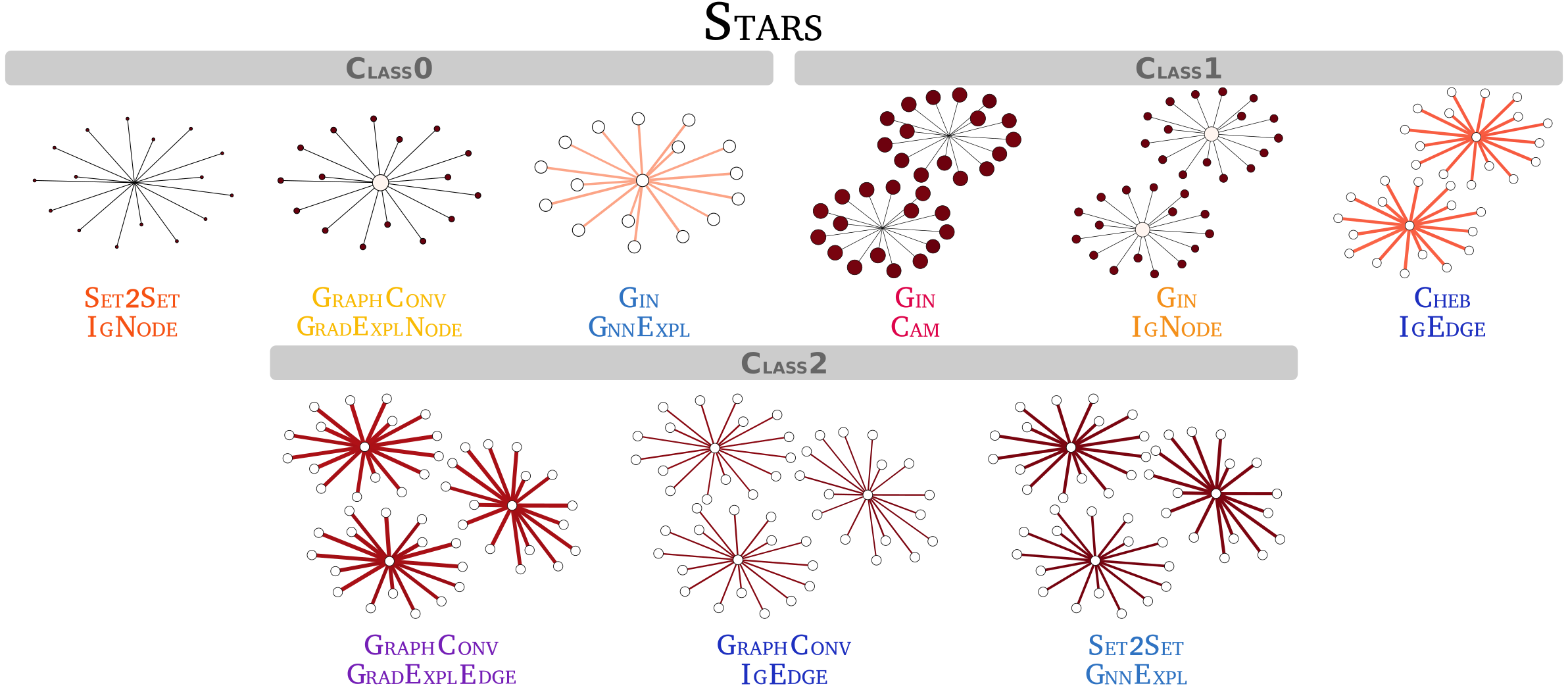}
    \caption{\rev{
Stability of the explanations for the three top-performing model-explanation pairs, for each of the three classes. The colors identify important edges (dark red), and the edge thickness the variability of the importance in the dataset.}}
    \label{fig:stanbility_d3}
\end{figure}

Figure \ref{fig:stanbility_d3} shows the stability of the explanations when restricted to the stars which identify the three classes. We remark that both node- and edge-based explainers are visualized, and thus the coloring and thickness may apply to either the nodes or the edges.

In class 0, explainers that produce a node mask (\ignode on \setset, and \gradexp on \graphconv) capture really well the star (intensity of the color) with a low standard deviation (size of the nodes), i.e., they are extremely stable. Remarkably, in the case of \graphconv the explainer gives no importance to the center of the star, while for \setset it does. 
On the other hand, \gnnexpl on \gin identifies the star but with a low intensity, and this explains the corresponding low plausibility observed in Figure~\ref{fig:pareto_d3}.

\begin{figure}[ht]
    \centering
        \includegraphics[width=0.99\textwidth]{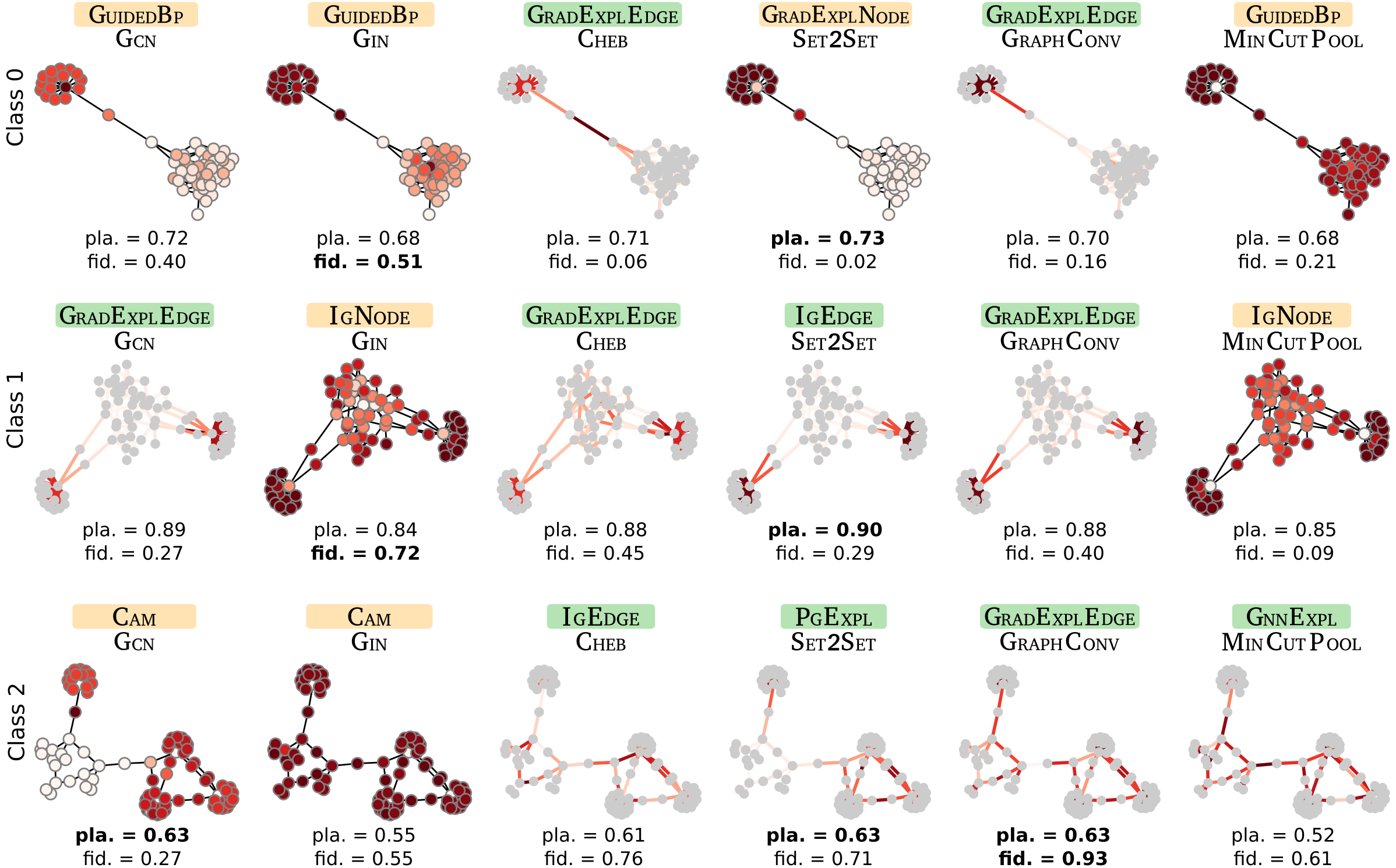}
    \caption{Examples of explanations provided for each model and its highest plausibility explainer, when applied to a random sample from \datasetgctwo. Each row shows the results for one of the two classes .The plausibility and fidelity values are those of the entire dataset, as reported in Figure~\ref{fig:pareto_d3}.}
    \label{fig:best_expl_d3}
\end{figure}

For class 1, all the three best model-explainer pairs capture the stars, even if \cam has a higher standard deviation. 
Surprisingly, \cam gives to the central node of the star always the highest importance (dark color and small size), while \ignode does the opposite (light color and large size). It is interesting to observe that both masks are reasonable ways to identify the two stars, and it may be difficult to recognize one or the other as the actual correct explanation. This difference is made even more interesting by the fact that the two explanations apply to the same model (\gin), and that the resulting fidelity and plausibility are essentially the same (see Figure \ref{fig:best_expl_d3}, central panel).

Finally, for class 2 all explainers capture the stars with high importance and low standard deviation, without significant differences.

Figure \ref{fig:best_expl_d3} reports, for each model and each class, an example explanation  provided by the corresponding best explainer. All masks are computed on a graph randomly drawn from the dataset for each class. 
% the examples of the  best explainer applied to each class and GNNs. As anticipated, in class 0 the star is captured, but only those explainer that capture also part of the ER graph have an higher fidelity. In class 1, all explainers producing an edge mask capture the two stars really well, and for the same reason of class 0 they have low fidelity. On the other hand, ig (node) that produces node mask have the highest fidelity. Finally, in the class 2 it seems that all the explainers producing edge mask not only captures the stars but they also try to connect them with paths into the ER graph. This paths allow the GNN to increase the magnitude of the signal, improvising the fidelity of the explanation.

\paragraph{\textbf{\datasetgcfour}} In this dataset we would like to explore how the node features affect the explanations. 
In particular, both classes have a BA base graph with attached one, two, or three houses, and each node has a random color, except for one house that has all nodes colored blue (class 0) or green (class 1). 
% Thus, the dataset structure does not give any a-priori clue on which class should be the most explainable, because identifying green or blue houses is essentially the same.
In particular, the two classes represent essentially the same type of pattern, and thus we have no reason to expect that a \emph{lazy} GNN should prefer one or the other, and consequently no way to anticipate a difference in their explainability. \rev{By laziness, specifically, we mean the GNN focusing on learning the discriminant features of class 0 (1) while predicting the remaining class 1 (0) by the absence of such discriminant feature, without properly modeling the underlying discriminant feature of the latter class.}

\begin{figure}[ht]
    \centering
    \includegraphics[width=0.8\textwidth]{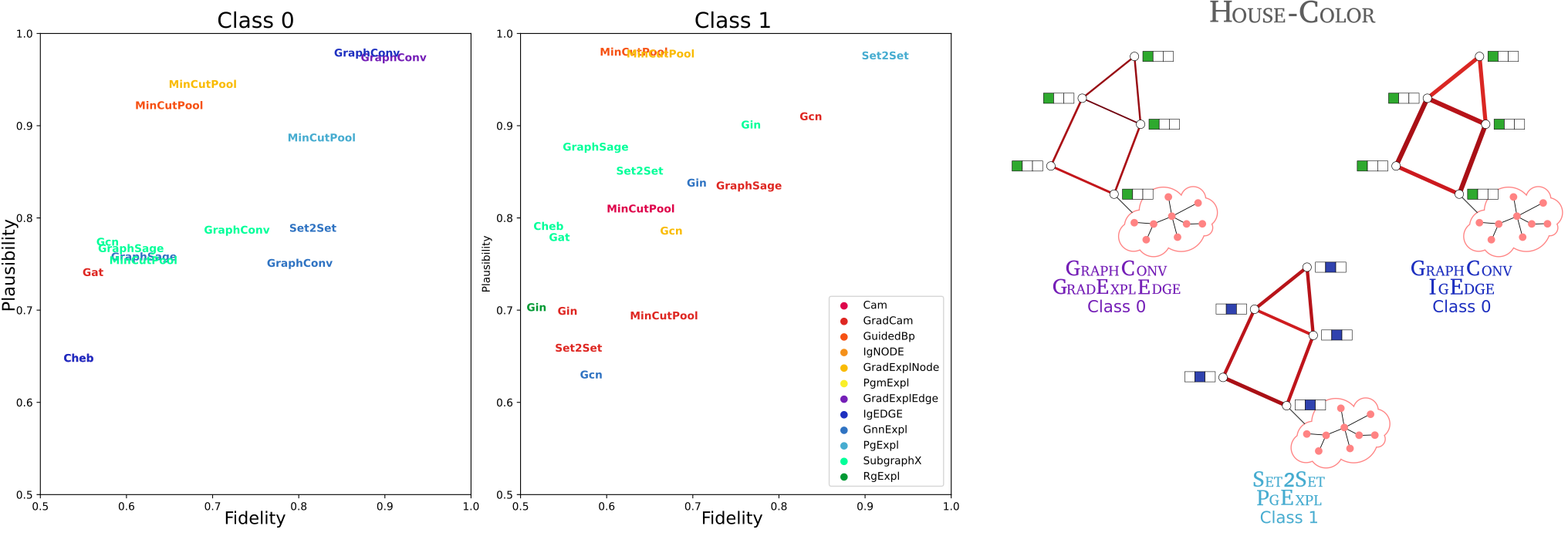}
    \caption{\rev{Left: fidelity and plausibility achieved by all the model-explainer pairs when applied to \datasetgcfour, where the two classes (0 and 1) are represented in the left and right panel. In each pair the name refers to the model, while the color identifies the explainer. The plot is limited to metrics larger than $0.5$ to simplify the visualization. Right: stability of the explanations for the three top-performing model-explanation pairs. The colors identify important edges (dark red), and the edge thickness the variability of the importance in the dataset.}}
    \label{fig:pareto_d4_and_stability}
\end{figure}

\begin{figure}[ht]
    \centering
        \includegraphics[width=0.99\textwidth]{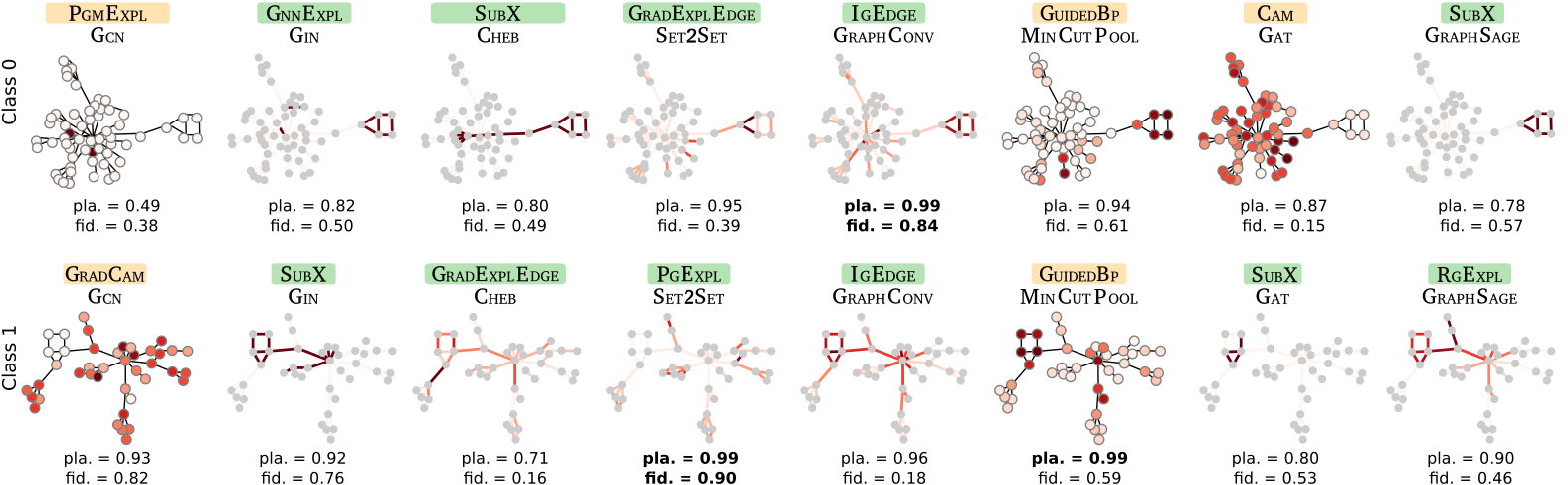}
    \caption{\rev{Examples of explanations provided for each model and its highest plausibility explainer, when applied to a random sample from \datasetgctwo. Each row shows the results for one of the two classes. The plausibility and fidelity values are those of the entire dataset, as reported in the right of Figure~\ref{fig:pareto_d4_and_stability}.}}
    \label{fig:best_expl_d4}
\end{figure}

The comparison of fidelity and plausibility in shown in the left panel of Figure \ref{fig:pareto_d4_and_stability}. To avoid visualizing too many results, the figure is limited to those pairs with both plausibility and fidelity greater than $0.5$. Also in this case a linear correlation between the two metrics is clearly present, even if there is a group of outliers with high plausibility and low fidelity (the \mincut models with explainers \gradexp and \guidedbp, for the two classes). In this case, a closer inspection of the single explanations reveals that the explanation masks are able to capture well the colored house, thus achieving a large plausibility, but they contain also a significant amount of noise that spoils the fidelity. An example of this behavior can be observed in the examples in Figure~\ref{fig:best_expl_d4}. Moreover, for this dataset the optimal pairs are based on \graphconv and \setset, with explanations \sal, \igedge, \pgexpl. In particular, edge-based explainers are the top performing ones also in this case.
Another remarkable aspect is the fact that for each model-explainer pair, only one between class 0 and class 1 achieves high scores (i.e., both high fidelity and high accuracy). This suggests again a {\em lazy} behavior for the GNN even in the case in which both classes are equally easy to explain (Figure~\ref{fig:pareto_d4_and_stability}), where some GNNs are learning to characterize class 0 and others class 1, but none modelling both.

% This may once again be motivated by GNN's \emph{laziness}, i.e., the fact that, in order to obtain a high accuracy, a GNN may just learn to identify patterns of one of the two classes, and simply assigning the opposite label to examples without this pattern. If this is the case, an explainer would consequently be able to provide an explanation only for one of the two classes, as is the case here.

In the right panel of Figure \ref{fig:pareto_d4_and_stability} we report the average explanation for the three top-performing pairs, which all comprise edge-based explainers. Differently from the previous datasets which had no node feature, the figure visualizes also the color of the node feature close to each node. 
In all cases the explanation is very strong across the dataset (dark red color), with a minimal standard deviation for \graphconv-\sal, and a maximal one for \graphconv-\igedge.

Examples of explanation masks computed on random samples from \datasetgcfour are reported in Figure \ref{fig:best_expl_d4}, where again we show for each class and for each model only the corresponding best performing explainer. Once again, explainers that produces edge features are more effective in terms of plausibility and fidelity. It is worth mentioning that there are no explainers having high fidelity in both classes, arguing the thesis that GNNs are lazy and learn only one class. On the other hand, at least one class for each GNN can be explained by an explainer. As one may expect, the only exception is the \gat architecture, because the attention mechanism strongly operate on node features, and none of the studied explainers operate on node features. 

\subsection{Node classification}

\subsubsection{RQ1: How does the architecture affect the explanations?}\label{subsec:rq1_nc}
To answer this question, we summarize the performance of the different architectures across all explainers by using the aggregation mechanism described in Section~\ref{sec:aggregation}. 
The plausibility and fidelity results for each of the two datasets \datasetncone and \datasetnctwo, and the overall results are reported in Table~\ref{tab:rq1_all_nc}.

The most explainable architectures are \sage for the plausibility metric (explained by \gradcam), and \gin for fidelity (explained by \gnnexpl). They are clearly identified as such by \textbf{RQ1.1} and \textbf{RQ1.2}, i.e., both as single best performing architectures and mean best performing ones\rev{, and both at the aggregate and single-dataset levels.} 
% The same holds for the single dataset as well, yet with a small variability suggesting that the answer is quite solid and stable.
The difference between the two metrics indicates that for \sage one gets explanations that better resemble the expected ground truth, and are thus closer to a human-level explanation. On the other hand, for \gin one obtains explanations with a higher fidelity, meaning that they better capture the actual patterns that the trained \gin uses in building its decisions. 

\rev{The hardest architecture to explain on average (\textbf{RQ1.3}) is identified to be \cheb, for both metrics and both at the aggregate and at dataset levels. The only exception is \datasetnctwo with \sage.}

\begin{table}[ht]
\small
\centering
\begin{tabular}{c|ccc|ccc}
&\multicolumn{3}{c|}{Plausibility} & \multicolumn{3}{c}{Fidelity}\\
&\textbf{All} & \datasetncone& \datasetnctwo &\textbf{All} & \datasetncone& \datasetnctwo \\
\hline 
{RQ1.1} & \textbf{\sage} & \sage & \gcn   & \textbf{\gin}  & \gcn      & \gin\\
{RQ1.2} & \textbf{\sage} & \rev{\sage}  &\sage  & \textbf{\gin}  & \gcn/\gin & \gin\\
{RQ1.3} & \textbf{\cheb} & \cheb & \cheb  & \rev{\textbf{\cheb}} & \sage     & \rev{\cheb} \\
\end{tabular}
\caption{\rev{Experimental answer to \textbf{RQ1} for node classification. The table shows the top-ranking architecture with respect to each subquestion \textbf{RQ1.1}, \textbf{RQ1.2}, \textbf{RQ1.3}, both for the single datasets \datasetncone and \datasetgctwo, and overall. The rankings are computed with respect to the plausibility and the fidelity metrics.}}
    \label{tab:rq1_all_nc}
\end{table}

\begin{figure}[ht]
     \centering
     \includegraphics[width=0.99\textwidth]{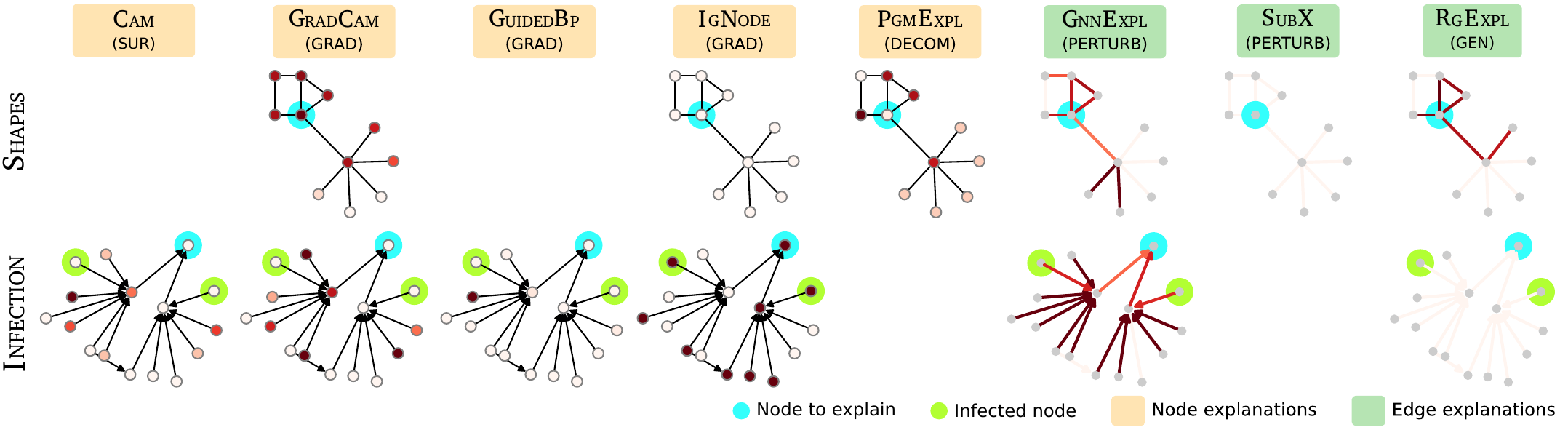}     
     \caption{\rev{Explanation masks (node- or edge-based) computed by the different explainers on the predictions of \sage on \datasetncone and \datasetnctwo. Each row visualizes the mask computed for a given random graph from each dataset. For each dataset, only the explainers which passed the filtering procedure are shown.}}
     \label{fig:rq1_nc}
 \end{figure}

Figure~\ref{fig:rq1_nc} shows examples of explanation masks computed by the different explainers on \sage, which is the architecture with the highest average plausibility (RQ1.2), over all explainers which passed the filtering procedure (Section \ref{sec:aggregation}).
For each dataset we visualize a sample node and its 2-hop neighborhood (directed in the case of \datasetnctwo, i.e., the set of nodes from which the ego node is reachable following two edges).
For both datasets we focused on nodes with label 1, i.e., the base of the roof in \datasetncone, and a node at distance one or two from an infected node in \datasetnctwo. 
This means that their associated ground truths are the entire house in \datasetncone, and a path of length one or two in \datasetnctwo (in this case multiple possible ground truth may exist, and only the one with the highest plausibility is considered for the computation of the metric).
Despite \sage being the model with the highest plausibility, it is clear from these examples that there is a high variability across nodes and explainers.
Moreover, for \datasetncone the house is well highlighted by \gradcam, \pgmexpl, \gnnexpl, \rev{and \rgexpl}, but the explanation additionally includes spurious nodes and edges with equally large importance. A similar situation can be observed for \datasetnctwo. In both cases, a finer inspection (see Section~\ref{subsec:rq3_nc}) reveals that the plausibility of \sage, despite being the highest on average over classes and explainers, is fairly limited for class 1: the values are below $0.6$ for class 1 in \datasetncone (Figure~\ref{fig:pareto_shapes}), and with high spread and mean around 0.6 for class 1 in \datasetnctwo (Figure~\ref{fig:pareto_infection}). We remark that for both datasets, other classes achieve an higher plausibility.

In any case, this variability across the same dataset has to be taken into account, and a word of caution is in order when using instance-based explanations, since the single masks are those which are actually utilized to inspect the model and try to extract information on its decisions. Their usefulness may vary largely from node to node.

\subsubsection{RQ2: How do explainers affect the explanations?}\label{subsec:rq2_nc}

Table~\ref{tab:rq2_all_nc} shows that, while \gradcam provides the single best explanation in terms of plausibility, \gnnexpl is the best one both in terms of mean performances (\textbf{RQ2.2}) for both metrics, and maximal performances (\textbf{RQ2.1}) for plausibility, and can be thus considered as the overall best performing explainer in this setting. It is worth remarking that this is a perturbation- and edge-based explainer.

When looking at average performances for the entire category the situation is instead different, and it turns out that gradient- (\textbf{RQ2.3}) and node-based (\textbf{RQ2.4}) explainers are to be preferred\rev{, with a uniform consensus across metrics and aggregation levels}. In particular, \pgexpl (perturbation- and edge-based) has very poor performances and this lower the aggregates scores of the corresponding categories.
This fact is possibly justifiable since we are dealing with node classification tasks, where local explanations like those provided by gradient based explainers may be more effective. Moreover, since meaningful explanation should be limited to a node's $k$-hop neighborhood, it is reasonable to expect that highlighting single nodes instead of edges is sufficient to explain the decision. In particular, the node itself is already defining the relevant neighborhood, and we may thus expect that the additional missing information required to elaborate a decision is given by the nodes' labels.

\begin{table}[ht]
\small
\centering
\begin{tabular}{c|ccc|ccc}
&\multicolumn{3}{c|}{Plausibility} & \multicolumn{3}{c}{Fidelity}\\
&\textbf{All} & \datasetncone& \datasetnctwo &\textbf{All} & \datasetncone& \datasetnctwo \\
\hline  
{RQ2.1} & \textbf{\gradcam} & \gradcam  & \gradcam & \textbf{\gnnexpl} & \pgexpl  & \gradcam\\
{RQ2.2} & \textbf{\gnnexpl} & \gradcam  &\gnnexpl  & \textbf{\gnnexpl} & \gnnexpl & \gnnexpl\\
{RQ2.3} & \textbf{Grad}     & Grad      & \rev{Grad}     & \textbf{Grad}     & \rev{Grad}     & Grad \\
{RQ2.4} & \textbf{Node}     & \rev{Node}     & Node     & \textbf{Node}     & \rev{Node}    & Node \\
\end{tabular}
    \caption{\rev{Experimental answer to \textbf{RQ2} for node classification. The table shows the top-ranking explainer with respect to each subquestion \textbf{RQ2.1}, \textbf{RQ2.2}, \textbf{RQ2.3}, \textbf{RQ2.4}, both for the single datasets \datasetncone and \datasetgctwo, and overall. The rankings are computed with respect to the plausibility and the fidelity metrics.}}
    \label{tab:rq2_all_nc}
\end{table}

\begin{figure}[ht]
     \centering
     \includegraphics[width=0.9\textwidth]{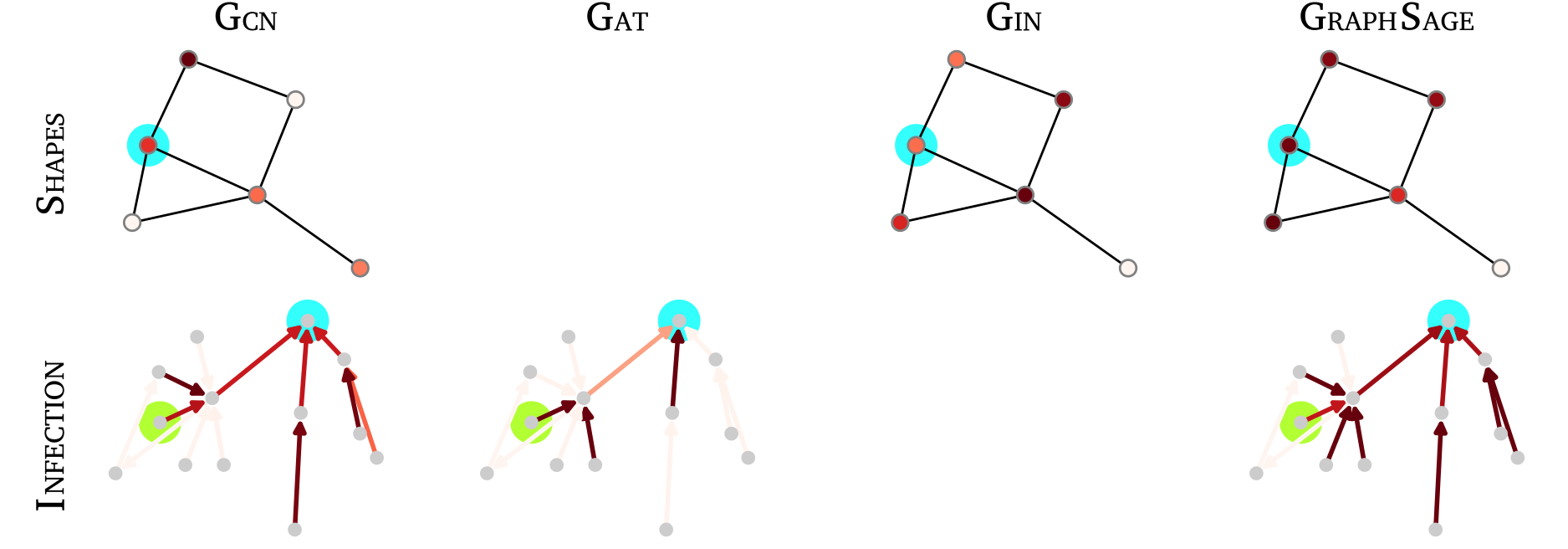}
     \caption{Explanation masks (node- or edge-based) computed on the predictions of \gnnexpl by the different explainers, both on \datasetncone and \datasetnctwo. Each row visualizes the mask computed for a given random graph from each dataset. For each dataset, only the explainers which passed the filtering procedure are shown.}
     \label{fig:rq2_nc}
 \end{figure}

In the same setting of Figure~\ref{fig:rq1_nc}, we report in Figure~\ref{fig:rq2_nc} examples of explanation masks obtained by \gnnexpl, which has the highest average plausibility over all architectures and classes - RQ2.2. Also here, architectures not passing the filtering step have been removed (Section~\ref{sec:aggregation}).
In this case a better localization of the explanations may be observed for \datasetncone for all models.
Indeed, all architectures have a plausibility above $0.8$ in class 1 in \datasetncone (Figure \ref{fig:pareto_shapes}).

\subsubsection{RQ3: How do different types of problems affect the explanations?}\label{subsec:rq3_nc}

We analyze the two datasets separately, in order to highlight the different aspects and problems that they represent.

\begin{figure}[ht]
    \centering
    \includegraphics[width=0.9\textwidth]{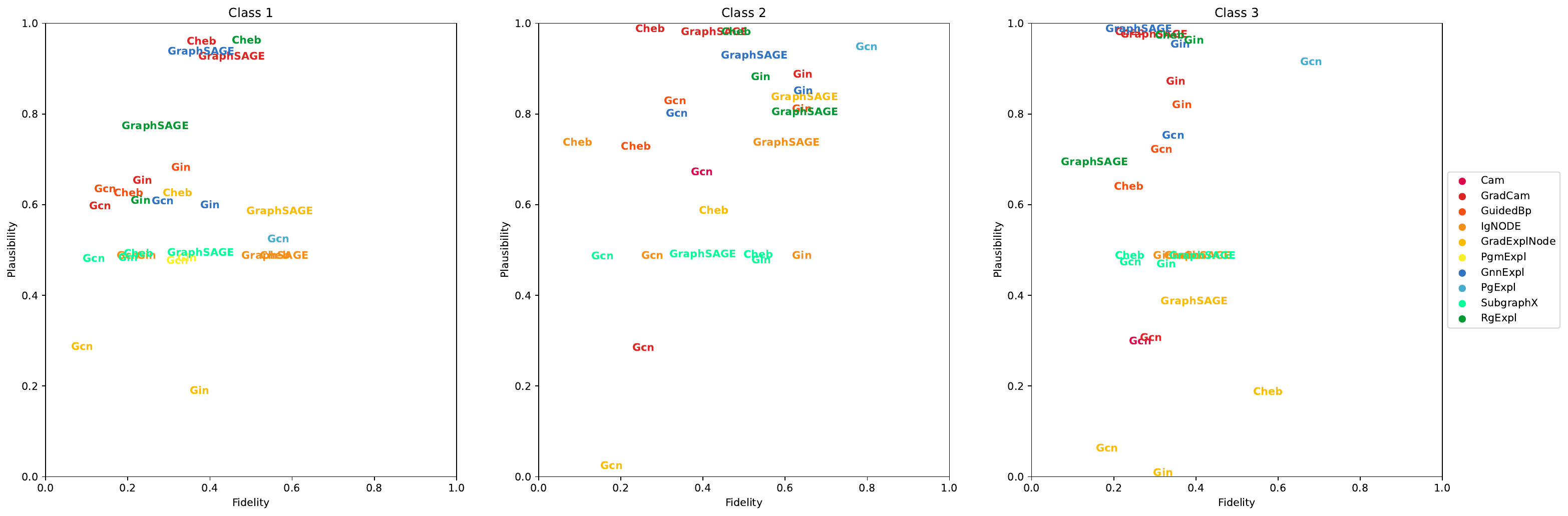}
    \caption{\rev{Fidelity and plausibility achieved by all the model-explainer pairs when applied to \datasetncone. In each pair the name refers to the model, while the color identifies the explainer. Class 0 is omitted from the visualization since it is less relevant for the discussion of the explainers.}}
    \label{fig:pareto_shapes}
\end{figure}

\paragraph{\textbf{\datasetncone}}
The comparative visualization of the plausibility and fidelity of each architecture-explainer pair is reported in Figure~\ref{fig:pareto_shapes}, where we visualize only the classes 1 to 3 (structural elements of a house, see Section~\ref{sec:dataset_nc}), while we omit class 0, which having no structure is less interesting from the point of view of explanations.

No clear correlation emerges between the two metrics, except partially for class \rev{2}, which is the only one with architecture-explainer pairs with both high fidelity and high plausibility.
Several pairs achieve very high plausibility (even close to $1$, and for all three classes), but always with low fidelity (values bounded by $0.6-0.7$ depending on the class).
The only outlier is the \gcn-\pgexpl pair for class 2 and 3, which has a plausibility of about $0.9$ and fidelity of $0.8$ (class 2), and $0.7$ (class 3).

The very high plausibility indicates that the explainers agree with the human definition of a ground truth, which is thus correctly defined to be the same for all classes (the house structure).
On the other hand, this in turn results in an observed low fidelity. This seems to suggest that it is enough to have a few house elements missing from the explanation (the plausibility is never equal to 1) to spoil the model predictions, and that indeed the entire motif has to be observed to compute a prediction.
It is possible moreover that an high plausibility is a proxy for a low sparsity of the explanation, which may thus miss some crucial edges. On the other hand, explanations with lower plausibility may be more spread, thus covering the entire house structure, even if with some spurious additional edge. This result highlights again that the sufficiency alone may be non very informative in some settings.

% This in turn results in the observed low fidelity: it is difficult for the explainers to identify an explanation which discriminates between the classes, which is confirming that the actual explanation is not discriminative enough for the trained models to compute their predictions, which is understandable if they coincide. T
% his result seems to indicate that this is a difficult to explain dataset, where high training and test accuracies are not easily motivated by the vast majority of the explainers, with the exception of \gcn with \pgexpl.

\begin{figure}[ht]
    \centering
    \includegraphics[width=0.99\textwidth]{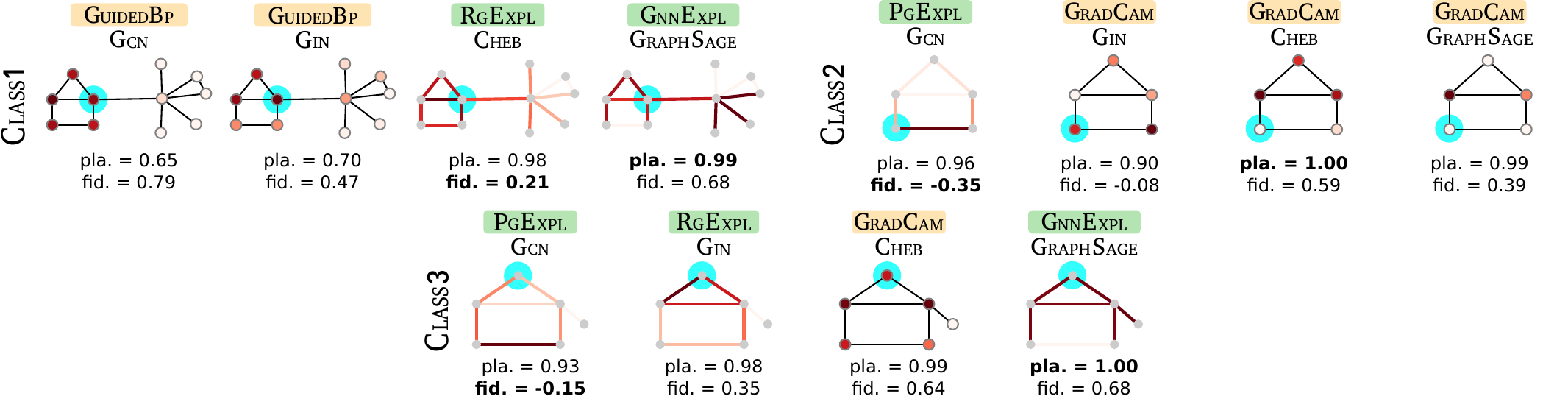}
    \caption{\rev{Examples of explanations provided for each model by its highest plausibility explainer, when applied to a random node from \datasetncone. Each row shows the results for one of the three house-structure classes.
The plausibility and fidelity values are those of the entire dataset, as reported in Figure~\ref{fig:pareto_shapes}.}}
    \label{fig:best_shapes}
\end{figure}

Regarding the well-performing \gcn-\pgexpl pair, a sort of two-class laziness can be observed: the explanations are good for class 2 and 3, while they have fairly limited plausibility and fidelity (around $0.5$) for class 1. This may suggest that the nodes in class 1 (the middle of the house) are classified by \gcn as not being in any other class. Examples in Figure~\ref{fig:best_shapes} shows explanation masks returned for each model by its explainer which reaches the highest plausibility. A significant overlapping between the explanation mask and the house structure (the ground truth) can be observed, even if in most cases the mask is at least partially spread also over other nodes.

\begin{figure}[ht]
    \centering
    \includegraphics[width=0.9\textwidth]{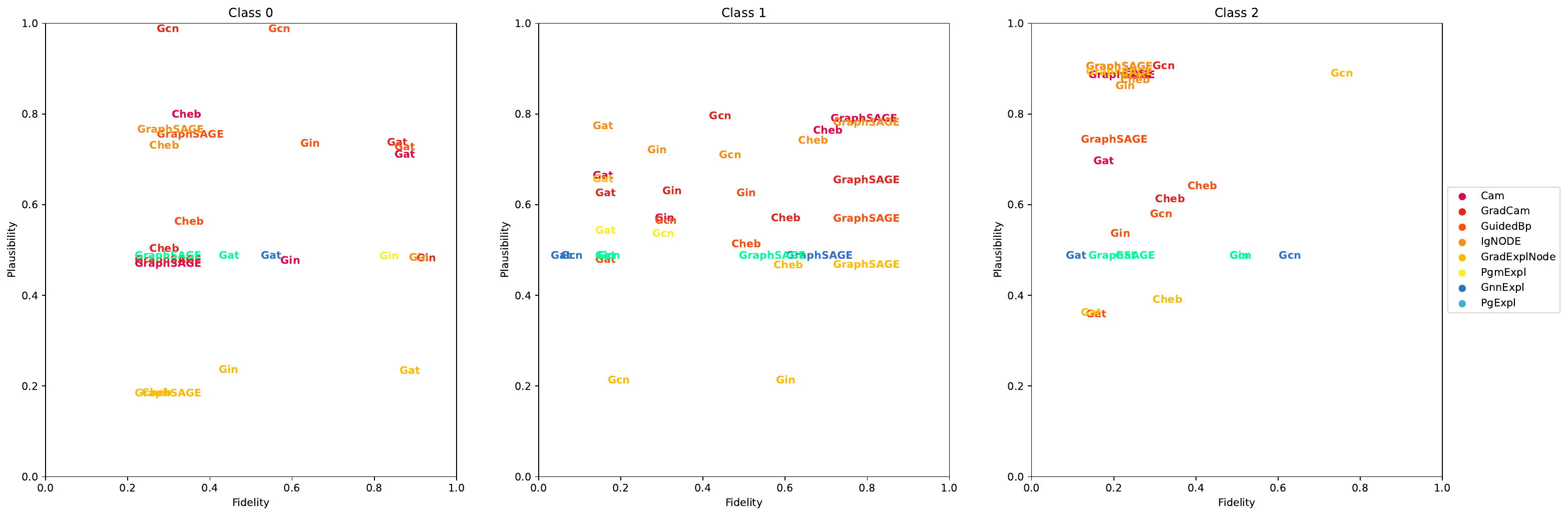}
    \caption{\rev{Fidelity and plausibility achieved by all the model-explainer pairs when applied to \datasetnctwo. In each pair the name refers to the model, while the color identifies the explainer.}}
    \label{fig:pareto_infection}
\end{figure}

\paragraph{\textbf{\datasetnctwo}}
The pairs of models and explanations are shown in Figure~\ref{fig:pareto_infection} according to their plausibility and fidelity on the three classes. 
We first remark that, out of the two edge-based explainers considered here, only \gnnexpl passes the filtering step (Section~\ref{sec:aggregation}), more specifically \gnnexpl with \gcn (for class 1,2) and with \gat and \sage (for all classes). Also in this case no correlation appears among the two metrics, while there are several pairs that reach a high value of only one of the two metrics. 
It is again possible that even relatively high values of the plausibility comprise explanations where some relevant part of the ground truth is missing, thus achieving low fidelity. For this dataset the necessity of covering the entire motif is even more clear than for \datasetncone, since here single nodes missing from an explanation may break the minimal paths connecting the node to an infected one. Exceptions are \sage-\gradcam and \sage-\ignode for class 1, and especially \gcn-\gradexp for class 2, which have both metrics above values of $0.8$.

With the exception of \gcn, class 2 exhibits a clear lack of explainability, indicating a preference of GNN architectures in focusing on class 0 and 1. This laziness is perfectly sensible, since it is much easier to model class 2 in terms of a lack of paths from nearby infected nodes (i.e., the negation of the other classes) than by trying to characterize all possible longer paths from an infected node.
%is a kind of "counterfactual one, as it is defined by a lack of paths from connected nodes. 
The exception to this pattern is \gcn-\gradexp, which has very high scores for class 2, but is instead significantly unexplainable for class 1 and class 0, where it is below the threshold for visualization. This is reasonable since \gcn has a simple message passing - aggregation mechanism, which may facilitate the identification of lack of infected nodes in the $2$-hop neighborhood (since the infected status is the node feature itself, see Section~\ref{sec:dataset_nc}), even if \sage has a completely analogous functioning mechanism, and it achieves nevertheless poorer results. At this stage, we are not able to identify the actual mechanism that make their performances so different.

\begin{figure}[ht]
    \centering
    \includegraphics[width=0.9\textwidth]{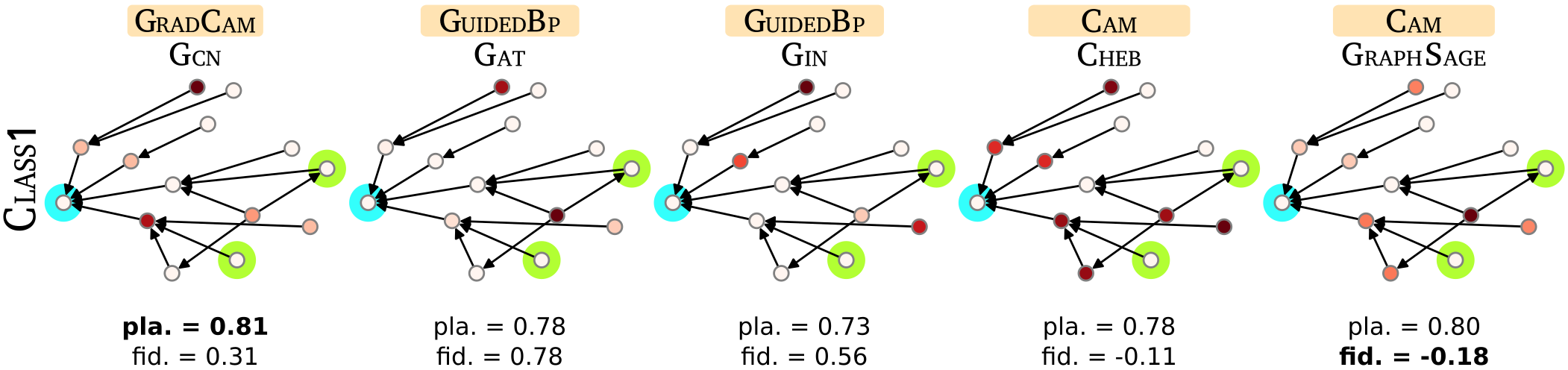}
    \caption{Examples of explanations provided for each model by its highest plausibility explainer, when applied to a random node from \datasetnctwo. Class 0 and class 2 are omitted since their visualization does not provide significant insights. The plausibility and fidelity values are those of the entire dataset, as reported in Figure~\ref{fig:pareto_infection}.}
    \label{fig:best_infection}
\end{figure}

Example of explanations are shown in Figure~\ref{fig:best_infection} for each model paired with the explainer which explains it with the highest plausibility. 
We show only class 1, since class 0 is trivial to visualize (the explanation is the actual node), while for class 2 the local networks happen to have too many edges and their visualization is not clear enough. We thus omit both since their visualization does not provide significant insights. In this case, it is difficult to observe a good accordance between the mask and the ground truth (a path connecting an infected node with the node to be explained). This happens despite the relatively high plausibilities (Figure~\ref{fig:pareto_infection}), and demonstrates again that care should be taken in considering single-instance explanations for the interpretation of a model's prediction.

\begin{table}[h!]
\small
\begin{tabular}{c c c }
\toprule
 & \multirow{2}{*}{\Large \textbf{Graph classification}} &  \multirow{2}{*}{\Large \textbf{Node classification}} \\
 & \\  
 \midrule
 \multicolumn{3}{l}{\multirow{2}{*}{\textit{\textbf{\normalsize RQ1: How does the GNN architecture affect the explanations?}}}}\\ 
 \multicolumn{3}{l}{} \\
 \midrule
 
 \cellcolor[HTML]{EFEFEF}RQ1.1& 
     \cellcolor[HTML]{EFEFEF}\begin{tabular}[c]{@{}c@{}}
        \graphconv excels in both plausibility  \\
        and fidelity, achieving the highest\\
        aggregated scores.
     \end{tabular} &  
 \multirow{-2}{*}{
    \begin{tabular}[c]{@{}c@{}}
        \sage provides explanations closer to the \\
        expected ground truth, akin to human-level\\
        explanations, while \gin attains elevated\\
        fidelity by effectively encapsulating the\\
        patterns acquired by the \gin model during\\
        its training.
    \end{tabular}} \\
%\\ \cline{0-1}

 RQ1.2 &  
    \begin{tabular}[c|]{@{}c@{}} 
        \gcn is the easiest architecture to \\
        explain, possibly due to its straightforward\\
        message-passing model.
    \end{tabular} \\ % \hline

 \cellcolor[HTML]{EFEFEF} RQ1.3 &  
     \cellcolor[HTML]{EFEFEF} \begin{tabular}[c]{@{}c@{}}
         \gin is the hardest to explain, which is\\
         surprising given its rather simple \\
         aggregation strategy. 
    \end{tabular} &  
    \cellcolor[HTML]{EFEFEF} \begin{tabular}[c]{@{}c@{}}
        \cheb is the hardest to explain due to its broader\\
        receptive field, aggregating information from more\\
        distant nodes compared to other networks.
        
    \end{tabular} \\
 %\midrule
  \midrule
 \multicolumn{3}{l}{\multirow{2}{*}{\textit{\textbf{\normalsize RQ2: How do explainers affect the explanations?}}}}\\ 
 \multicolumn{3}{l}{} \\ 
 \midrule
 
 \cellcolor[HTML]{EFEFEF} RQ2.1  &  
    \cellcolor[HTML]{EFEFEF}\begin{tabular}[c]{@{}c@{}} 
        The best explainer differs if evaluated in\\
        terms of plausibility or fidelity, but\\
        the two best performing ones are both \\
        edge- and gradient-based. 
    \end{tabular} &
    \cellcolor[HTML]{EFEFEF}\begin{tabular}[c]{@{}c@{}}
        The best explainer differs if evaluated in terms\\
        of plausibility (\gradcam) or fidelity (\gnnexpl).\\
    \end{tabular} \\
        % \hline

 RQ2.2  &      
    \begin{tabular}[c]{@{}c@{}}
        \subgraphx explains the maximal\\
        number of models.\\ 
    \end{tabular}  & 
    \begin{tabular}[c]{@{}c@{}}
        \gnnexpl explains the maximal \\
        number of models. \\ 
    \end{tabular} \\
    % \hline
    
 \cellcolor[HTML]{EFEFEF}RQ2.3  &  
    \cellcolor[HTML]{EFEFEF}\begin{tabular}[c]{@{}c@{}} 
        Perturbation methods excel in plausibility,\\
        while generative methods perform best in\\
        fidelity, but may not capture the\\
        expected ground truth.
    \end{tabular} & 
    \cellcolor[HTML]{EFEFEF}\begin{tabular}[c]{@{}c@{}} 
        Gradient methods excel in both plausibility\\
        and fidelity.
    \end{tabular}\\  %\hline    
 RQ2.4  &  
    \begin{tabular}[c]{@{}c@{}} 
        Edge-based explainers outperform \\
        node-based ones, possibly because they\\
        are designed for graph-explanation tasks,\\
        while node-based explainers are adapted\\
        from other contexts.
    \end{tabular} & 
    \begin{tabular}[c]{@{}c@{}} 
        Node-based explainers outperform edge-based.\\
        The motivation remains unclear, \\
        necessitating future investigation.
    \end{tabular} \\   

 \midrule
 \multicolumn{3}{l}{\multirow{2}{*}{\textit{\textbf{\normalsize RQ3: How do different types of problems affect the explanations?}}}}\\ 
 \multicolumn{3}{l}{} \\ 
 
 \midrule

\cellcolor[HTML]{EFEFEF}\begin{tabular}[c]{@{}c@{}} 
    Motif-based\\explanations
   \end{tabular} &
  \multicolumn{2}{c}{
    \cellcolor[HTML]{EFEFEF}\begin{tabular}[c]{@{}c@{}}
            
        In learning problems where the underlying ground truth is a motif, care should be taken in\\
        designing the dataset. In some popular scenarios, the ground-truth motif contains a smaller\\
        minimal discriminant subgraph (MDS). For example, in \datasetgcone, the expected explanation\\
        is a 3x3 grid, while the MDS is a simple square. This clearly undermines a correct evaluation\\
        of the explanations being extracted. The problem can also affect the evaluation of real-world\\
        explanations by domain experts.

        %In some evaluation settings there is a clear mismatch between the ground truth \\
        %explanation and the \textbf{MDF} (i.e., the simplest subgraph discriminant of\\
        %the classes),raising the question of why the GNN should prefer a more complex \\
        %motif to be learned instead of a simpler pattern. For example, in \datasetgcone\\
        %the expected explanation is a 3x3 grid, while the MDS is a simple square.\\
        %We emphasize that more attention should be given in the design of synthetic datasets\\
        %without such ambivalent behaviors, for example by avoiding too complex motif as\\
        %target explanation which might contain nested MDS sufficient for solving\\
        %the downstream task.
        \end{tabular} }\\ % \hline

 \begin{tabular}[c]{@{}c@{}} 
        Class-specific\\ explanations
   \end{tabular} &
    \multicolumn{2}{c}{
    \begin{tabular}[c]{@{}c@{}} 

        %The phenomenon of \textbf{laziness} notably affects explainability, with the GNN effectively \\ 
        %explaining only the class it has learned the key feature for. This behavior is quantified in\\
        %our metrics and evident in the Pareto plots of Section~\ref{sec:results} (Figures \ref{fig:pareto_d1_and_stability} and \ref{fig:pareto_d4_and_stability},where top-right pairs\\
        %distinctly correspond to a single class in each (architecture, explainer) pair.

        %The phenomenon of \textbf{laziness} impacts explainability, as only the class for which the GNN \\
        %has learned the discriminant feature will be explained. Hence, it is crucial to identify the GNN\\
        %learned class and evaluate explanations within that class.
        
        When considering explanations for both positive and negative predictions, the phenomenon of\\
        laziness severely affects the quality of the explanation being extracted. GNNs typically tend\\
        to learn features for only one of the two classes, and predict the other as a "default" option.\\ 
        Extracting explanations for the "default" class is useless and potentially misleading.

        %.This can be seen in Figures \ref{fig:pareto_d1_and_stability} and \ref{fig:pareto_d4_and_stability} of \\
        %Section~\ref{sec:results}, where the top-right pairs (architecture, explainer) are different.
        %This means that an architecture learns only a single class
        
        %in the Pareto plots of Section~\ref{sec:results} (Figures \ref{fig:pareto_d1_and_stability} and \ref{fig:pareto_d4_and_stability}), where the top-right pairs are clearly\\
        %distinct for the two classes, meaning that such (architecture, explainer)-pair is explaining\\         only a single class.
    \end{tabular} }\\ % \hline

\cellcolor[HTML]{EFEFEF} \begin{tabular}[c]{@{}c@{}} 
        Aggregation-\\dependent\\ explanations
   \end{tabular} &
    \multicolumn{2}{c}{
   \cellcolor[HTML]{EFEFEF} \begin{tabular}[c]{@{}c@{}}
        
        The sum aggregator, used in tasks like counting substructures (e.g., \datasetgcthree),\\
        makes the graph-level embedding dependent on the number of nodes in the graph. \\
        However, in fidelity evaluation, explanations involve smaller subgraphs, potentially\\
        introducing a distribution shift that harms model fidelity, as also previously noted in~\cite{zhao2022consistency}.
        
        % models the distribution of nodes in training graphs as $P(y|n_{v_{train}})$. However,\\
        % in fidelity evaluation, explanations involve smaller subgraphs, so $P(y|n_{v_{expls}})$\\
        % differs from $P(y|n_{v_{train}})$. Using smaller graphs as GNN input can introduce distribution \\
        % shifts that affect model fidelity, as noted in~\cite{zhao2022consistency}.
        
        %The sum aggregator is sometimes made necessary for solving specific \\
        %tasks, like counting substructures as in \datasetgcthree. Informally and \\
        %loosely speaking, we could say that it permits to model the distribution \\
        %of the number of nodes in the training graphs $P(y|n_{v_{train}})$. When it \\
        %comes to evaluating fidelity, however, given that the explanations \\
        %consist of subgraphs of (possibly) much lower size, it follows that \\
        %$P(y|n_{v_{expls}}) \ne P(y|n_{v_{train}})$. Consequently, feeding the GNN \\
        %with much smaller graphs could result in distribution shifts that might \\
        %adversely affect the model's fidelity \\
        %assessment as reported also by~\cite{zhao2022consistency}. \\
    \end{tabular} }\\  
    
\end{tabular}
\caption{\rev{Summary of the main lessons learned, divided by research question.}}
\label{tab:summary_res}
\end{table}

In Table \ref{tab:summary_res}, we present a comprehensive summary of the principal outcomes derived from our research. Each row within the table is dedicated to addressing a specific research question.

\section{Discussion}\label{sec:discussion}

The main objective of this work is to experimentally study the effectiveness of explainers on different GNNs and types of data, identifying current pitfalls and formulating possible future directions in the field of GNN explainability. Given that GNNs may learn different concepts, possibly less intuitive, than those expected by humans, defining ground truths is not a trivial task and may be prone to human biases. A first simple remark is that GNNs, as neural networks in general, tend to be lazy and learn a "default" option for one of the classes. For this reason, a high accuracy does not necessarily imply having learned the ground-truth concept for the class.
A useful insight that emerged from our analysis is that these human biases can often be detected by comparing plausibility and fidelity. Indeed, an explainer with high fidelity and low plausibility (or vice-versa) clearly indicates a discrepancy between what is considered to be the ground truth and the concept learned by the GNN. It is important to remark, however, that fidelity alone may not be the optimal choice for both graph and node classification. In the case of graph classification, the aggregation used to convert nodes embedding into a graph embedding directly influences the fidelity. In particular, sum aggregation, which is needed to allow GNN to count substructures, often negatively affects fidelity, with the mere ground-truth structure achieving relatively low fidelity because of the reduced number of nodes which in turn reduces the norm of the overall embedding. 

The aggregation mechanism is a crucial component of GNNs and its decision directly affects the quality of the explanation. Nonetheless, there is a lack of works studying the impact of the aggregation mechanism on explainability. This is an interesting direction for further research.
Reliably measuring fidelity can be tricky for node classification too. On the one hand, comprehensiveness is poorly defined when explaining node predictions (see Section~\ref{sec:metrics}). On the other hand, measuring fidelity only in terms of sufficiency introduces a bias that favours larger explanations. Indeed, finding the optimal metric for evaluating explainers is still an open problem that deserves further investigation.

Identifying a general category of explainers working consistently better than others is challenging. However, our results suggest that gradient-based explainers are more suited in explaining node classification networks. We conjecture that this is due to the fact that in node classification, the gradient is computed only on the neighborhood of the node under investigation, limiting the receptive field of the network. On the other hand, the category of explainers that best explain GNNs for graph classification are those that focus on edges, be it by perturbation or gradient. In general, edge-based explainers outperform node-based ones whenever node features are not available. 
%It is worth mentioning that edge-based explainers require GNNs to support the learning of edge weights. Any GNN can however be easily adapted to work with edge-based explainers by using constant values as edge weight.

Concerning GNN architectures, there is a substantial difference in their explainability, regardless of the explainer that best suits each of them. We believe that this result is surprising yet not fully understood, given that explainers are usually aimed to be model agnostic. 
Given the importance of explaining predictions, it would be advisable to include explainability as a metric to be optimized when designing novel GNN architectures. \rev{Finally,
we would like to highlight that this work did not consider the most recent and complex architectures, like diffusion-based~\cite{NIPS2016_390e9825, gasteiger2022diffusion, Tamil_Selvi_2021} and spatio-temporal models~\cite{ tgn_icml_grl2020, MICHELI202285,longa2023graph}, which are typically omitted from the evaluation of explainers in the literature. We leave an investigation on the usability of existing explainers for these architectures, and on the need for ad-hoc explainers specifically designed for their characteristics, to future research.}

\section{Future directions}\label{sec:future}
Several explainers have been proposed and tested for the classical GCN architecture~\cite{kipf2016semi}, but we argue that significant work has yet to be developed to fully grasp the complex interaction of models and explanations and their interpretability.
In the following we outline some promising research directions we believe deserve further investigation.

\subsection{Going beyond instance-level explanations}
While many studies focus on instance-level explanations, model-level explanations are less explored~\cite{wang2024gnnboundary,yu2024mage,muller2024graphchef,nian2024globally,GLGExplainer,yuan2020xgnn}. 
A deeper understanding of global explanations could provide a more comprehensive insight into the inner workings of GNNs. This approach can reveal how features are aggregated and utilized across the entire model, offering a broader perspective that complements instance-level insights.
Apart from extracting explanations for providing human-aware guidance on the behavior of a trained GNN, several works also use explanations during the training of the model as a regularization term. 
It has been shown, in fact, that explanation-aware training of GNNs can result in better generalization, faster convergence, and intrinsically higher explainability~\cite{shi2023engage,spinelli2022meta,giunchiglia2022towards}.
Bringing this design choice even further, a promising recent trend is to condition the GNN to output predictions uniquely from its explanations. % a framework often referred to as \textit{self-explainable} models. In this way, if the model is faithful to the extracted explanation, the explanation unambiguously depicts the underlying reason for the model's predictions. 
In this framework, the GNN is typically divided into two components: first an explanation extractor takes as input the entire graph and outputs the explanation, then a classifier takes as input the explanation and makes the final prediction. Explanations can come in different forms, like subgraphs~\cite{miao2022interpretable, serra2024l2xgnnlearningexplaingraph}, prototypes similarity~\cite{zhang2022protgnn, ragno2022prototype}, or activation values for semantic concepts~\cite{magister2022encoding}.
Following this schema, if the model is faithful to the extracted explanation then the explanation unambiguously depicts the underlying reason for the model's predictions.
Unfortunately, recent studies have shown that despite this training-aware conditioning the resulting models are often unfaithful to the explanations, meaning that it does not fully depict the intended model behavior \cite{christiansen2023faithful}, an issue shared with classic post-hoc explainers.
Therefore, further research is needed in the study of trustworthy and effective explainers, whether they are used at training or evaluation time.

\subsection{What is an explanation}\label{subsec:whatis}
A direction that warrants further investigation is the definition of explanations in GNNs. Current explainability-related studies typically consider a subgraph as an explanation. However, we argue that in some real-world applications, this subgraph may not provide a sufficiently faithful explanation. For example, in a recommendation system, a user might purchase an item not because of its inherent features, but because it is currently trending. In this case, the explanation should highlight the high degree of the item's node, rather than its enclosing subgraph. An additional example is when local predictions are influenced by a global property of the network. In traffic flow prediction, for instance, the congestion of a particular street may be due to its role as the sole connection between two parts of the city. This situation can be explained by the high betweenness centrality of that street, indicating its critical position in the network.
Another intriguing direction involves investigating the causality of explanations~\cite{lin2021generative,wang2022reinforced,lin2022orphicx}. This entails understanding the minimal modifications required to the computational graph of a node to alter its prediction. Such causal explanations can provide deeper insights into what the model has learned to make predictions and how node features are aggregated and utilized by the model.

\subsection{The role of the dataset}
%We used only synthetic datasets in our experiments because they provide ground truth explanations necessary for our evaluations. However, despite the predefined ground truth, we discovered that some well-known benchmark datasets have design issues (see Section \ref{sec:results})\cite{faber2021comparing}. Therefore, it is crucial to reconsider these synthetic benchmarks, paying close attention to potential shortcuts and ensuring that minimal explanations are provided. 
We used only synthetic datasets in our experiments because they provide the ground truth explanations necessary for our evaluations. However, despite the predefined ground truth, some well-known benchmark datasets have design issues~\cite{faber2021comparing}. Therefore, it is crucial to reconsider these synthetic benchmarks, paying close attention to potential shortcuts and ensuring that minimal explanations are provided. Furthermore, as discussed in Subsection~\ref{subsec:whatis}, it is essential to define synthetic datasets that match the expected ground truth in real-world scenarios, where the explanation may differ from a local structure.

\section{Conclusion}\label{sec:conclusion}
In this survey we proposed an extensive experimental study to quantify the effectiveness of the existing explainers and to obtain actionable recommendations to select the optimal method for a given task.  For this comparison we evaluated ten explainers on eight different GNN architectures, all chosen to represent the most commonly utilized instances in a vast taxonomy of existing solutions. These methods have been tested on six different datasets for both graph and node classification, carefully designed or adapted to model interesting and challenging aspects of real-world datasets. As a result of our experimental study, we were able to describe significant criticalities in the common explainer evaluation methods and to identify recurring patterns that make some categories of explainers preferable in certain situations.  In particular, we proposed insights on which explainer to use, and how to use it, depending on the available data. Our findings naturally point to promising future research directions, and especially highlight once more that much has yet to be understood to achieve a satisfactory GNN explainability.

\bibliographystyle{plain}
\bibliography{my_biblio}

\newpage
\appendixpage
\appendix

\section{Detailed Results}\label{sup:tabs}
In this section, we present the raw plausibility and fidelity values for each dataset, model, and explainer, thus without the aggregations presented in the main text. Note that entries marked with \filter~ indicate that these explanations did not meet the criteria outlined in Section \ref{sec:aggregation} and were removed accordingly.

\begin{table}[h!]
\footnotesize
\begin{tabular}{l|cccccc}
Explainers             & \gcn & \gin  & \cheb   & \setset  & \mincut  & \graphconv \\ \hline
\cam                   & 0.837   & 0.211   & 0.239   & 0.657   & \filter & \filter \\
\gradcam               & 0.876   & 0.789   & 0.697   & 0.660   & \filter & \filter \\
\guidedbp              & 0.684   & \filter & 0.907   & 0.197   & 0.169   & \filter \\
\ignode                & 0.364   & 0.108   & 0.423   & 0.176   & 0.174   & 0.097 \\
\gradexp               & 0.608   & \filter & 0.487   & 0.195   & 0.169   & \filter \\
\pgmexpl               & 0.487   & 0.484   & 0.486   & 0.484   & 0.486   & 0.487\\
\sal                   & 0.948   &         & 0.837   & 0.321   &         & 0.924\\
\igedge                & 0.528   &         & 0.837   & 0.209   &         & 0.975\\
\gnnexpl               & 0.735   & 0.327   &         & 0.449   &0.813    & 0.667\\
\pgexpl                & 0.407   & 0.295   & 0.935   & 0.703   &0.771    & \filter \\
\subgraphx             & 0.648   & 0.698   & 0.897   & 0.823   &0.698    & 0.858\\
\rgexpl                & 0.983   & 0.972   & 0.685   & 0.089   &         & 0.998         
\end{tabular}
\caption{Plausibility \datasetgcone}
\end{table}

% If you use beamer only pass "xcolor=table" option, i.e. \documentclass[xcolor=table]{beamer}https://www.overleaf.com/project/63563a047c2bad478f2955a7
\begin{table}[ht]
\footnotesize
\begin{tabular}{l|cccccccccccc}

Explainers  & \multicolumn{2}{c}{\gcn} & \multicolumn{2}{c}{\gin}  & \multicolumn{2}{c}{\cheb}   & \multicolumn{2}{c}{\setset}  & \multicolumn{2}{c}{\mincut}  & \multicolumn{2}{c}{\graphconv} \\ \hline
Class & 0 & 1 & 0 & 1 & 0 & 1 & 0 & 1 & 0 & 1 & 0 & 1 \\ \hline
\cam & \filter & 0.574 & 0.197 & 0.333 & \filter  & 0.187 & 0.176 & 0.147 & \filter & \filter & \filter & \filter \\
\gradcam & 0.390 & 0.626 & 0.166 & 0.519 & 0.395 & 0.580 & 0.419 & 0.066 & \filter & \filter & \filter & \filter \\
\guidedbp & 0.585 & 0.628 & \filter & \filter & 0.663 & 0.651 & 0.621 & 0.146 & 0.432 & 0.176 & \filter & \filter \\
\ignode & 0.490 & 0.475 & \filter & 0.121 & 0.081 & 0.556 & 0.347 & 0.075 & 0.455 & 0.144 & \filter & 0.131 \\
\gradexp & 0.602 & 0.608 & \filter & \filter & 0.015 & 0.169 & 0.575 & 0.185 & 0.436 & 0.187 & 0.043 & \filter \\
\pgmexpl & 0.404 & 0.436 & 0.400 & 0.378 & 0.439 & 0.364 & 0.397 & 0.313 & 0.381 & 0.349 & 0.400 & 0.388 \\
\sal & 0.344 & 0.723 &  &  & 0.282 & 0.495 & 0.523 & 0.099 &  &  & 0.048 & 0.601 \\
\igedge & 0.469 & 0.434 &  &  & 0.282 & 0.495 & 0.671 & 0.126 &  &  & 0.045 & 0.769 \\
\gnnexpl & 0.325 & 0.584 & 0.184 & 0.262 &  &  & 0.401 & 0.053 & 0.384 & 0.539 & 0.189 & 0.531 \\
\pgexpl & 0.221 & 0.500 & 0.117 & 0.129 & 0.102 & 0.812 & 0.136 & 0.480 & 0.148 & 0.731 & \filter & \filter \\
\subgraphx & 0.329 & 0.593 & 0.287 & 0.414 & 0.442 & 0.772 & 0.298 & 0.641 & 0.190 & 0.526 & 0.284 & 0.890 \\
\rgexpl & 0.370 & 0.770 & 0.043 & 0.796 & 0.637 & 0.430 & 0.701 & 0.126 &  &  & 0.044 & 0.816
\end{tabular}
\caption{F1 fidelity \datasetgcone}
\end{table}

\begin{table}[ht]
\footnotesize
\begin{tabular}{l|cccccccccccc}
Explainers  & \multicolumn{2}{c}{\gcn} & \multicolumn{2}{c}{\gin}  & \multicolumn{2}{c}{\cheb}   & \multicolumn{2}{c}{\setset}  & \multicolumn{2}{c}{\mincut}  & \multicolumn{2}{c}{\graphconv} \\ \hline
Class & 0 & 1 & 0 & 1 & 0 & 1 & 0 & 1 & 0 & 1 & 0 & 1 \\ \hline
\cam & 0.087 & 0.230 & \filter & \filter & 0.961 & 0.092 & 0.409 & \filter & 0.160 & 0.420 & \filter & \filter \\
\gradcam & 0.866 & 0.823 & \filter & \filter & 0.479 & 0.481 & 0.957 & 0.896 & 0.919 & 0.875 & 0.147 & 0.189 \\
\guidedbp & 0.908 & 0.142 & \filter & \filter & \filter & \filter & 0.946 & 0.048 & 0.737 & 0.249 & \filter & \filter \\
\ignode & 0.761 & 0.067 & \filter & 0.186 & 0.464 & 0.571 & 0.777 & 0.043 & \filter & 0.201 & \filter & 0.161 \\
\gradexp & 0.905 & 0.139 & \filter & \filter & \filter & \filter & 0.953 & 0.032 & 0.729 & 0.521 & \filter & \filter \\
\pgmexpl & 0.488 & 0.484 & 0.488 & 0.485 & 0.488 & 0.485 & 0.486 & 0.484 & 0.489 & 0.485 & 0.487 & 0.486 \\
\sal & 0.849 & 0.838 &  &  & 0.454 & 0.396 & 0.300 & 0.268 &  &  & 0.823 & 0.749 \\
\igedge & 0.544 & 0.580 &  &  & 0.454 & 0.396 & 0.085 & 0.151 &  &  & 0.815 & 0.928 \\
\gnnexpl & 0.362 & 0.544 & 0.525 & 0.782 &  &  & 0.496 & 0.516 & 0.531 & 0.653 & 0.527 & 0.785 \\
\pgexpl & 0.795 & 0.756 & 0.452 & 0.336 & 0.984 & 0.886 & 0.679 & 0.621 & 0.978 & 0.968 & \filter & \filter \\
\subgraphx & 0.321 & 0.706 & 0.593 & 0.757 & 0.474 & 0.744 & 0.202 & 0.558 & 0.594 & 0.752 & 0.594 & 0.752 \\
\rgexpl & 0.529 & 0.702 & 0.763 & 0.686 & 0.212 & 0.338 & 0.090 & 0.148 &  &  & 0.817 & 0.659
\end{tabular}
\caption{Plausibility \datasetgctwo}
\end{table}

\begin{table}[ht]
\footnotesize
\begin{tabular}{l|cccccccccccc}
Explainers  & \multicolumn{2}{c}{\gcn} & \multicolumn{2}{c}{\gin}  & \multicolumn{2}{c}{\cheb}   & \multicolumn{2}{c}{\setset}  & \multicolumn{2}{c}{\mincut}  & \multicolumn{2}{c}{\graphconv} \\ \hline
Class & 0 & 1 & 0 & 1 & 0 & 1 & 0 & 1 & 0 & 1 & 0 & 1 \\ \hline
\cam & 0.613 & 0.086 & \filter & \filter & 0.234 & 0.147 & 0.482 & \filter & 0.787 & 0.254 & \filter & \filter \\
\gradcam & 0.386 & 0.567 & \filter & \filter & 0.571 & 0.168 & 0.338 & 0.563 & 0.193 & 0.684 & 0.275 & 0.272 \\
\guidedbp & 0.215 & 0.192 & \filter & \filter & \filter & \filter & 0.520 & 0.523 & 0.391 & 0.378 & \filter & \filter \\
\ignode & 0.399 & 0.175 & \filter & 0.115 & 0.209 & 0.135 & 0.274 & 0.140 & \filter & 0.595 & \filter & 0.126 \\
\gradexp & 0.068 & 0.074 & \filter & \filter & \filter & \filter & 0.083 & 0.084 & 0.377 & 0.006 & \filter & \filter \\
\pgmexpl & 0.392 & 0.475 & 0.400 & 0.466 & 0.430 & 0.478 & 0.383 & 0.440 & 0.474 & 0.355 & 0.401 & 0.481 \\
\sal & 0.286 & 0.708 &  &  & 0.602 & 0.235 & 0.593 & 0.087 &  &  & 0.033 & 0.427 \\
\igedge & 0.577 & 0.432 &  &  & 0.602 & 0.235 & 0.854 & 0.056 &  &  & 0.034 & 0.637 \\
\gnnexpl & 0.563 & 0.566 & 0.140 & 0.729 &  &  & 0.254 & 0.387 & 0.355 & 0.608 & 0.186 & 0.732 \\
\pgexpl & 0.121 & 0.784 & 0.225 & 0.186 & 0.092 & 0.79 & 0.134 & 0.57 & -0.015 & 0.661 & \filter & \filter \\
\subgraphx & 0.495 & 0.680 & 0.293 & 0.637 & 0.509 & 0.674 & 0.733 & 0.484 & 0.261 & 0.524 & 0.296 & 0.633 \\
\rgexpl & 0.273 & 0.693 & 0.087 & 0.638 & 0.212 & 0.338 & 0.758 & 0.097 &  &  & 0.817 & 0.659
\end{tabular}
\caption{F1 fidelity \datasetgctwo}
\end{table}

\begin{table}[ht]
\scriptsize
\begin{tabular}{l|cccccccccccccccccc}
Explainers  & \multicolumn{3}{c}{\gcn} & \multicolumn{3}{c}{\gin}  & \multicolumn{3}{c}{\cheb}   & \multicolumn{3}{c}{\setset}  & \multicolumn{3}{c}{\mincut}  & \multicolumn{3}{c}{\graphconv} \\ \hline
Class & 0 & 1 & 2 & 0 & 1 & 2  & 0 & 1 & 2  & 0 & 1 & 2  & 0 & 1 & 2  & 0 & 1 & 2  \\ \hline
\cam & \filter & 0.52 & 0.63 & 0.34 & 0.81 & 0.55 & 0.39 & 0.48 & 0.52 & \filter & \filter & \filter & \filter & \filter & \filter & 0.67 & 0.20 & 0.41 \\
\gradcam & 0.39 & 0.27 & 0.42 & \filter & \filter & \filter & 0.32 & 0.14 & 0.37 & 0.71 & 0.88 & 0.60 & \filter & \filter & \filter & 0.49 & 0.53 & 0.50 \\
\guidedbp & 0.72 & 0.28 & \filter & 0.68 & 0.56 & \filter & 0.70 & 0.22 & \filter & 0.71 & \filter & \filter & 0.68 & 0.40 & 0.36 & 0.69 & 0.43 & 0.38 \\
\ignode & \filter & 0.11 & 0.37 & 0.67 & 0.84 & \filter & 0.69 & 0.15 & 0.37 & 0.73 & 0.11 & 0.37 & 0.68 & 0.85 & \filter & 0.69 & 0.47 & 0.37 \\
\gradexp & 0.70 & 0.19 & \filter & 0.36 & 0.41 & \filter & 0.69 & 0.17 & \filter & 0.71 & \filter & \filter & 0.68 & 0.40 & 0.36 & 0.69 & 0.44 & 0.36 \\
\pgmexpl & 0.49 & 0.50 & 0.50 & 0.49 & 0.50 & 0.50 & 0.49 & 0.50 & 0.50 & 0.49 & 0.50 & 0.50 & 0.49 & 0.50 & 0.50 & 0.49 & 0.50 & 0.50 \\
\sal & 0.72 & 0.89 & 0.62 &  &  &  & 0.71 & 0.88 & 0.60 & 0.72 & 0.77 & 0.59 &  &  &  & 0.70 & 0.88 & 0.63 \\
\igedge & 0.72 & 0.89 & 0.63 &  &  &  & 0.71 & 0.88 & 0.61 & 0.72 & 0.90 & 0.63 &  &  &  & 0.63 & 0.66 & 0.63 \\
\gnnexpl & 0.47 & 0.77 & 0.58 & 0.33 & 0.26 & 0.51 &  &  &  & 0.39 & 0.87 & 0.63 & 0.23 & 0.10 & 0.52 & 0.30 & 0.52 & 0.59 \\
\pgexpl & 0.54 & 0.58 & 0.54 & 0.34 & 0.19 & 0.38 & 0.34 & 0.21 & 0.42 & 0.71 & 0.90 & 0.63 & 0.28 & 0.11 & 0.38 & \filter & \filter & \filter \\
\subgraphx & 0.48 & 0.46 & 0.47 & 0.47 & 0.39 & 0.49 & 0.47 & 0.44 & 0.50 & 0.46 & 0.42 & 0.48 &  &  &  & 0.47 & 0.37 & 0.53 \\
\rgexpl & 0.71 & 0.69 & 0.54 & 0.49 & 0.49 & 0.50 & 0.71 & 0.70 & 0.54 & 0.70 & 0.70 & 0.55 &  &  &  & 0.54 & 0.52 & 0.51
\end{tabular}
\caption{Plausibility \datasetgcthree}
\end{table}

\begin{table}[ht]
\scriptsize
\begin{tabular}{l|cccccccccccccccccc}
Explainers  & \multicolumn{3}{c}{\gcn} & \multicolumn{3}{c}{\gin}  & \multicolumn{3}{c}{\cheb}   & \multicolumn{3}{c}{\setset}  & \multicolumn{3}{c}{\mincut}  & \multicolumn{3}{c}{\graphconv} \\ \hline
Class & 0 & 1 & 2 & 0 & 1 & 2  & 0 & 1 & 2  & 0 & 1 & 2  & 0 & 1 & 2  & 0 & 1 & 2  \\ \hline
\cam & \filter & 0.54 & 0.27 & 0.41 & 0.73 & 0.55 & 0.17 & 0.54 & 0.24 & \filter & \filter & \filter & \filter & \filter & \filter & 0.21 & 0.42 & 0.61 \\
\gradcam & 0.49 & 0.34 & 0.34 & \filter & \filter & \filter & 0.08 & 0.25 & 0.32 & 0.13 & 0.38 & 0.59 & \filter & \filter & \filter & 0.39 & 0.53 & 0.58 \\
\guidedbp & 0.40 & 0.64 & \filter & 0.51 & 0.59 & \filter & 0.54 & 0.65 & \filter & 0.59 & \filter & \filter & 0.21 & 0.38 & 0.50 & 0.52 & 0.61 & 0.68 \\
\ignode & \filter & 0.14 & 0.25 & 0.09 & 0.72 & \filter & 0.28 & 0.34 & 0.31 & 0.02 & 0.24 & 0.25 & 0.19 & 0.09 & nan & 0.15 & 0.48 & 0.51 \\
\gradexp & 0.02 & 0.18 & \filter & 0.54 & 0.61 & \filter & 0.54 & 0.66 & \filter & 0.35 & \filter & \filter & 0.40 & 0.63 & 0.63 & 0.74 & 0.60 & 0.63 \\
\pgmexpl & 0.39 & 0.38 & 0.40 & 0.38 & 0.39 & 0.35 & 0.38 & 0.39 & 0.41 & 0.40 & 0.38 & 0.39 & 0.38 & 0.37 & 0.42 & 0.35 & 0.39 & 0.40 \\
\sal & 0.01 & 0.27 & 0.79 &  &  &  & 0.06 & 0.45 & 0.76 & 0.01 & 0.39 & 0.68 &  &  &  & 0.16 & 0.40 & 0.93 \\
\igedge & 0.01 & 0.29 & 0.84 &  &  &  & 0.06 & 0.45 & 0.76 & 0.01 & 0.29 & 0.84 &  &  &  & 0.24 & 0.40 & 0.90 \\
\gnnexpl & 0.18 & 0.34 & 0.71 & 0.79 & 0.33 & 0.57 &  &  &  & 0.59 & 0.30 & 0.86 & 0.72 & 0.48 & 0.61 & 0.68 & 0.40 & 0.87 \\
\pgexpl & 0.07 & 0.37 & 0.26 & 0.28 & 0.33 & 0.05 & 0.64 & 0.13 & 0.04 & 0.01 & 0.43 & 0.71 & 0.76 & 0.13 & 0.11 & \filter & \filter & \filter \\
\subgraphx & 0.30 & 0.21 & 0.26 & 0.25 & 0.28 & 0.25 & 0.32 & 0.21 & 0.33 & 0.43 & 0.21 & 0.23 &  &  &  & 0.33 & 0.26 & 0.28 \\
\rgexpl & 0.02 & 0.19 & 0.39 & 0.28 & 0.26 & 0.25 & 0.01 & 0.22 & 0.39 & 0.01 & 0.10 & 0.34 &  &  &  & 0.14 & 0.28 & 0.15
\end{tabular}
\caption{F1 fidelity \datasetgcthree}
\end{table}

\begin{table}[ht]
\scriptsize
\begin{tabular}{l|cccccccccccccccc}
Explainers  & \multicolumn{2}{c}{\gcn} & \multicolumn{2}{c}{\gin}  & \multicolumn{2}{c}{\cheb}   & \multicolumn{2}{c}{\setset}  & \multicolumn{2}{c}{\mincut}  & \multicolumn{2}{c}{\graphconv}& \multicolumn{2}{c}{\gat}& \multicolumn{2}{c}{\sage} \\ \hline
Class & 0 & 1 & 0 & 1 & 0 & 1 & 0 & 1 & 0 & 1 & 0 & 1 & 0 & 1 & 0 & 1 \\ \hline
\cam & 0.02 & 0.06 & 0.14 & 0.05 & 0.26 & 0.29 & 0.77 & 0.04 & 0.48 & 0.83 & 0.25 & 0.39 & 0.87 & 0.10 & \filter & \filter \\
\gradcam & 0.14 & 0.93 & 0.54 & 0.72 & 0.76 & 0.42 & 0.54 & 0.68 & 0.32 & 0.71 & 0.27 & 0.92 & 0.76 & 0.49 & 0.31 & 0.85 \\
\guidedbp & 0.00 & \filter & \filter & \filter & \filter & \filter & \filter & \filter & 0.94 & 1.00 & \filter & \filter & \filter & \filter & \filter & \filter \\
\ignode & 0.08 & \filter & 0.03 & 0.00 & 0.00 & 0.29 & 0.03 & \filter & 0.01 & 0.01 & 0.01 & 0.02 & 0.03 & 0.01 & 0.17 & \filter \\
\gradexp & 0.00 & 0.80 & \filter & \filter & \filter & \filter & \filter & \filter & 0.96 & 0.99 & \filter & \filter & \filter & \filter & \filter & \filter \\
\pgmexpl & 0.49 & 0.49 & 0.49 & 0.49 & 0.49 & 0.49 & 0.49 & 0.49 & 0.50 & 0.50 & 0.49 & 0.50 & 0.49 & 0.50 & 0.49 & 0.49 \\
\sal & 0.44 & 0.35 &  &  & 0.66 & 0.72 & 0.95 & 0.90 &  &  & 0.99 & 0.95 &  &  &  &  \\
\igedge & 0.29 & 0.26 &  &  & 0.66 & 0.72 & 0.28 & 0.67 &  &  & 1.00 & 0.96 &  &  &  &  \\
\gnnexpl & 0.26 & 0.65 & 0.82 & 0.85 &  &  & 0.81 & 0.47 & 0.47 & 0.87 & 0.77 & 0.35 & 0.86 & 0.79 & 0.77 & 0.47 \\
\pgexpl & \filter & \filter & \filter & \filter & \filter & \filter & 0.03 & 0.99 & 0.90 & 0.36 & \filter & \filter & \filter & \filter & \filter & \filter \\
\subgraphx & 0.79 & 0.84 & 0.81 & 0.92 & 0.80 & 0.81 & 0.79 & 0.87 & 0.77 & 0.82 & 0.80 & 0.84 & 0.75 & 0.79 & 0.78 & 0.89 \\
\rgexpl & 0.40 & 0.44 & 0.45 & 0.72 & 0.96 & 0.38 & 0.48 & 0.47 &  &  & 0.42 & 0.35 & 0.46 & 0.48 & 0.49 & 0.90
\end{tabular}
\caption{Plausibility \datasetgcfour}
\end{table}

\begin{table}[ht]
\scriptsize
\begin{tabular}{l|cccccccccccccccc}
Explainers  & \multicolumn{2}{c}{\gcn} & \multicolumn{2}{c}{\gin}  & \multicolumn{2}{c}{\cheb}   & \multicolumn{2}{c}{\setset}  & \multicolumn{2}{c}{\mincut}  & \multicolumn{2}{c}{\graphconv}& \multicolumn{2}{c}{\gat}& \multicolumn{2}{c}{\sage} \\ \hline
Class & 0 & 1 & 0 & 1 & 0 & 1 & 0 & 1 & 0 & 1 & 0 & 1 & 0 & 1 & 0 & 1 \\ \hline
\cam & 0.02 & 0.11 & 0.11 & 0.06 & 0.07 & 0.04 & 0.17 & 0.08 & 0.29 & 0.60 & 0.22 & 0.20 & 0.15 & 0.15 & \filter & \filter \\
\gradcam & 0.12 & 0.82 & 0.39 & 0.54 & 0.42 & 0.25 & 0.35 & 0.54 & 0.29 & 0.63 & 0.18 & 0.50 & 0.55 & 0.38 & 0.19 & 0.73 \\
\guidedbp & 0.65 & \filter & \filter & \filter & \filter & \filter & \filter & \filter & 0.61 & 0.59 & \filter & \filter & \filter & \filter & \filter & \filter \\
\ignode & 0.05 & \filter & 0.03 & 0.39 & 0.09 & 0.48 & 0.08 & \filter & 0.45 & -0.05 & 0.08 & 0.01 & 0.13 & 0.63 & 0.06 & \filter \\
\gradexp & 0.61 & 0.66 & \filter & \filter & \filter & \filter & \filter & \filter & 0.65 & 0.62 & \filter & \filter & \filter & \filter & \filter & \filter \\
\pgmexpl & 0.38 & 0.38 & 0.37 & 0.39 & 0.34 & 0.40 & 0.40 & 0.40 & 0.34 & 0.38 & 0.34 & 0.40 & 0.46 & 0.47 & 0.38 & 0.38 \\
\sal & 0.18 & 0.24 &&& 0.53 & 0.17 & 0.40 & 0.40 &&& 0.87 & 0.19 &&&&\\
\igedge & 0.15 & 0.24 &&& 0.53 & 0.17 & 0.25 & 0.24 &&& 0.84 & 0.18 &&&&\\
\gnnexpl & 0.15 & 0.57 & 0.50 & 0.69 &&& 0.79 & 0.23 & 0.29 & 0.33 & 0.76 & 0.03 & 0.45 & 0.47 & 0.58 & 0.18 \\
\pgexpl & \filter & \filter & \filter & \filter & \filter & \filter & 0.01 & 0.90 & 0.79 & 0.03 & \filter & \filter & \filter & \filter & \filter & \filter \\
\subgraphx & 0.56 & 0.44 & 0.49 & 0.76 & 0.47 & 0.52 & 0.49 & 0.61 & 0.58 & 0.44 & 0.69 & 0.38 & 0.35 & 0.53 & 0.57 & 0.55 \\
\rgexpl & 0.11 & 0.33 & 0.14 & 0.51 & 0.49 & 0.10 & 0.09 & 0.20 &&& 0.16 & 0.06 & 0.28 & 0.28 & 0.19 & 0.46
\end{tabular}
\caption{F1 fidelity \datasetgcfour}
\end{table}

\begin{table}[ht]
\footnotesize
\begin{tabular}{l|cccccccccccccccc}
Explainers  & \multicolumn{4}{c}{\gcn} & \multicolumn{4}{c}{\gin}  & \multicolumn{4}{c}{\cheb}  & \multicolumn{4}{c}{\sage} \\ \hline
Class & 0 & 1 & 2 & 3 & 0 & 1 & 2 & 3 & 0 & 1 & 2 & 3 & 0 & 1 & 2 & 3 \\\hline
\cam &  & \filter & 0.68 & 0.31 &  & \filter & \filter & \filter &  & \filter & \filter & 0.00 &  & \filter & \filter & \filter \\
\gradcam &  & 0.61 & 0.30 & 0.32 &  & 0.67 & 0.90 & 0.88 &  & 0.97 & 1.00 & 0.99 &  & 0.94 & 0.99 & 0.99 \\
\guidedbp &  & 0.65 & 0.84 & 0.73 &  & 0.69 & 0.82 & 0.83 &  & 0.64 & 0.74 & 0.65 &  & \filter & \filter & \filter \\
\ignode &  & 0.50 & 0.50 & 0.50 &  & 0.50 & 0.50 & 0.50 &  & 0.50 & 0.75 & 0.50 &  & 0.50 & 0.75 & 0.50 \\
\gradexp &  & 0.30 & 0.04 & 0.08 &  & 0.20 & 0.00 & 0.02 &  & 0.64 & 0.60 & 0.20 &  & 0.60 & 0.85 & 0.40 \\
\pgmexpl &  & 0.49 & \filter & \filter &  & 0.50 & \filter & \filter &  & \filter & \filter & \filter &  & \filter & \filter & \filter \\
\gnnexpl &  & 0.62 & 0.81 & 0.77 &  & 0.61 & 0.86 & 0.97 &  &  &  &  &  & 0.95 & 0.94 & 1.00 \\
\pgexpl &  & 0.54 & 0.96 & 0.93 &  & \filter & \filter & \filter &  & \filter & \filter & \filter &  & \filter & \filter & \filter \\
\subgraphx &  & 0.49 & 0.50 & 0.49 &  & 0.50 & 0.49 & 0.48 &  & 0.51 & 0.50 & 0.50 &  & 0.51 & 0.50 & 0.50 \\
\rgexpl &  &  &  &  &  & 0.62 & 0.89 & 0.98 &  & 0.98 & 0.99 & 0.99 &  & 0.79 & 0.82 & 0.71
\end{tabular}
\caption{Plausibility \datasetncone}
\end{table}

\begin{table}[ht]
\scriptsize
\begin{tabular}{l|cccccccccccccccc}
Explainers  & \multicolumn{4}{c}{\gcn} & \multicolumn{4}{c}{\gin}  & \multicolumn{4}{c}{\cheb}  & \multicolumn{4}{c}{\sage} \\ \hline
Class & 0 & 1 & 2 & 3 & 0 & 1 & 2 & 3 & 0 & 1 & 2 & 3 & 0 & 1 & 2 & 3 \\\hline
\cam & 0.13 & \filter & 0.35 & 0.58 & \filter & \filter & \filter & \filter & 0.74 & \filter & \filter & 0.56 & \filter & \filter & \filter & \filter \\
\gradcam & 0.03 & 0.81 & 0.60 & 0.54 & 0.74 & 0.63 & -0.09 & 0.43 & 0.79 & 0.40 & 0.59 & 0.64 & 0.46 & 0.35 & 0.39 & 0.62 \\
\guidedbp & 0.00 & 0.79 & 0.47 & 0.49 & 0.89 & 0.47 & -0.08 & 0.40 & 0.80 & 0.71 & 0.65 & 0.65 & \filter & \filter & \filter & \filter \\
\ignode & 0.00 & 0.70 & 0.56 & 0.35 & 0.77 & 0.61 & -0.08 & 0.48 & 0.14 & 0.09 & 0.90 & 0.43 & 0.44 & 0.17 & 0.09 & 0.44 \\
\gradexp & \filter & 0.89 & 0.74 & 0.73 & \filter & 0.39 & -0.10 & 0.48 & 0.44 & 0.50 & 0.32 & 0.06 & \filter & 0.15 & 0.01 & 0.45 \\
\pgmexpl & -0.01 & 0.49 & \filter & \filter & 0.50 & 0.44 & \filter & \filter & 0.29 & \filter & \filter & \filter & 0.49 & \filter & \filter & \filter \\
\gnnexpl & 0.12 & 0.55 & 0.46 & 0.44 & 0.74 & 0.34 & -0.09 & 0.41 &  &  &  &  & 0.76 & 0.48 & 0.22 & 0.69 \\
\pgexpl & 0.02 & 0.06 & -0.35 & -0.15 & \filter & \filter & \filter & \filter & \filter & \filter & \filter & \filter & \filter & \filter & \filter & \filter \\
\subgraphx & 0.01 & 0.84 & 0.77 & 0.63 & 0.87 & 0.69 & 0.09 & 0.47 & 0.72 & 0.67 & 0.13 & 0.64 & 0.55 & 0.48 & 0.44 & 0.41 \\
\rgexpl &  &  &  &  & 0.92 & 0.64 & -0.10 & 0.35 & 0.63 & 0.21 & 0.22 & 0.47 & 0.91 & 0.68 & -0.01 & 0.87
\end{tabular}
\caption{F1 fidelity \datasetncone}
\end{table}

\begin{table}[ht]
\footnotesize
\begin{tabular}{l|ccccccccccccccc}
Explainers  & \multicolumn{3}{c}{\gcn} & \multicolumn{3}{c}{\gat}  & \multicolumn{3}{c}{\gin}  & \multicolumn{3}{c}{\cheb}  & \multicolumn{3}{c}{\sage} \\ \hline
Class & 0 & 1 & 2 & 0 & 1 & 2 & 0 & 1 & 2 & 0 & 1 & 2 & 0 & 1 & 2 \\ \hline
\cam & \filter & \filter & \filter & 0.72 & 0.68 & 0.71 & 0.49 & 0.58 & 0.10 & 0.81 & 0.78 & 0.90 & 0.48 & 0.80 & 0.90 \\
\gradcam & 1.00 & 0.81 & 0.92 & 0.75 & 0.64 & \filter & 0.49 & 0.64 & 0.90 & 0.52 & 0.58 & 0.63 & 0.49 & 0.67 & 0.92 \\
\guidedbp & 1.00 & 0.58 & 0.59 & 0.74 & 0.49 & 0.37 & 0.75 & 0.64 & 0.55 & 0.58 & 0.53 & 0.65 & 0.77 & 0.58 & 0.76 \\
\ignode & \filter & 0.72 & 0.90 & \filter & 0.79 & \filter & 0.50 & 0.73 & 0.87 & 0.74 & 0.75 & 0.89 & 0.78 & 0.79 & 0.92 \\
\gradexp & 0.00 & 0.23 & 0.90 & 0.25 & 0.67 & 0.37 & 0.25 & 0.22 & \filter & 0.20 & 0.48 & 0.40 & 0.20 & 0.48 & 0.90 \\
\pgmexpl & \filter & 0.55 & 0.10 & \filter & 0.56 & \filter & 0.50 & 0.10 & \filter & \filter & \filter & \filter & \filter & \filter & \filter \\
\gnnexpl & \filter & 0.50 & 0.50 & 0.50 & 0.50 & 0.50 & \filter & \filter & \filter & \filter & \filter & \filter & 0.50 & 0.50 & 0.50 \\
\pgexpl & \filter & \filter & \filter & \filter & \filter & \filter & \filter & \filter & \filter & \filter & \filter & \filter & \filter & \filter & \filter \\
\subgraphx & 0.50 & 0.50 & 0.50 & 0.50 & 0.50 & 0.50 & 0.50 & 0.50 & 0.50 & 0.50 & 0.50 & 0.50 & 0.50 & 0.50 & 0.50 \\
\rgexpl &  &  &  &  &  &  &  &  &  &  &  &  &  &  & 
\end{tabular}
\caption{Plausibility \datasetnctwo}
\end{table}

\begin{table}[ht]
\footnotesize
\begin{tabular}{l|ccccccccccccccc}
Explainers  & \multicolumn{3}{c}{\gcn} & \multicolumn{3}{c}{\gat}  & \multicolumn{3}{c}{\gin}  & \multicolumn{3}{c}{\cheb}  & \multicolumn{3}{c}{\sage} \\ \hline
Class & 0 & 1 & 2 & 0 & 1 & 2 & 0 & 1 & 2 & 0 & 1 & 2 & 0 & 1 & 2 \\ \hline

\cam & \filter & \filter & \filter & -0.41 & 0.78 & 0.75 & 0.05 & 0.53 & \filter & 0.49 & -0.11 & 0.63 & 0.64 & -0.18 & 0.77 \\
\gradcam & 0.55 & 0.31 & 0.51 & -0.38 & 0.77 & \filter & -0.50 & 0.50 & 0.60 & 0.58 & 0.06 & 0.50 & 0.64 & -0.19 & 0.78 \\
\guidedbp & 0.10 & 0.53 & 0.52 & -0.41 & 0.77 & 0.78 & -0.03 & 0.20 & 0.68 & 0.48 & 0.22 & 0.37 & 0.55 & -0.19 & 0.80 \\
\ignode & \filter & 0.27 & 0.64 & \filter & 0.78 & \filter & -0.47 & 0.56 & 0.66 & 0.58 & -0.05 & 0.64 & 0.63 & -0.19 & 0.78 \\
\gradexp & 1.00 & 0.72 & -0.21 & -0.43 & 0.78 & 0.80 & 0.30 & 0.04 &  \filter & 0.61 & 0.05 & 0.51 & 0.64 & -0.19 & 0.78 \\
\pgmexpl & \filter & 0.54 & \filter & \filter & 0.77 & \filter & -0.35 &  \filter &  \filter & \filter & \filter & \filter & \multicolumn{1}{l}{\filter} & \multicolumn{1}{l}{\filter} & \multicolumn{1}{l}{\filter} \\
\gnnexpl & \filter & 0.91 & 0.00 & 0.13 & 0.95 & 0.86 &  \filter &  \filter &  \filter & \filter & \filter & \filter & 1.00 & 0.00 & 1.00 \\
\pgexpl & \filter & \filter & \filter & \filter & \filter & \filter & \filter & \filter & \filter & \filter & \filter & \filter &  \filter &  \filter &  \filter \\
\subgraphx & 1.00 & 0.76 & 0.20 & 0.30 & 0.77 & 0.66 & 0.99 & 0.77 & 0.20 & \filter & \filter & \filter & 0.64 & 0.19 & 0.77 \\
\rgexpl &  &  &  &  &  &  &  &  &  &  &  &  &  &  & 
\end{tabular}
\caption{F1 fidelity \datasetnctwo}
\end{table}

%% Putting figures
\section{Additional explanation examples}\label{sup:tab}
In the following we report additional plots of explanations corresponding to random samples for each dataset, relative to the most explainable architecture (RQ1.2) as reported in the upper part of Table \ref{tab:rq1_all} and  \ref{tab:rq1_all_nc} (plausibility), for respectively graph and node classification. Missing entries in the plot are referred to explanations that did not meet the criteria outlined in Section \ref{sec:aggregation}.

\begin{figure}[ht]
    \centering
    \includegraphics[width=0.75\textwidth]{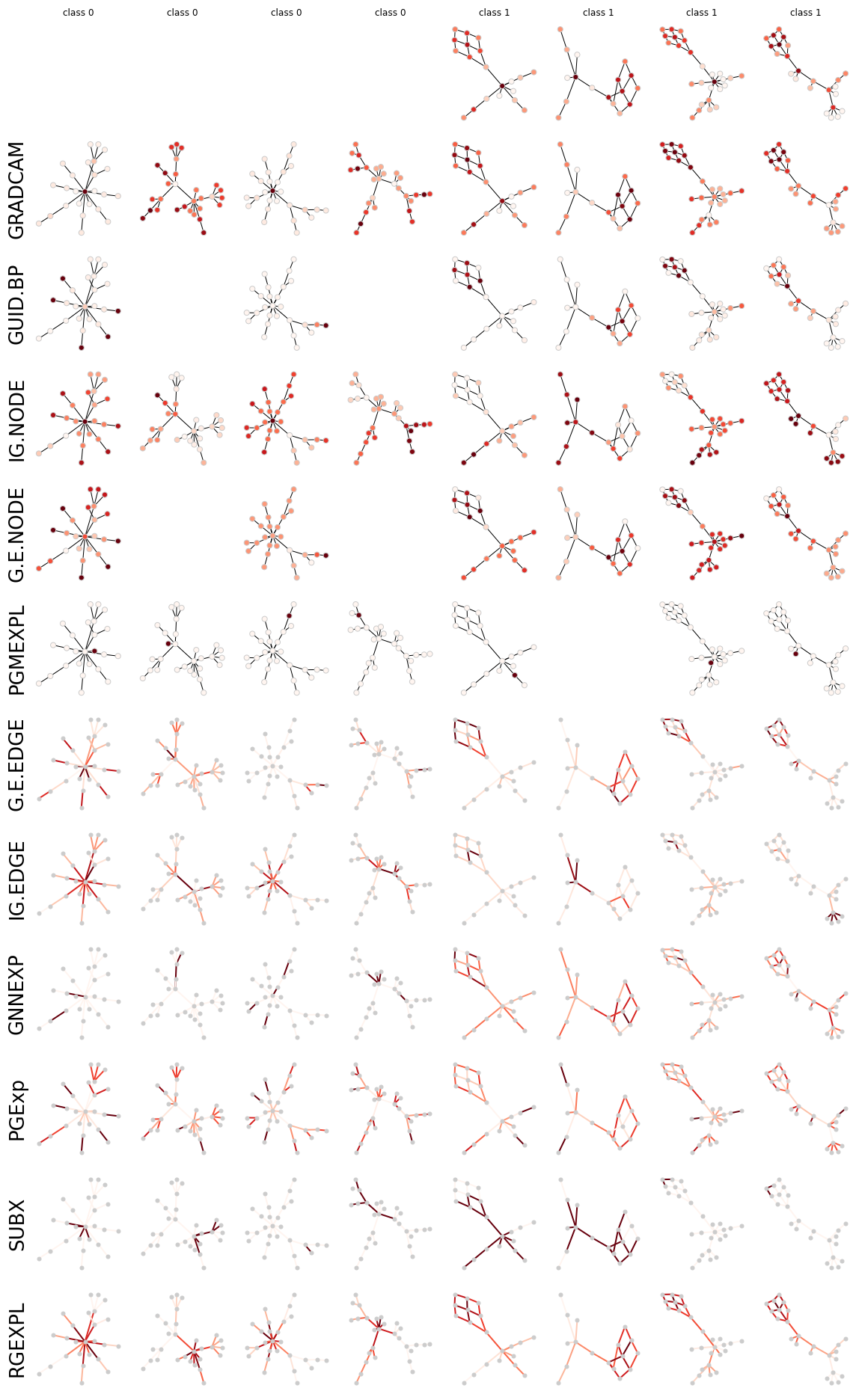}
    \caption{\rev{Examples of \gcn explanations for all explainers on the \datasetgcone dataset.}}
\end{figure}
\begin{figure}[ht]
    \centering
    \includegraphics[width=0.75\textwidth]{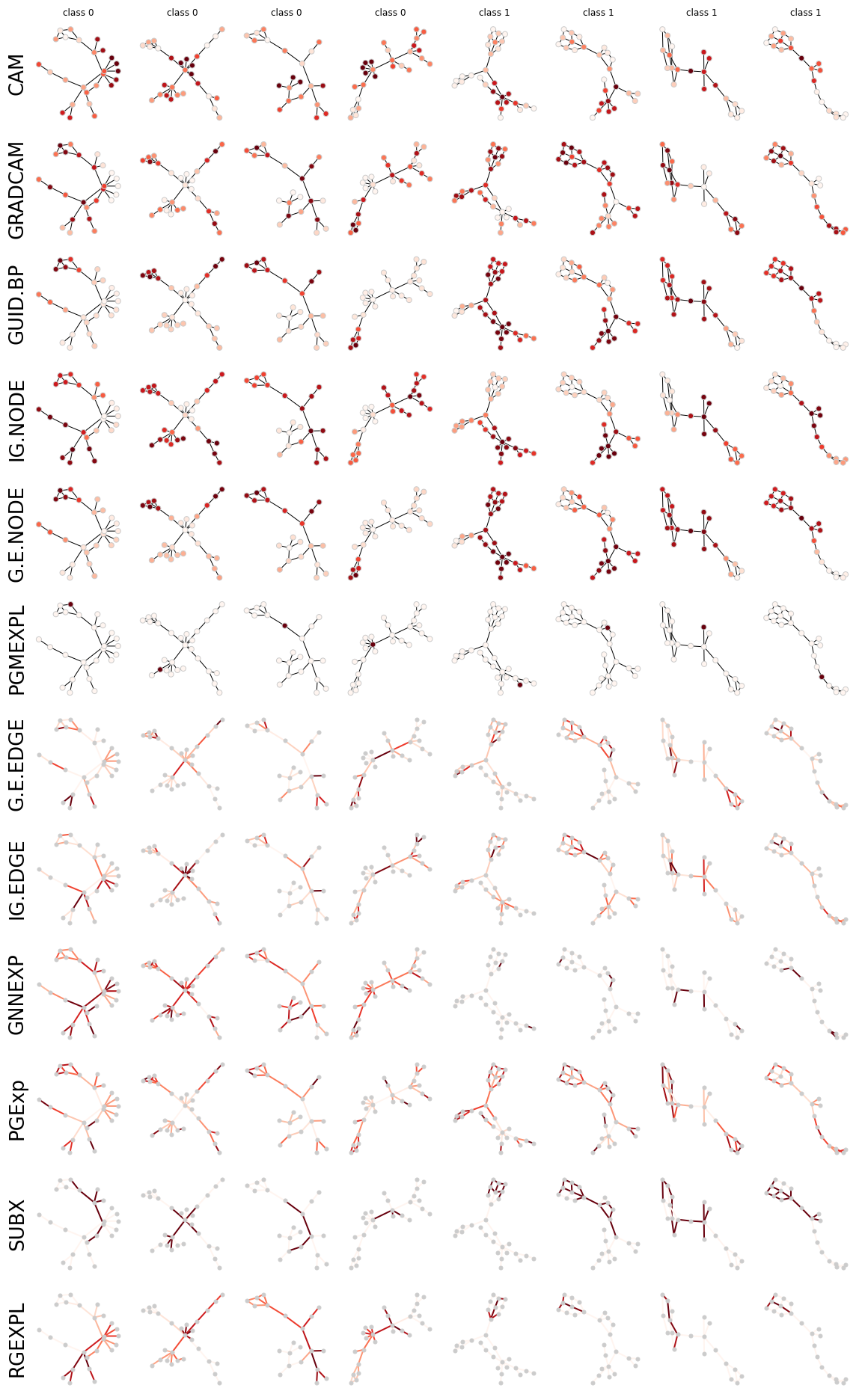}
    \caption{\rev{Examples of \gcn explanations for all explainers on the \datasetgctwo dataset.}}
\end{figure}
\begin{figure}[ht]
    \centering
    \includegraphics[width=0.75\textwidth]{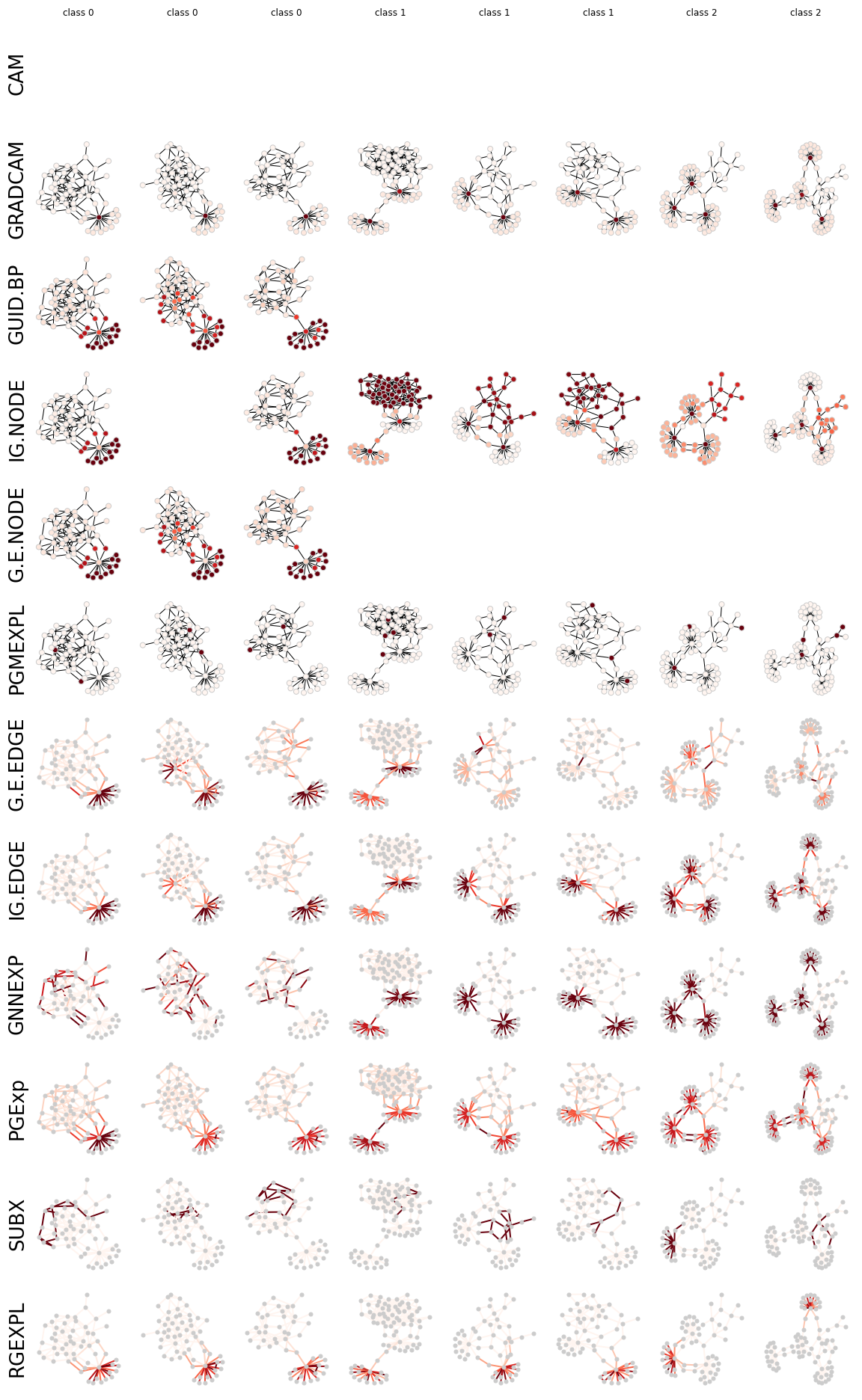}
    \caption{\rev{Examples of \setset explanations for all explainers on the \datasetgcthree dataset.}}
\end{figure}
\begin{figure}[ht]
    \centering
    \includegraphics[width=0.75\textwidth]{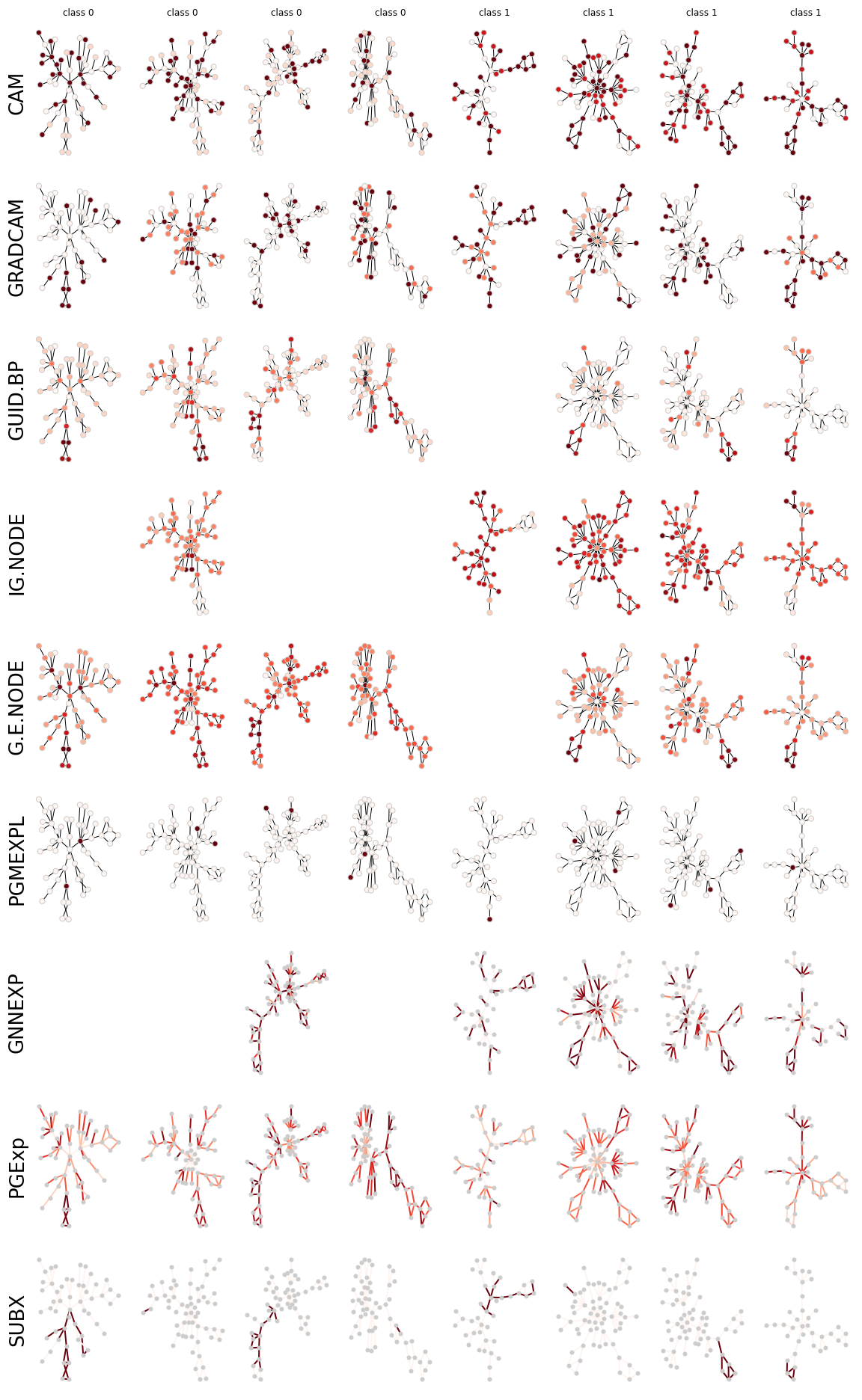}
    \caption{\rev{Examples of \mincut explanations for all explainers on the \datasetgcfour dataset.}}
\end{figure}
\begin{figure}[ht]
    \centering
    \includegraphics[width=0.55\textwidth]{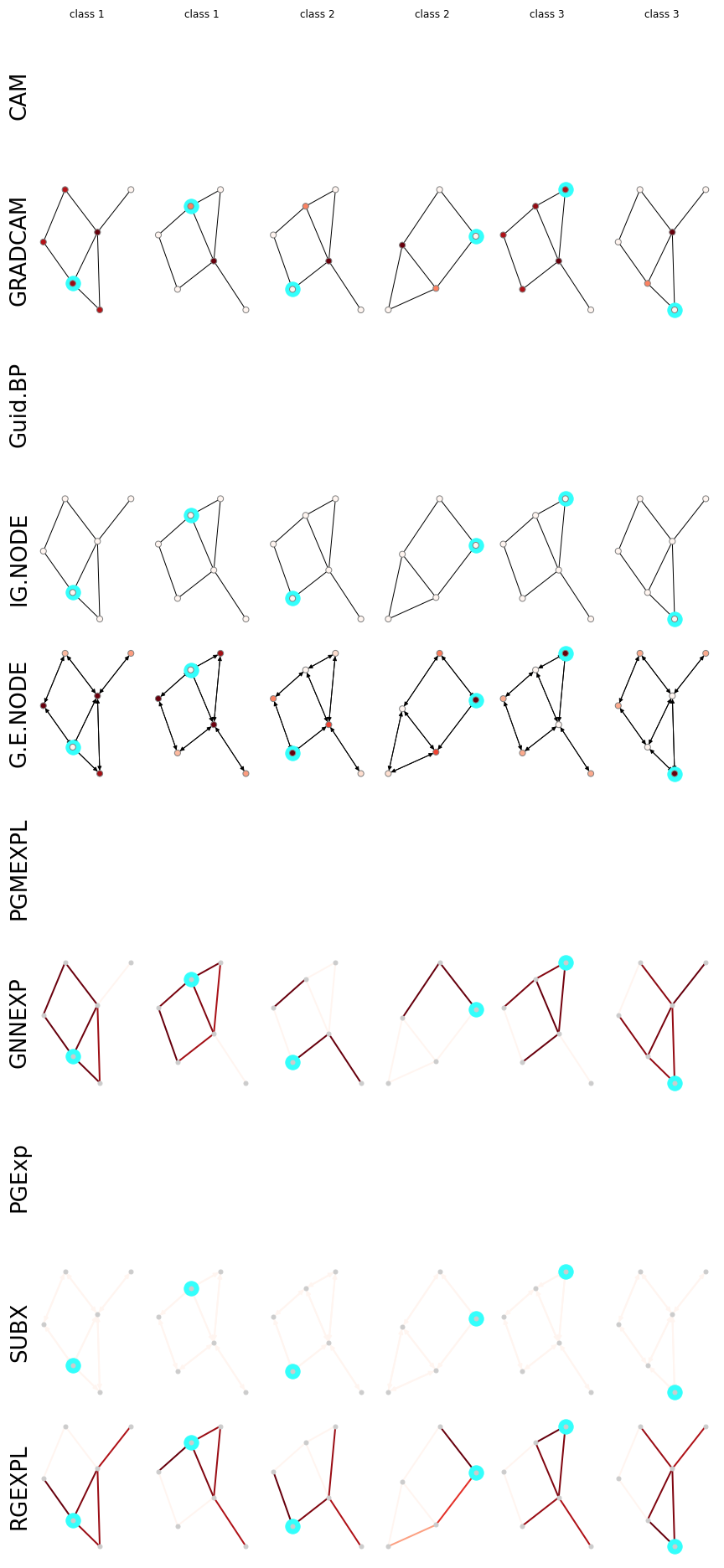}
    \caption{\rev{Examples of \sage explanations for all explainers on the \datasetncone dataset.}}
\end{figure}
\begin{figure}[ht]
    \centering
    \includegraphics[width=0.55\textwidth]{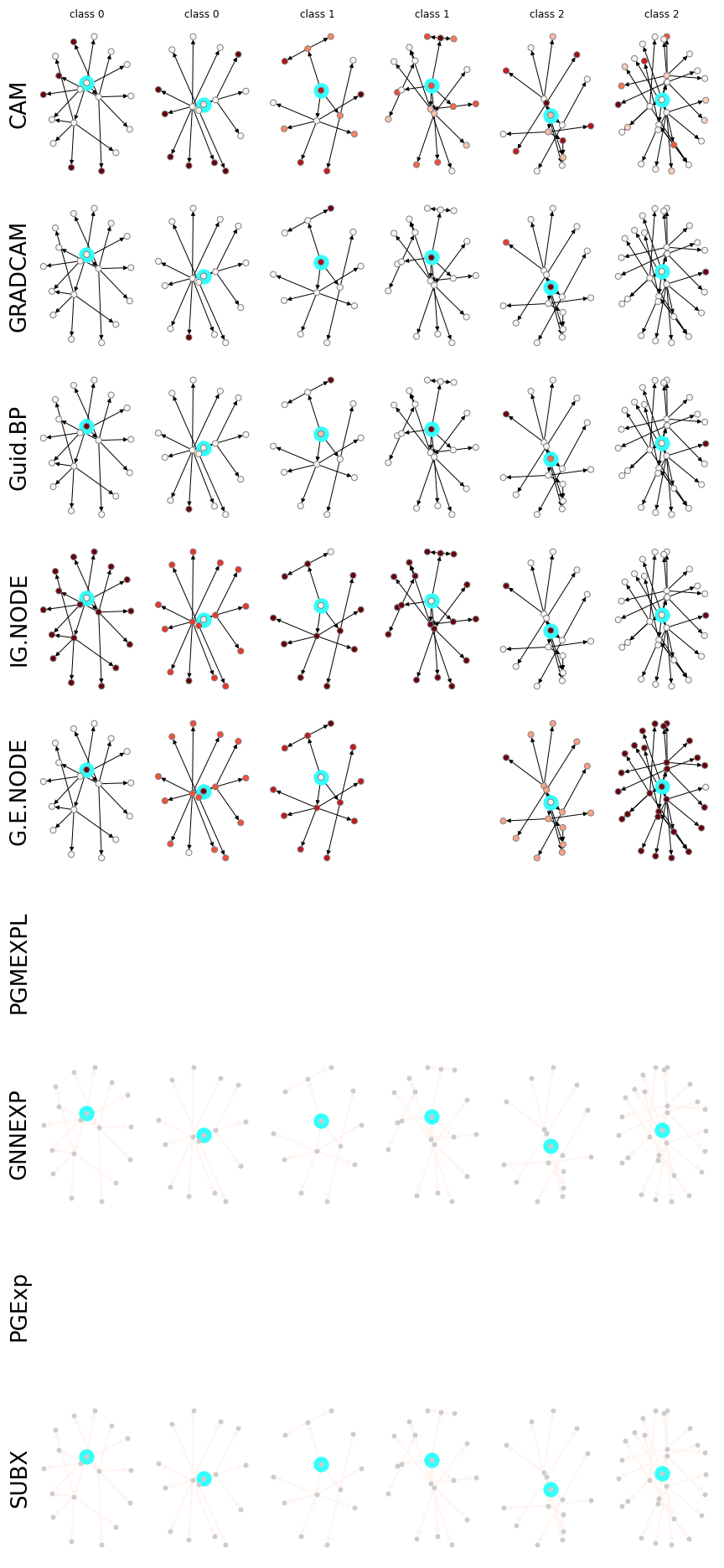}
    \caption{\rev{Examples of \sage explanations for all explainers on the \datasetnctwo dataset.}}
\end{figure}

\end{document}